%% file: egpaper_for_review.tex
\documentclass[10pt,twocolumn,letterpaper]{article}

\usepackage{iccv}
\usepackage{times}
\usepackage{epsfig}
\usepackage{graphicx}
\usepackage{amsmath}
\usepackage{amssymb}

\usepackage{pifont}
\newcommand{\cmark}{\ding{51}}%
\newcommand{\xmark}{\ding{55}}%
\usepackage{booktabs}
\usepackage{csquotes}
\usepackage{diagbox}
\usepackage{multirow}
\usepackage{multicol}
\usepackage{bbm}
\usepackage{algorithm}
\usepackage{algpseudocode}
\usepackage{xcolor}
\usepackage{subcaption}
\usepackage{graphicx}
\usepackage{tabularray}
\usepackage{array}
\newcolumntype{C}[1]{>{\centering\arraybackslash}p{#1}}
\newcolumntype{L}[1]{>{\raggedright\arraybackslash}p{#1}}
\usepackage[normalem]{ulem}
\usepackage{etoc}

\usepackage[pagebackref=true,breaklinks=true,letterpaper=true,colorlinks,bookmarks=false]{hyperref}

\iccvfinalcopy 

\def\iccvPaperID{4163} 

\ificcvfinal\fi

\begin{document}

\title{Model Calibration in Dense Classification with Adaptive Label Perturbation}

\author{Jiawei Liu \;\, Changkun Ye \;\, Shan Wang \;\, Ruikai Cui \;\, Jing Zhang \;\, Kaihao Zhang \;\, Nick Barnes
\\
The Australian National University\\
}

\maketitle
\ificcvfinal\thispagestyle{empty}\fi


\begin{abstract}
For safety-related applications, it is crucial to produce trustworthy deep neural networks whose prediction is associated with confidence that can represent the likelihood of correctness for subsequent decision-making. Existing dense binary classification models are prone to being over-confident. To improve model calibration, we propose Adaptive Stochastic Label Perturbation (ASLP) which learns a unique label perturbation level for each training image. ASLP employs our proposed Self-Calibrating Binary Cross Entropy (SC-BCE) loss, which unifies label perturbation processes including stochastic approaches (like DisturbLabel), and label smoothing, to correct calibration while maintaining classification rates. ASLP follows Maximum Entropy Inference of classic statistical mechanics to maximise prediction entropy with respect to missing information. It performs this while: (1) preserving classification accuracy on known data as a conservative solution, or (2) specifically improves model calibration degree by minimising the gap between the prediction accuracy and expected confidence of the target training label. Extensive results demonstrate that $\text{ASLP}$ can significantly improve calibration degrees of dense binary classification models on both in-distribution and out-of-distribution data. The code is available on \url{https://github.com/Carlisle-Liu/ASLP}.
\end{abstract}

\etocdepthtag.toc{mtchapter}
\etocsettagdepth{mtchapter}{subsection}
\etocsettagdepth{mtappendix}{none}

\section{Introduction}
\label{sec:introduction}
Binary segmentation aims to differentiate foreground areas from the background in images. Its tasks include Salient Object Detection~\cite{ucnet++}, Camouflaged Object Detection~\cite{camouflaged_obj}, Smoke Detection~\cite{yan2022transmission}, \etc. Performance in these tasks has been significantly advanced using the strong representation powers of Deep Neural Networks (DNNs). However, with complex structures and a tremendous number of parameters, DNNs are prone to over-fitting to training data and producing over-confident predictions in the real world \cite{guo2017calibration}. Such issues can render the model predictions unreliable in decision making or utilisation in downstream tasks. 

\begin{figure}[t!]
    \centering
    \includegraphics[width=\linewidth]{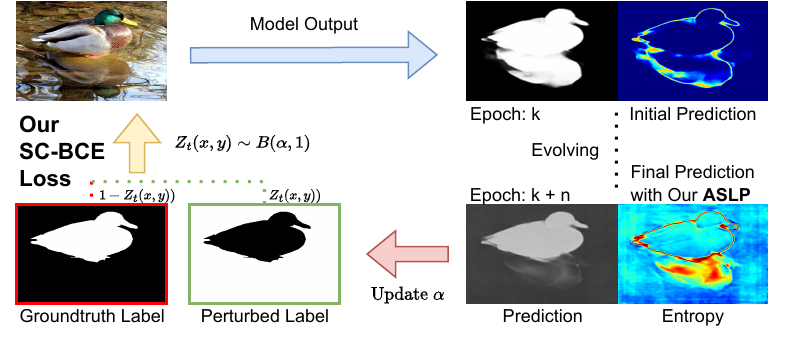}
    \vspace{-20pt}
    \caption{Applying Adaptive Label Perturbation during training can effectively moderate predictions at incorrect areas, highlighting them with high entropy values (red). $Z_{t}(x, y)$ is a sample-wise Bernoulli variable, parameterized by $\alpha$, at the $t^{\text{th}}$ iteration. After $k$ iterations, update $\alpha$ using Eq.~\eqref{eq:ASLP_MEI_Learning_Rule} to adjust the likelihood (or level) of label perturbation to increase
    entropy for incorrect predictions and so correct model calibration. 
    The Perturbed Label (shown inverted) replaces the Groundtruth Label with probability $\alpha$.
    }
    \label{fig:intro_figure}
    \vspace{-10pt}
\end{figure}

Recently, a growing body of literature has been proposed to address model mis-calibration problems in DNNs. They can be roughly categorised as: (1) post-hoc operations, such as temperature scaling \cite{guo2017calibration}, Platt scaling \cite{platt1999probabilistic}, \etc, (2) training objective approaches \cite{karandikar2021soft}, like MMCE \cite{MMCE}, soft calibration objective \cite{karandikar2021soft}, focal loss \cite{focal_loss,ghosh2022adafocal}, and (3) data/label augmentation techniques, \eg label smoothing \cite{muller2019does} and mixup \cite{zhang2018mixup}. We propose an Adaptive Label Perturbation which learns a unique label perturbation level for each training image. As illustrated in Fig.~\ref{fig:intro_figure}, training with Adaptive Stochastic Label Perturbation, a form of ALP, can effectively moderate incorrect predictions and highlight them with high entropy values.

Adaptive Label Perturbation employs our proposed Self-Calibrating Binary Cross Entropy (SC-BCE) loss, which unifies label perturbation processes including stochastic approaches (like DisturbLabel \cite{Xie_2016_CVPR}), and label smoothing \cite{szegedy2016rethinking} to correct calibration while maintaining classification accuracy.
SC-BCE loss is equivalent to 
a factored combination of (i) a BCE loss \wrt groundtruth label, and (ii) a BCE loss \wrt a uniform binary categorical distribution.
The former enhances dense binary classification performance and the latter improves the model calibration degree. 
Our method can be connected to Maximum Entropy Inference \cite{jaynes1957information} of classic statistical mechanics, to maximise prediction entropy with respect to missing information while preserving the classification accuracy on known data.

The proposed Adaptive Label Perturbation (ALP) can approximate Maximum Entropy Inference \cite{jaynes1957information} to maximise prediction entropy while preserving the ideal dense classification performance on known data. This represents a conservative solution that adopts classification accuracy as a proxy for known data and assumes maximum disorder on unknown data. 
We also present an alternative ALP solution that, instead, takes model calibration degree as a proxy for known data, using a calibration regulariser which constrains the expected confidence of individual supervision signal to not  drop below the ideal accuracy on the validation set.
This effectively minimises the gap between the distributions of prediction confidence and prediction accuracy, which is the source of model mis-calibration.


Our contributions can be summarised as: 
(i) We propose Adaptive Stochastic Label Perturbation that 
learns a sample-wise label perturbation level to improve model calibration; 
(ii) We present a Self-Calibrating Binary Cross Entropy loss that unifies label perturbation processes including stochastic approaches and label smoothing;
(iii) Following Maximum Entropy Inference \cite{jaynes1957information}
we show that Adaptive Stochastic Label Perturbation ($\text{ASLP}_{\text{MEI}}$), can maximise the prediction entropy while preserving the ideal dense classification accuracy, and 
(iv) We present an alternative Adaptive Stochastic Label Perturbation ($\text{ASLP}_{\text{MC}}$) solution to maximise model calibration degree, which achieves state-of-the-art performance in terms of model calibration degree on both in-distribution and out-of-distribution data. We thoroughly evaluate our method on Salient Object Detection and demonstrate its effectiveness for Camouflaged Object Detection, Smoke Detection and Semantic Segmentation.


\section{Related Works}
\label{sec:related_works}
\noindent{\bf Model Calibration:}
Model calibration methods can be roughly divided into three categories of approach: (1) post hoc processing (2) training object, and (3) input/label augmentation. The first category utilises a validation set to align the prediction confidence distribution with prediction accuracy distribution. It includes histogram binning \cite{zadrozny2001obtaining}, isotropic regression \cite{zadrozny2002transforming}, Platt scaling \cite{platt1999probabilistic,niculescu2005predicting}, Bayesian binning into quantiles \cite{naeini2015obtaining}, Dirichlet scaling \cite{kull2019beyond}, mix-n-match \cite{zhang2020mix} and temperature scaling \cite{guo2017calibration,yu2022robust}. 

The second category focuses on designing training objectives that (in)directly improve  model calibration degree. Some methods address the in-continuity of expected calibration error, a widely adopted model calibration measure, and propose trainable calibration objectives like maximum mean calibration error \cite{MMCE}, soft calibration objective \cite{karandikar2021soft}. Other works discover that certain existing training objects are beneficial to model calibration, \eg Brier loss \cite{brier1950verification,degroot1983comparison}, confidence penalty \cite{pereyra2017regularizing} and focal loss \cite{focal_loss,ghosh2022adafocal}.

The third category employs data or label augmentation techniques to regularise the prediction confidence distribution. Mixup \cite{zhang2018mixup,thulasidasan2019mixup} explores the neighbourhood of training data through random interpolation of input images and associated labels to improve model calibration degree. Label smoothing \cite{muller2019does} augments the one-hot training labels with softer versions to prevent the model being over-confident.

\noindent{\bf Salient Object Detection:}
Inspired by pioneering work \cite{itti1998model}, traditional Saliency Object Detection (SOD) methods rely on various heuristic priors with handcraft features to explore low-level cures \cite{achanta2009frequency,jiang2013salient,jiang2013submodular,liu2010learning,wang2016kernelized}. However, these methods cannot cope with complex scenes because of the limited representation ability of handcrafted features \cite{borji2019salient}.
%
%
Recently, deep learning based SOD methods broke the bottleneck of traditional methods due to the powerful capability of neural networks, achieving improvemed performance \cite{chen2018reverse,ji2020context,qin2019basnet,xu2021locate,liu2018picanet,wang2016saliency,wang2019iterative}. 
Early deep SOD methods use multi-layer perception to predict a map with a pixel-wise score for each image \cite{wei2012geodesic,cheng2014global}. These approaches rely on fixed fully connected layers and thus severely limit the ability of spatial information extraction. Later methods address this issue via using fully convolutional networks (FCNs) \cite{long2015fully}.

Most contemporary SOD methods are designed based on FCNs with various schemes to improve performance. One of the most popular strategies is to fuse multi-scale information extracted from different layers and aggregate them in the network \cite{zhang2018bi,wu2019stacked,gao2020highly,pang2020multi,zhao2020suppress,zhang2020weakly}. 
Attention modules are also applied to capture powerful multi-scale features via exploring relationship between local and global information \cite{liu2018picanet,piao2019depth,zhao2019pyramid,hu2020sac}.
Training SOD networks using auxiliary supervision is also a popular strategy \cite{wei2020label,wei2020label,tang2021disentangled}. For example, the body map and detail map are utilized by \cite{wei2020label} to help the network focus on center areas and edges, respectively. Skeleton \cite{liu2020dynamic} and uncertainty \cite{tang2021disentangled} are applied to the training processing due to their important roles in taking photos.

\noindent{\bf Noisy Label:}
Noisy labels refer to incorrect ground truth classes/values in classification/regression tasks. They arise from data collection or annotation processes, and exist commonly in real-world datasets \cite{algan2021image}. Efforts are put to identify the noisy labels and exclude them from network training in various computer vision tasks, deeming their incorporation as harmful. \cite{zhang2020learning} proposes a framework that learns from noisy labels, being a collection of predictions from classic SOD methods. The framework approximates the noise distribution in order to recover clean labels for model training. 

Differently from data augmentation techniques that are applied simultaneously to training samples and corresponding labels to generate more training data, one may artificially corrupt the label. We refer to this category of approaches as {\em label perturbation}, which includes label smoothing approaches 
\cite{szegedy2016rethinking}, and DisturbLabel, Xie \etal \cite{Xie_2016_CVPR}. In image classification, Xie \etal \cite{Xie_2016_CVPR} shows that randomly replacing training labels with a prior distribution leads to a regularising effect, preventing overfitting. Our work is different from \cite{Xie_2016_CVPR} by employing labels corrupted to different scales to enhance the model calibration degrees for both in-distribution and out-of-distribution data. Further, in performing this, we assume noise that varies with different samples, making our method more adaptable.


\begin{figure*}
    \centering
    \includegraphics[width=0.9\textwidth]{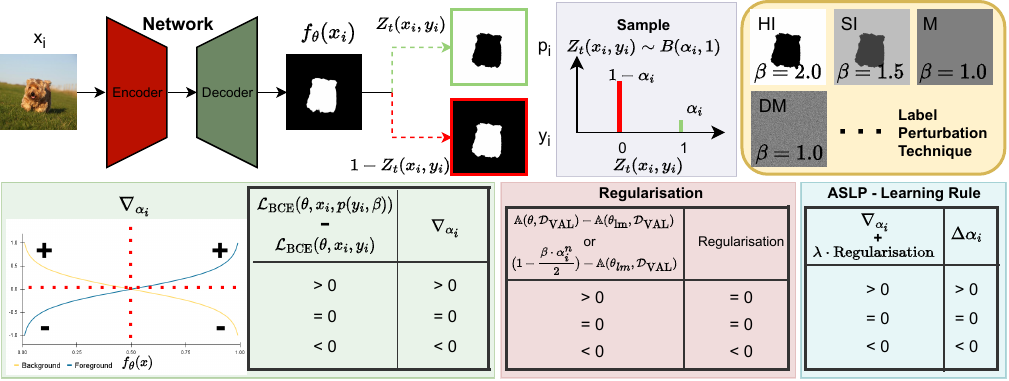}
    \vspace{-8pt}
    \caption{The method overview is comprised of model implementation (top) and Adaptive Stochastic Label Perturbation (ASLP) learning rule (bottom). In each iteration, the model uses a Bernoulli variable to sample a supervision, which can be a groundtruth label $y_{i}$ or a perturbed label $p_{i}$ and computes a sample-specific $\nabla_{\alpha_{i}}$ based on the prediction. 
    Regularisation 
    is computed with Eq.~\ref{eq:ASLP_MEI_Learning_Rule} ($\text{ASLP}_{\text{MEI}}$) or Eq.~\ref{eq:Calibration_Regularisation_for_ASLP_MC} ($\text{ASLP}_{\text{MC}}$) on a validation set after each training epoch. The ASLP learning rule combines Grad-$\alpha_{i}$ and a factored regularisation to update label perturbation probability $\alpha_{i}$ for each individual sample.
    }
    \label{fig:ASLP_ECE_overview}
    \vspace{-10pt}
\end{figure*}

\section{Proposed Method}
We first lay out the task setting in Sec.~\ref{sub_sec:task_setting}. Then we introduce our proposed Self-Calibrating Binary Cross Entropy loss in Sec.~\ref{sub_sec:sc_bce_loss} and prove its connection to Maximum Entropy Inference \cite{jaynes1957information} in Sec.~\ref{sub_sec:maximum_entropy_inference}. Lastly, we detail our major contribution - Adaptive Label Perturbation in Sec.~\ref{sub_sec:adaptive_stochastic_label_perturbation}.

\subsection{Task Setting}
\label{sub_sec:task_setting}
Binary segmentation problems aim to differentiate between foreground object(s) and background. They can be formulated as a pixel-wise binary classification problem. Given an independent and identically distributed (i.i.d) training dataset $\mathcal{D}_{\text{TR}} = \{x_{i}, y_{i}\}_{i=1}^{N}$ drawn from an unknown joint distribution of training images and groundtruth labels $P(\mathcal{X}, \mathcal{Y})$, a neural network model parameterised by $\theta$
is employed to predict labels for an input image $x \in \mathcal{X}$: $f_{\theta}(x) \in (0, 1)^{1 \times H \times W}$. We use $\hat{y}$ and $P_{\hat{y}}$ to denote the winning class and its associated probability respectively. 
The groundtruth label $y \in \{0, 1\}^{1 \times H \times W}$  represents the foreground pixels with \enquote{1} and background with \enquote{0}. In the following equations, we omit the spatial indexes $H$ and $W$ for simplicity.
A perfectly calibrated model has $\mathrm{P}(\hat{y} = y | P_{\hat{y}}) = P_{\hat{y}}, \, \forall P_{\hat{y}} \in (0, 1)$. 
That is, in the entire range of prediction probabilities, prediction with probability $P_{\hat{y}}$ has exactly $P_{\hat{y}}$ chance to be correct. 
The calibration degree of a model $f_{\theta}(\cdot)$ over a distribution $\mathcal{D}$ is quantified with Expected Calibration Error (ECE), defined as $\mathbb{E}_{f_{\theta}(x)} [ | \mathrm{P}(\hat{y} = y | f_{\theta}(x)) - f_{\theta}(x) | ]$. 



\subsection{Self-Calibrating Binary Cross Entropy Loss}
\label{sub_sec:sc_bce_loss}
We propose a Self-Calibrating Binary Cross Entropy (SC-BCE) loss that unifies Label Smoothing \cite{szegedy2016rethinking}, DisturbLabel \cite{Xie_2016_CVPR} and Stochastic Label Perturbation as:
\begin{equation}
\begin{aligned}
    & \mathcal{L}_{\text{SC-BCE}}(\theta, X, Y, \alpha, \beta) \\ 
    = & \mathbb{E}_{x,y \in X,Y} \Bigl[ (1 - Z_{t}(x,y)) \cdot \mathcal{L}_{\text{BCE}}(\theta, x, y) \\
    & \qquad \qquad + Z_{t}(x,y) \cdot \mathcal{L}_{\text{BCE}}(\theta, x, p(y, \beta)) \Bigr]\\
    & \beta \in [0, 2], \quad \alpha \in [0, \frac{1}{\beta}),
\end{aligned}
\label{eq:SC-BCE_loss_definition}
\end{equation}
where $Z_{t}(x,y) \sim B(1, \alpha)$ follows a Bernoulli distribution with $\alpha$ probability to be $1$ and $1 - \alpha$ chance to be $0$, $t$ denotes the training epoch, $\alpha \in [0, \frac{1}{\beta})$ and $\beta \in [0, 2]$ are Label Perturbation Probability (LPP) and Label Perturbation Strength (LPS) respectively, 
$p(y, \beta) = (1 - \beta) \cdot y + \frac{\beta}{2}, \beta \in [0, 2]$ is a perturbed label
and $\mathcal{L}_{\text{BCE}}(\theta, x, y)$\footnote{$\mathcal{L}_{\text{BCE}}(\theta, x, y) = -y \cdot \log(f_{\theta}(x)) - (1 - y) \cdot \log(1 - f_{\theta}(x))$} is a Binary Cross Entropy (BCE) loss computed for training pair $(x, y)$. For $\alpha=1$, the label perturbation equation follows the label smoothing equation for a binary label \cite{szegedy2016rethinking}. In the proposed SC-BCE loss, different label perturbations can be applied by setting (i) Label Smoothing \cite{szegedy2016rethinking}: $\alpha = 1$ and $\beta \in [0, 1)$, (2) DisturbLabel: $\beta = 1$ and $\alpha \in (0, 1)$, and (3) Stochastic Label Perturbation (SLP): $\beta \in (0, 2]$ and $\alpha \in (0, \frac{1}{\beta})$. For example, Hard Inversion (HI) that inverts the label category as shown in Fig.~\ref{fig:ASLP_ECE_overview}, can be stochastically applied by setting $\beta = 2$ and $\alpha \in (0, \frac{1}{2})$. 

In the implementation of SLP, the supervision for an individual training image in each epoch is sampled by drawing from a Bernoulli distribution. That is, the individual supervision can take the form of the groundtruth label or perturbed label in each training iteration. The overall function of SLP can be connected to that of a smoothed label by taking expectation of the Bernoulli variable: $\mathbb{E}_{Z_{t}}[(1 - Z_{t}(y)) \cdot Y + Z_{t}(y) \cdot p(y, \beta)], \, \forall y \in Y$. 
Taking the expectation over the Bernoulli variation in each iteration is too expensive to implement in model training.
Instead, following \cite{Xie_2016_CVPR}, the expectation of stochastically perturbed label is approximated by taking expectation across training epochs: $\mathbb{E}_{t \in T}[(1 - Z_{t}(y)) \cdot Y + Z_{t}(y) \cdot p(y, \beta)], \, \forall y \in Y$, where $T$ is the total number of training epochs and $Z_{t}(y)$ is a variable drawn from a Bernoulli distribution for the $\text{t}^{\text{th}}$ epoch.

\subsection{Maximum Entropy Inference}
\label{sub_sec:maximum_entropy_inference}
Maximum Entropy Inference (MEI), assuming minimum distribution commitment in respect to missing information, was initially proposed by Jaynes \cite{jaynes1957information}. That is, the probability distribution should have maximum Shannon entropy subject to the partially available information.
Thus, in the complete absence of information, Shannon entropy for a binary prediction defined in Eq.~\eqref{eq:entropy_definition} should be maximised:
\begin{equation}
\begin{aligned}
    \mathbb{H}(f_{\theta}(X)) = \mathbb{E}_{x \in X} \Bigl[&- f_{\theta}(x) \cdot \log f_{\theta}(x) -\\
    & \bigl( 1 - f_{\theta}(x) \bigr) \cdot \log \bigl(1 - f_{\theta}(x) \bigr) \Bigr]
\label{eq:entropy_definition}
\end{aligned}
\end{equation}
For salient object detection and other binary segmentation problems,
maximising 
Eq.~\eqref{eq:entropy_definition} can be achieved with a binary uniform categorical distribution.

The proposed SC-BCE loss can be transformed into a factored combination of a BCE loss \wrt groundtruth label (the constraints of the data) and a BCE loss \wrt a binary uniform categorical distribution (See Appendix~\ref{A_sec:connection_between_SC-BCE_and_MEI} for derivation) as:
\begin{equation}
\begin{aligned}
    &\mathcal{L}_{\text{SC-BCE}}(\theta, X, Y, \alpha, \beta)\\ 
    = & \mathbb{E}_{x,y \in X,Y} \Bigl[ (1 - \beta  Z_{t}(x, y)) \cdot \mathcal{L}_{\text{BCE}}(\theta, x, y)\\
    & \qquad \qquad + \beta Z_{t}(x, y) \cdot \mathcal{L}_{\text{BCE}}(\theta, x, u) \Bigr],
\label{eq:overall_training_loss_definition}
\end{aligned}
\end{equation}
where $Z_{t}(x,y) \sim B(1, \alpha)$, $u$ is a binary uniform categorical distribution, and minimising the second term pushes the prediction distribution towards a uniform binary categorical distribution, equivalently maximising the inference entropy. Therefore, our proposed SC-BCE loss, a combination of a regular BCE loss and a BCE loss with a perturbed label, effectively performs a type of
MEI.
That is, the regular BCE loss component improves the model's binary classification accuracy in the presence of information
while the perturbed label maximizes prediction entropy with respect to missing information in order to close the gap between the available training data and the entire data distribution.



\subsection{Adaptive Label Perturbation}
\label{sub_sec:adaptive_stochastic_label_perturbation}
Stochastic Label Perturbation (SLP) uses a single label perturbation probability and perturbation strength for the entire training dataset.
However, this approach cannot adapt to predictive error that varies for different input images. To address this, we propose an Adaptive Stochastic Label Augmentation (ASLP) method 
to adjust the label augmentation probability for individual training samples. That is we allow the variable in Eq.~\eqref{eq:SC-BCE_loss_definition} to be drawn from a per training image Bernoulli distribution with sample-specific label perturbation probability as:
\begin{equation}
\begin{aligned}
Z_{t}(x, y) \sim B(1, \alpha_{x, y}), \, \forall x, y \in X, Y
\end{aligned}
\label{eq:sample_specific_Bernoulli_variable}
\end{equation}
where $\alpha_{x, y}$ is the label perturbation probability for sample (image-label) pair $(x, y)$. Initially, we set all label perturbation probabilities to $\{\alpha_{i} = 0\}_{i=1}^{N}$ and train a model with a regular BCE loss without label augmentation techniques, which is equivalent to $\mathcal{L}_{\text{SC-BCE}}(\theta, X, Y, \alpha=0, \beta=0)$. The trained model has weight $\theta_{\text{lm}}$ and its accuracy on the validation set, $\mathbb{A}(\theta_{lm}, \mathcal{D}_{\text{VAL}})$, is held as an ideal performance. Subsequently, we select a label perturbation technique and continue to train the model with SC-BCE loss with a learning rule to update the label perturbation probability for individual training samples.

We propose the learning rule for $\alpha$, ($\text{ASLP}_{\text{MEI}}$) to approximate maximum entropy inference. The rule has two components: (1) $\nabla_{\alpha_{i}} = (2 / \beta) \cdot \partial \mathbb{E}_{Z_{t}}(x_{i}, y_{i}) \bigl[ \mathcal{L}_{\text{SC-BCE}}(X, Y, \theta, \{\alpha_{i}\}_{i=1}^{N}, \beta) \bigr] / \partial \alpha_{i}$ is the derivative of the expectation of SC-BCE over the Bernoulli variable \wrt $\alpha_{i}$. We divide this by $\beta / 2$ to ensure that different perturbation techniques (varying $\beta$ values) have the same convergence speed (See derivation in Appendix~\ref{A_sec:Grad_Alpha_Derivation}),
and (2) Accuracy Regularization to encourage maintenance of prediction accuracy. The rule is:
\begin{equation}
\begin{aligned}
    & \alpha_{i}^{n + 1} 
    = \; \alpha_{i}^{n} + \eta \cdot (\nabla_{\alpha_{i}}
    + \lambda \cdot \text{Reg}_A), \ \ \ \text{for} \ i = 1, \dots, N,\\
    &\nabla_{\alpha_{i}} = \frac{2 \cdot \bigl( \mathcal{L}_{\text{BCE}}(\theta, x_{i}, p(y_{i}, \beta)) - \mathcal{L}_{\text{BCE}}(\theta, x_{i}, y_{i}) \bigr) }{\beta}, \\
    & \text{Reg}_{\text{A}} = \min \Bigl(\frac{\mathbb{A}(\theta, \mathcal{D}_{\text{VAL}}) - \mathbb{A}(\theta_{lm}, \mathcal{D}_{\text{VAL}})}{\mathbb{A}(\theta_{lm}, \mathcal{D}_{\text{VAL}}) }, 0\Bigr),
\label{eq:ASLP_MEI_Learning_Rule}
\end{aligned}
\end{equation}
where $\eta$ and $\lambda$ are hyperparameters controlling the updating pace of label perturbation probability and the regularisation strength respectively, $\mathbb{A}(\theta, \mathcal{D}_{\text{VAL}})$ and $\mathbb{A}(\theta_{lm}, and \mathcal{D}_{\text{VAL}})$ denote the current and ideal accuracy on the validation set separately.
$\nabla_{\alpha_{i}}$ 
aims to increase label perturbation probability to confident and correct samples and otherwise for incorrectly classified samples. For example, it returns a large positive value for correct predictions with small BCE loss value \wrt to groundtruth label $y_{i}$ and large BCE loss value \wrt to perturbed label $p(y_{i}, \beta)$.
The \enquote{Accuracy Regularisation} ($\text{Reg}_{\text{A}}$) is designed to reduce the overall perturbation probability if the accuracy on the validation set reduces to be below the local minima. It returns 0 if there is no accuracy drop on the validation set and a large decrease will overwhelm the $\nabla_{\alpha_{i}}$ value and reduce the sample label perturbation probability.
Intuitively, $\text{ASLP}_{\text{MEI}}$ aims to construct a model that preserves the ideal classification accuracy while otherwise maximising the entropy \cite{jaynes1957information}. Note that which particular examples are classified correctly are able to change, but the accuracy is constrained to remain the same. Intuitively, having a model that better captures ignorance may lead to changes in the treatment of test examples that are distant from training distribution. Note, however that adopting classification accuracy as proxy for known data and otherwise maximizing entropy is a conservative strategy 
and we find that it results in the model being significantly under-confident.

The model mis-calibration arises from the distribution mismatch between prediction confidence and prediction accuracy \cite{focal_loss}. We offer an alternative model that uses the model calibration as proxy for known data and maximises the prediction entropy in respect to unknown data $\text{ASLP}_{\text{MC}}$.
The learning rule replaces the \enquote{Accuracy Regularisation} in Eq.~\eqref{eq:ASLP_MEI_Learning_Rule} with a \enquote{Calibration Regularisation} ($\text{Reg}_{\text{C}}$) as:
\begin{equation}
    \text{Reg}_{\text{C}} = \min \Bigl( \bigl(\mathrm{1} - \frac{\beta \cdot \alpha_{i}^{n}}{2} \bigr) - \mathbb{A}(\theta_{lm}, \mathcal{D}_{\text{VAL}}), 0\Bigr), 
\label{eq:Calibration_Regularisation_for_ASLP_MC}
\end{equation}
where $\mathrm{1} - (\beta \cdot \alpha_{i}^{n} / 2)$ denotes the expected confidence of  the perturbed label (Derivation in Appendix~\ref{A_sec:confidence_of_ecpectation_of_augmented_label}). For example, a foreground label \enquote{1} with 5\% chance of being inverted to \enquote{0} has an expected confidence of 0.95.
\enquote{Calibration Regularisation} constrains the expected confidence of perturbed label of each sample to not drop below the ideal classification accuracy on validation set, preventing the model from becoming under-confident. Note that we can also have an updating rule $\text{ALS}_{\text{MC}}$ to learn per-image label perturbation strength (adaptive $\beta$ and fixed $\alpha = 1$) (See Appendix~\ref{A_sec:Adaptive_Label_Smoothing}).


\section{Experiments and Results}
\label{sec:experiments_and_results}
We verify the proposed method primarily on Salient Object Detection and also implement it for Camouflaged Object Detection, Smoke Detection and Semantic Segmentation tasks and report their results in the Appendices.
\subsection{Implementation Details}
\label{sub_sec:implementation_details}
\noindent\textbf{Evaluation Metrics:}
We use Equal-Width Expected Calibration Error ($\text{ECE}_{\text{EW}}$) \cite{guo2017calibration} and Equal-Width Over-confidence Error ($\text{OE}_{\text{EW}}$) \cite{thulasidasan2019mixup} with 10 bins ($B = 10$) to evaluate the model calibration degrees. 
Additionally, we adopt $\text{ECE}_{\text{EM}}$ \cite{nguyen-oconnor-2015-posterior}, $\text{ECE}_{\text{DEBIAS}}$ \cite{kumar2019verified} and $\text{ECE}_{\text{SWEEP}}$ \cite{roelofs2022mitigating} to corroborate with the results of $\text{ECE}_{\text{EW}}$ (See Appendix~\ref{A_sec:model_calibration_benchmark_with_ECE_EM_SWEEP_DEBIAS}).

\noindent\textbf{Datasets:}
The proposed methods are trained with the DUTS-TR \cite{DUTS-TE} training dataset. It is divided into a training set $|\mathcal{D}_{\text{TR}}| = 9,553$ and validation set $|\mathcal{D}_{\text{VAL}}| = 1,000$. We use six testing datasets, including DUTS-TE \cite{DUTS-TE}, DUT-OMRON \cite{DUT-OMRON}, SOD \cite{SOD}, PASCAL-S \cite{PASCAL-S}, ECSSD \cite{ECSSD}, HKU-IS \cite{HKU-IS}, to evaluate the model calibration degree.

\begin{table*}[htb!]
\centering
\scriptsize
\renewcommand{\arraystretch}{1.2}
\renewcommand{\tabcolsep}{1.5mm}
\caption{Salient object detection model calibration degree benchmark. Results are evaluated in with $\text{ECE}_{\text{EW}}$ and $\text{OE}_{\text{EW}}$ with 10 bins (units in \%).
See Appendix~\ref{A_sec:model_calibration_benchmark_with_ECE_EM_SWEEP_DEBIAS} for evaluations with $\text{ECE}_{\text{EM}}$ \cite{nguyen-oconnor-2015-posterior}, $\text{ECE}_{\text{DEBIAS}}$ \cite{kumar2019verified} and $\text{ECE}_{\text{SWEEP}}$ \cite{roelofs2022mitigating}. 
}
\vspace{-2pt}
\begin{tabular}{cl|c|cc|cc|cc|cc|cc|cc}
\toprule
\multicolumn{2}{c|}{\multirow{2}{*}{Methods}} & {\multirow{2}{*}{Year}} & \multicolumn{2}{c|}{DUTS-TE \cite{DUTS-TE}} & \multicolumn{2}{c|}{DUT-OMRON \cite{DUT-OMRON}} & \multicolumn{2}{c|}{PASCAL-S \cite{PASCAL-S}} & \multicolumn{2}{c|}{SOD \cite{SOD}} & \multicolumn{2}{c|}{ECSSD \cite{ECSSD}} & \multicolumn{2}{c}{HKU-IS \cite{HKU-IS}}\\
& & & $\text{ECE} \downarrow$ & $\text{OE} \downarrow$ & $\text{ECE} \downarrow$ & $\text{OE} \downarrow$ & $\text{ECE} \downarrow$ & $\text{OE} \downarrow$ & $\text{ECE} \downarrow$ & $\text{OE} \downarrow$ & $\text{ECE} \downarrow$ & $\text{OE} \downarrow$ & $\text{ECE} \downarrow$ & $\text{OE} \downarrow$ \\
\midrule
\multirow{18}{*}{\parbox{1.0cm}{SOD \\ Methods}} 
& MSRNet \cite{li2017instance} & 2017 & 2.57 & 2.34 & 3.32 & 3.16 & 3.44 & 3.23 & 6.42 & 6.14 & 0.97 & 0.94 & 0.92 & 0.87\\
& SRM \cite{Wang_2017_ICCV} & 2017 & 4.02 & 3.72 & 4.19 & 3.96 & 4.88 & 4.59 & 9.93 & 9.58 & 2.53 & 2.35 & 1.86 & 1.72\\
& Amulet \cite{zhang2017amulet} & 2017 & 5.67 & 5.28 & 5.84 & 5.49 & 5.76 & 5.43 & 10.03 & 9.59 & 2.56 & 2.42 & 1.98 & 1.87\\
& BMPM \cite{zhang2018bi} & 2018 & 3.74 & 3.52 & 4.52 & 4.37 & 4.88 & 4.68 & 8.16 & 7.93 & 1.95 & 1.89 & 1.58 & 1.53 \\
& DGRL \cite{wang2018detect} & 2018 & 4.12 & 3.86 & 4.41 & 4.21 & 5.01 & 4.77 & 8.44 & 8.20 & 2.13 & 2.02 & 1.63 & 1.53\\
& PAGR \cite{zhang2018progressive} & 2018 & 4.04 & 3.79 & 5.14 & 4.96 & 5.64 & 5.37 & 12.17 & 11.87 & 2.84 & 2.70 & 1.62 & 1.54\\
& PiCANet \cite{liu2018picanet} & 2018 & 5.12 & 4.90 & 4.84 & 4.70 & 8.14 & 7.92 & 10.50 & 10.30 & 3.48 & 3.39 & 2.55 & 2.47\\
& CPD \cite{wu2019cascaded} & 2019 & 3.97 & 3.78 & 4.20 & 4.06 & 5.37 & 5.17 & 9.65 & 9.39 & 2.29 & 2.19 & 1.99 & 1.90\\
& BASNet \cite{qin2019basnet} & 2019 & 5.00 & 4.86 & 4.93 & 4.83 & 6.50 & 6.36 & 10.40 & 10.27 & 2.74 & 2.70 & 2.30 & 2.26\\
& EGNet \cite{zhao2019egnet} & 2019 & 3.33 & 3.14 & 3.66 & 3.50 & 5.42 & 5.19 & 8.04 & 7.79 & 1.98 & 1.88 & 1.47 & 1.40\\
& AFNet \cite{feng2019attentive} & 2019 & 3.95 & 3.74 & 4.25 & 4.09 & 5.06 & 4.84 & 8.15 & 8.02 & 2.38 & 2.27 & 1.87 & 1.78\\
& PoolNet \cite{Liu21PamiPoolNet} & 2019 & 3.33 & 3.12 & 3.86 & 3.70  & 5.32 & 5.07 & 8.14 & 7.87 & 2.00 & 1.90 & 1.82 & 1.75\\
& GCPANet \cite{chen2020global} & 2020 & 3.18 & 2.99 & 3.99 & 3.84 & 4.16 & 3.97 & 7.05 & 6.88 & 1.61 & 1.54 & 1.27 & 1.21\\
& MINet \cite{pang2020multi} & 2020 & 3.65 & 3.48 & 4.45 & 4.29 & 4.94 & 4.75 & 8.01 & 7.89 & 2.13 & 2.03 & 1.74 & 1.65 \\
& $\text{F}^{3}\text{Met}$ \cite{wei2020f3net} & 2020 & 3.67 & 3.50 & 4.25 & 4.10 & 4.85 & 4.67 & 7.95 & 7.78 & 2.26 & 2.16 & 1.92 & 1.83 \\
& EBMGSOD \cite{jing_ebm_sod21} & 2021 & 3.45 & 3.29 & 4.11 & 3.95 & 4.79 & 4.61 & 7.48 & 7.30 & 2.14 & 2.05 & 1.79 & 1.70\\
& ICON \cite{zhuge2021salient} & 2021 & 2.89 & 2.76 & 3.84 & 3.71 & 4.08 & 3.95 & 6.70 & 6.55 & 1.56 & 1.49 & 1.38 & 1.32\\
& PFSNet \cite{ma2021pyramidal} & 2021 & 2.94 & 2.72 & 3.95 & 3.81 & 4.45 & 4.27 & 7.59 & 7.39 & 2.41 & 2.25 & 2.06 & 1.96\\
& EDN \cite{wu2022edn} & 2022 & 3.62 & 3.47 & 4.02 & 3.90 & 4.89 & 4.74 & 8.81 & 8.66 & 2.20 & 2.13 & 1.65 & 1.58\\
\hline
\multirow{7}{*}{\parbox{1.0cm}{Model \\ Calibration \\ Methods}}
& Brier Loss \cite{brier1950verification} & 1950 & 2.77 & 2.58 & 3.55 & 3.38 & 3.90 & 3.70 & 6.40 & 6.16 & 1.37 & 1.30 & 1.04 & 0.99 \\
& Temperature Scaling \cite{guo2017calibration} & 2017 & 2.53 & 2.34 & 3.18 & 3.03 & 3.56 & 3.36 & 6.32 & 6.05 & 0.96 & 0.93 & 0.83 & 0.70\\
& MMCE \cite{MMCE} & 2018 & 2.86 & 2.67 & 3.56 & 3.41 & 4.00 & 3.81 & 6.85 & 6.63 & 1.41 & 1.35 & 1.18 & 1.13\\
& Label Smoothing \cite{muller2019does} & 2019 & 2.00 & 1.79 & 2.89 & 2.71 & 3.04 & 2.83 & 5.97 & 5.69 & {\color{blue} 0.83} & 0.68 & {\color{blue} 0.82} & 0.47\\
& Mixup \cite{thulasidasan2019mixup} & 2019 & 2.45 & 2.25 & 3.41 & 3.23 & 3.13 & 2.99 & {\color{blue} 5.82} & 5.70 & 1.41 & {\color{blue} 0.18} & 3.83 & 0.05\\
& Focal Loss \cite{focal_loss} & 2020 & 2.25 & 2.08 & 3.10 & 2.82 & 3.40 & 3.13 & 6.21 & 5.98 & 1.41 & 1.03 & 1.24 & 0.77\\
& AdaFocal \cite{ghosh2022adafocal} & 2022 & {\color{blue}1.61} & 1.41 & {\color{blue}2.31} & 1.84 & {\color{blue}2.53} & 2.27 & 5.88 & 5.47 & 1.63 & 0.79 & 1.35 & 0.52\\
\hline
\multirow{2}{*}{\parbox{1.0cm}{Our \\ Methods}}
& $\text{ASLP}_{\text{MC}}$  & 2023 & {\color{red} \textbf{1.40}} & {\color{blue} 1.22} & {\color{red} \textbf{1.99}} & {\color{blue} 1.83} & {\color{red} \textbf{2.31}} & {\color{blue} 2.10} & {\color{red} \textbf{5.50}} & {\color{blue} 5.17} & {\color{red} \textbf{0.48}} &  0.20 & {\color{red} \textbf{0.79}} & {\color{blue} 0.17}\\
& $\text{ASLP}_{\text{MEI}}$ & 2023 & 27.9 & {\color{red} \textbf{0.01}} & 26.0 & {\color{red} \textbf{0.00}} & 26.1 & {\color{red} \textbf{0.00}} & 22.4 & {\color{red} \textbf{0.00}} & 29.9 & {\color{red} \textbf{0.00}} & 30.5 & {\color{red} \textbf{0.00}}\\
\bottomrule
\end{tabular}
\label{tab:SOD_Calibration_Degree_Benchmark}
\end{table*}

\noindent \textbf{Compared Methods:}
We compare with both SOD models and model calibration methods
in terms of model calibration degrees. The SOD models include: MSRNet \cite{li2017instance}, SRM \cite{Wang_2017_ICCV}, Amulet \cite{zhang2017amulet}, BMPM \cite{zhang2018bi}, DGRL \cite{wang2018detect}, PAGR \cite{zhang2018progressive}, PiCANet \cite{liu2018picanet}, CPD \cite{wu2019cascaded}, BASNet \cite{qin2019basnet}, EGNet \cite{zhao2019egnet}, AFNet \cite{feng2019attentive}, PoolNet \cite{Liu21PamiPoolNet}, GCPANet \cite{chen2020global}, MINet \cite{pang2020multi}, $\text{F}^{3}\text{Met}$ \cite{wei2020f3net}, EBMGSOD \cite{jing_ebm_sod21}, ICON \cite{zhuge2021salient}, EDN \cite{wu2022edn}. We evaluate ECE on their published prediction results, or results produced with their released model weights. We also compare with model calibration methods include: Temperature Scaling (TS) \cite{guo2017calibration}, Brier Loss \cite{brier1950verification}, MMCE \cite{MMCE}, Label Smoothing \cite{muller2019does}, Mixup \cite{thulasidasan2019mixup}, Focal Loss \cite{focal_loss} and AdaFocal \cite{ghosh2022adafocal} implemented on our baseline model.

\noindent\textbf{Baseline Model Structure:}
We implement our method in the Pytorch framework. Our model has a simple U-Net structure, comprising of a ResNet50 encoder \cite{he2016deep} and a decoder, where the former is initialised with ImageNet-pretrained weights and the later by default. We also experiment with VGG16 \cite{simonyan2014very} and Swin transformer \cite{liu2021swin} encoders and report their results in Appendix~\ref{A_sec:Experiments_with_Additional_Backbones}.

\noindent\textbf{Label Perturbation Techniques:}
We experiment with four different label perturbation strategies: (1) Hard Inversion $T(Y, \alpha, \beta=1.0)$, (2) Soft Inversion (SI) $T(Y, \alpha, \beta=0.75)$, (3) Moderation (M) $T(Y, \alpha, \beta=0.5)$ and Dynamic Moderation (DM) $T(Y, \alpha, \beta=0.5)$ with additional 
Gaussian noise. See Appendix~\ref{A_sub_sec:SLP_Implementation} for implementation details.

\noindent\textbf{Training Details:} 
Each model is trained with an Adam optimiser for 30 epochs. The learning rate is initialised to $2.5 \times 10^{-5}$, and decays by a factor of 0.9 for each epoch after the $10^{\text{th}}$ epoch. All training images are scaled to $384 \times 384$. Basic data augmentation techniques, including random flipping, random translation and random cropping, are applied.

\noindent\textbf{Hyperparameters:}
The reported model calibration results associated with ASLP are obtained by setting $\eta = 0.002$ and $\lambda = 2,000$. 
We set the number of bins to $B = 10$ for ECE and OE evaluation metrics. 

\begin{figure*}
    \centering
    \includegraphics[width=0.9\textwidth]{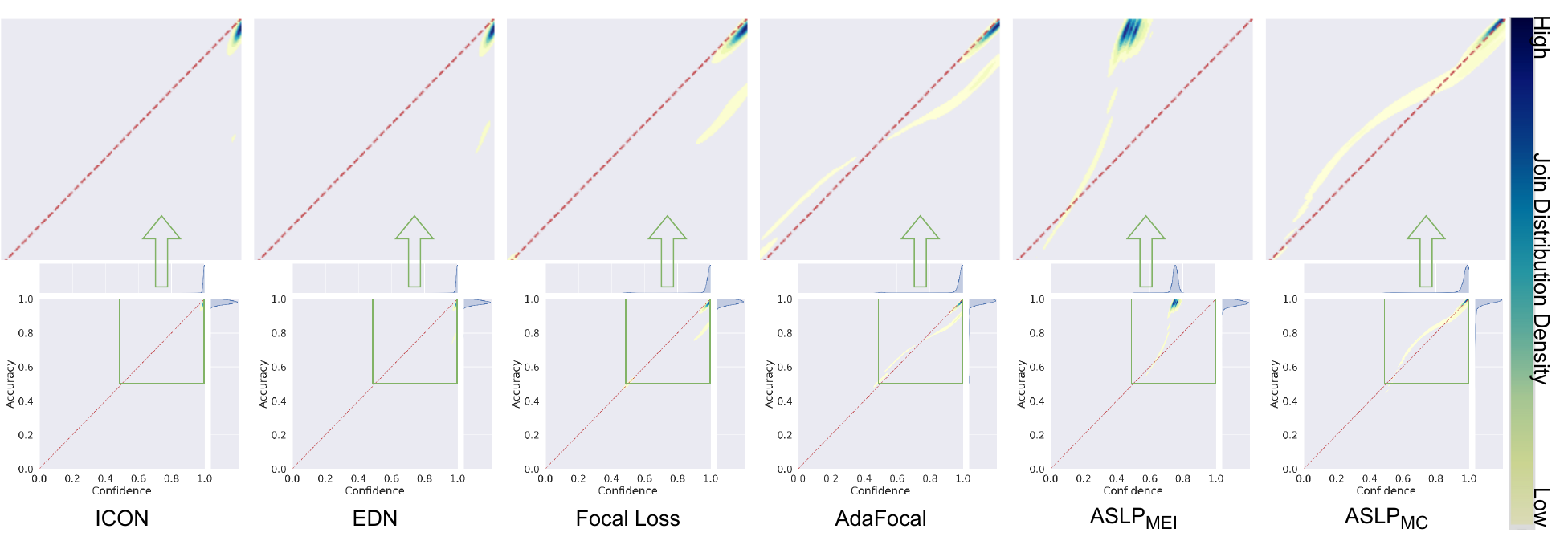}
    \vspace{-8pt}
    \caption{Joint distribution of prediction confidence (horizontal axis) and prediction accuracy (vertical axis) on the DUTS-TE dataset. A perfectly calibrated model has an identical confidence distribution and accuracy distribution, denoted as the oracle ({\color{red}diagonal red dotted line}). The joint distribution of a better calibrated model is more aligned with the oracle line, especially its high density area.
    See Appendix~\ref{A_sec:More_Joint_Plot} for results of other methods and on other testing datasets.
    }
    \label{fig:joint_distribution_plot}
\end{figure*}

\subsection{Model Calibration Degree Performance}
Tab.~\ref{tab:SOD_Calibration_Degree_Benchmark} presents the calibration degree of existing SOD models, existing model calibration methods and our proposed technique on the six SOD testing datasets. Our proposed $\text{ASLP}_{\text{MC}}$, designed to optimise the model calibration degree, achieves the best ECE performances on all testing datasets. 
In addition, $\text{ASLP}_{\text{MC}}$ also obtains the second-best OE performances on all six testing datasets, outperformed only by our $\text{ASLP}_{\text{MEI}}$.
On the other hand, $\text{ASLP}_{\text{MEI}}$, though almost eliminates the over-confidence issue completely, is significantly mis-calibrated on the six testing datasets. This can be attributed to it being significantly under-confident rather than over-confident. The observed performances of $\text{ASLP}_{\text{MEI}}$ are in accordance with its design - assuming minimum distribution commitment with respect to missing information. That is, in the presence of limited training data, to maximise the prediction entropy while maintaining the prediction accuracy for in-distribution data. 

Fig.~\ref{fig:joint_distribution_plot} presents the joint distribution of prediction confidence and prediction accuracy 
of some of the best calibrated methods and our proposed technique on DUTS-TE dataset 
(See Appendix~\ref{A_sec:More_Joint_Plot} for other testing datasets and other methods). Existing SOD methods produce extremely confident predictions whose confidence scores are nearly 100\% for the majority of samples. On the contrary, the prediction accuracy is on average lower than prediction confidence, resulting in the model being over-confident. On the other hand, existing model calibration methods are generally more calibrated than the SOD methods which in general do not strive to improve model calibration degree.

Our proposed $\text{ASLP}_{\text{MC}}$ produces the most calibrated model whose joint distribution is closer to the oracle than those of the compared calibration methods and SOD methods. AdaFocal \cite{ghosh2022adafocal} produces the second most calibrated model on DUTS-TE. However, the majority of their joint distribution ({\color{blue} blue} high density area) is slightly to the right bottom of the oracle line, making their model slightly less calibrated. Whereas the high density area of our joint distribution is well aligned with the oracle line, showing $\text{ASLP}_{\text{MC}}$ is more calibrated. Despite the small difference on the plot, $\text{ASLP}_{\text{MC}}$ improves over AdaFocal by more than 10\% in terms of ECE scores on DUTS-TE dataset.
We can also observe that $\text{ASLP}_{\text{MEI}}$ is significantly under-confident for in-distribution data with its joint distribution being at the top-left side of the oracle line. Its prediction confidences are limited to between 70\% and 80\% while the prediction accuracies are generally above 90\%.

\begin{figure}
    \centering
    \includegraphics[width=\linewidth]{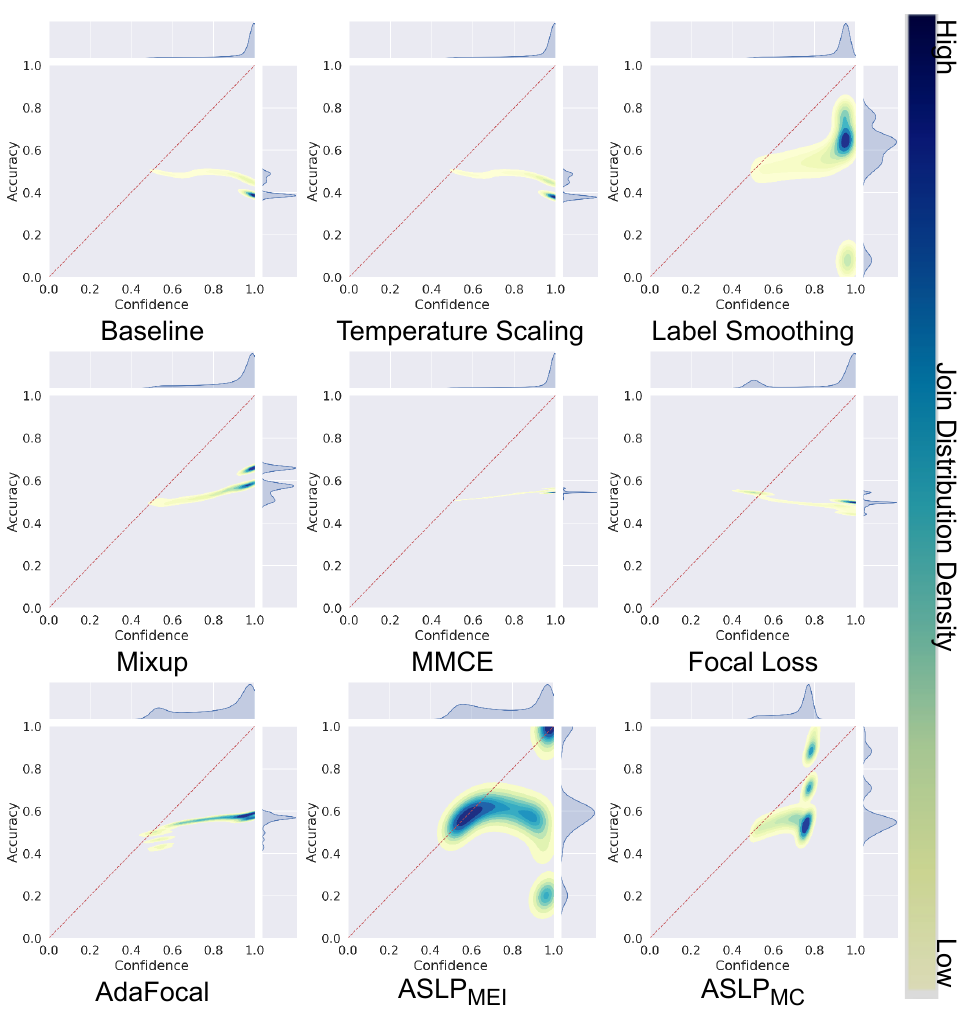}
    \vspace{-20pt}
    \caption{Joint distribution of prediction confidence (horizontal axis) and prediction accuracy (vertical axis) on the Describable Texture Dataset \cite{cimpoi2014describing}, of compared model calibration methods and our proposed $\text{ASLP}_{\text{MC}}$ and $\text{ASLP}_{\text{MEI}}$.
    }
    \label{fig:joint_plot_texture100}
    \vspace{-10pt}
\end{figure}

\subsection{Model Calibration Degrees on Out-of-Distribution Dataset}
We compare our proposed method with existing model calibration methods in terms of model calibration degrees on Out-of-Distribution data. We consider texture images, where salient objects are completely absent, as OoD samples for the SOD task. 
We use Describable Texture Dataset \cite{cimpoi2014describing} to evaluate the model calibration degrees on OoD samples.
Fig.~\ref{fig:joint_plot_texture100} shows the joint distribution of prediction confidence and prediction accuracy of various model calibration methods and our proposed techniques. It can be seen that the baseline model produces extremely confident predictions for OoD data. However, its accuracy is only 41.88\%, 
worse 
than a uniform prior in a binary classification task. We also observe that Temperature Scaling does not calibrate the model under data distribution shift in accordance with literature \cite{ovadia2019can,focal_loss}. Our $\text{ASLP}_{\text{MC}}$, being the most calibrated for in-distribution data, is also more calibrated on the OoD samples than the existing model calibration methods by a large margin as shown in Tab.~\ref{tab:TEXTURE100_quantitative}. On the other hand, our $\text{ASLP}_{\text{MEI}}$ is more successful in handling OoD data. It is the most calibrated on OoD data, with a larger proportion of the distribution aligned with the oracle line. As shown in Tab.~\ref{tab:TEXTURE100_quantitative}, it outperforms existing model calibration methods in terms of both ECE and OE by significant margins. This can be attributed to its minimum distribution assumption in the presence of limited training data.

\begin{table}[tbh!]
    \centering
    \scriptsize
    \renewcommand{\arraystretch}{1.2}
    \renewcommand{\tabcolsep}{1.0mm}
    \caption{Model calibration methods and our $\text{ASLP}_{\text{MC}}$ and $\text{ASLP}_{\text{MEI}}$ are evaluated on the Out-of-Distribution dataset, Describable Texture Dataset \cite{cimpoi2014describing}, in terms of 
    $\text{ECE}_{\text{EW}}$ and $\text{OE}_{\text{EW}}$ with 10 bins,
    and Accuracy (ACC).}
    \vspace{-8pt}
    \begin{tabular}{L{0.32\linewidth}|C{0.15\linewidth}C{0.15\linewidth}C{0.15\linewidth}}
    \toprule
    {\multirow{2}{*}{\parbox{3.6cm}{Method}}} & \multicolumn{3}{c}{Evaluation (\%)} \\
    & ECE $\downarrow$ & OE $\downarrow$ & ACC $\uparrow$ \\
    \midrule
    Baseline & 52.36 & 51.05 & 41.88\\
    \hline
    Brier Loss \cite{brier1950verification} & 38.85 & 37.18 & 53.62\\
    Temperature Scaling \cite{guo2017calibration} & 51.95 & 50.46 & 41.59\\
    Label Smoothing \cite{muller2019does} & 37.22 & 35.48 & 55.41\\
    MMCE \cite{MMCE} & 40.64 & 39.67 & 54.39\\
    Mixup \cite{thulasidasan2019mixup} & 31.07 & 29.10 & 58.71\\
    Focal Loss \cite{focal_loss} & 40.01 & 38.43 & 49.71\\
    AdaFocal \cite{ghosh2022adafocal} & 27.55 & 25.07 & 55.39\\
    \hline
    $\text{ASLP}_{\text{MC}}$ & 18.31 & 16.37 & 61.93\\
    $\text{ASLP}_{\text{MEI}}$ & \textbf{13.43} & \textbf{8.40} & \textbf{62.47}\\
    \bottomrule
    \end{tabular}
    \label{tab:TEXTURE100_quantitative}
    \vspace{-10pt}
\end{table}

\begin{table*}[htb!]
\centering
\scriptsize
\renewcommand{\arraystretch}{1.1}
\renewcommand{\tabcolsep}{1.5mm}
\caption{Ablation: Effect of Stochastic Label Perturbation (SLP) and Adaptive Stochastic Label Perturbation (ASLP) with different label perturbation techniques on the model calibration degrees evaluated on Expected Calibration Error (ECE) and Over-confidence Error (OE). The proposed ASLP is generalised to an Adaptive Label Smoothing (ALS) technique that adaptively tunes the label softening scale ($\beta_{\text{ada}}$).}
\vspace{-8pt}
\begin{tabular}{l|ccc|cc|cc|cc|cc|cc|cc}
\toprule
\multirow{2}{*}{Methods} & \multicolumn{3}{c|}{Perturbation Params} & \multicolumn{2}{c|}{DUTS-TE \cite{DUTS-TE}} & \multicolumn{2}{c|}{DUT-OMRON \cite{DUT-OMRON}} & \multicolumn{2}{c|}{PASCAL-S \cite{PASCAL-S}} & \multicolumn{2}{c|}{SOD \cite{SOD}} & \multicolumn{2}{c|}{ECSSD \cite{ECSSD}} & \multicolumn{2}{c}{HKU-IS \cite{HKU-IS}}\\
& $\alpha$ & $\beta$ & e & $\text{ECE} \downarrow$ & $\text{OE} \downarrow$ & $\text{ECE} \downarrow$ & $\text{OE} \downarrow$ & $\text{ECE} \downarrow$ & $\text{OE} \downarrow$ & $\text{ECE} \downarrow$ & $\text{OE} \downarrow$ & $\text{ECE} \downarrow$ & $\text{OE} \downarrow$ & $\text{ECE} \downarrow$ & $\text{OE} \downarrow$ \\
\midrule
Baseline (\enquote{B}) & 0 & 0 & 0 & 3.48 & 3.29 & 4.17 & 4.02 & 4.60 & 4.41 & 7.42 & 7.17 & 1.93 & 1.86 & 1.64 & 1.59\\
\hline
$\text{SLP}_{\text{HI}}^{\alpha=0.01}$ & 0.01 & 1.0 & \xmark & 2.21 & 1.84 & 2.96 & 2.78 & 3.11 & 2.82 & 6.09 & 5.80 & 1.03 & 0.68 & 1.01 & 0.53\\
$\text{SLP}_{\text{SI}}^{\alpha=0.02}$ & 0.02 & 0.75 & \xmark & 2.25 & 2.05 & 3.00 & 2.82 & 3.05 & 2.83 & 6.40 & 6.09 & 0.93 & 0.84 & 0.87 & 0.60\\
$\text{SLP}_{\text{M}}^{\alpha=0.03}$ & 0.03 & 0.5 & \xmark & 2.24 & 2.03 & 3.17 & 2.97 & 3.41 & 3.20 & 6.26 & 5.97 & 0.83 & 0.77 & 0.96 & 0.81 \\
$\text{SLP}_{\text{DM}}^{\alpha=0.03}$ & 0.03 & 0.5 & \cmark & 2.29 & 2.09 & 3.00 & 2.83 & 3.47 & 3.24 & 6.72 & 6.43 & 1.13 & 1.04 & 0.96 & 0.80\\
$\text{LS}^{\beta=0.03}$ & 1.0 & 0.03 & \xmark & 2.20 & 1.99 & 3.09 & 2.91 & 3.24 & 3.03 & 6.27 & 5.99 & 1.03 & 0.78 & 0.92 & 0.67\\
\hline
$\text{ASLP}_{\text{MC}}^{\text{HI}}$ & $\alpha_{\text{ada}}$ & 1.0 & \xmark & 1.40 & 1.22 & 1.99 & 1.83 & 2.31 & 2.10 & 5.50 & 5.17 & 0.48 &  0.20 & 0.79 & 0.17\\
$\text{ASLP}_{\text{MC}}^{\text{SI}}$ & $\alpha_{\text{ada}}$ & 0.75 & \xmark & 1.51 & 1.29 & 2.14 & 1.95 & 2.29 & 2.07 & 5.12 & 4.80 & 0.61 & 0.34 & 0.84 & 0.22\\
$\text{ASLP}_{\text{MC}}^{\text{M}}$ & $\alpha_{\text{ada}}$ & 0.5 & \xmark & 1.47 & 1.27 & 1.87 & 1.80 & 2.37 & 2.13 & 5.63 & 5.29 & 0.51 & 0.23 & 0.80 & 0.20\\
$\text{ASLP}_{\text{MC}}^{\text{M}}$ & $\alpha_{\text{ada}}$ & 0.5 & \cmark & 1.64 & 1.20 & 1.94 & 1.75 & 2.03 & 1.81 & 4.14 & 3.84 & 0.80 & 0.42 & 0.87 & 0.42 \\
$\text{ALS}$ & 1.0 & $\beta_{\text{ada}}$ & \xmark & 1.46 & 1.25 & 2.07 & 1.87 & 2.30 & 2.10 & 5.44 & 5.18 & 0.61 & 0.25 & 0.81 & 0.32\\
\bottomrule
\end{tabular}
\label{tab:Ablation_Study}
\vspace{-10pt}
\end{table*}

\subsection{Discussion}
\noindent\textbf{Adaptive Stochastic Label Perturbation:}
We study the effect of ASLP on ECE and OE and present the experimental results in Tab.~\ref{tab:Ablation_Study}. It shows that ASLPs significantly outperforms the baseline model, \enquote{B}, which does not adopt any model calibration measures. In addition, ASLPs also outperforms their static counterparts (SLPs) which use a single $\alpha$ for the entire dataset. This can be attributed to the approach modelling variance of noise with input image.


\noindent\textbf{Generalisation of Adaptive Label Smoothing:}
We generalise the proposed ASLP to label smoothing, developing an Adaptive Label Smoothing (ALS) that fixes the label perturbation probability to 100\%, akin to the label smoothing technique \cite{muller2019does}, and tunes a smoothing factor for each training sample. As shown in Tab.~\ref{tab:Ablation_Study}, ALS effectively reduces the ECE and OE scores over its static version $\text{LS}^{\beta=0.03}$, and achieves similar performances with $\text{ASLP}_{\text{MC}}$ approaches on the six testing datasets. It manifests that our proposed ASLP can be generalised onto other label perturbation techniques as a measure to calibrate the SOD models.

\noindent\textbf{Compatibility with SOTA SOD Models}
We retrain EBMGSOD \cite{jing_ebm_sod21}, ICON \cite{zhuge2021salient} and EDN \cite{wu2022edn} with the proposed $\text{ASLP}_{\text{MC}}$ and find significant improvements in terms of model calibration degrees without compromising their classification abilities (See Appendix~\ref{A_sec:Generalisation_to_SOD_Models}).

\noindent\textbf{Compatibility with Different Backbones:} We demonstrate that our proposed method is also compatible with VGG16 \cite{simonyan2014very} and Swin transformer \cite{liu2021swin} backbones. See Appendix~\ref{A_sec:Experiments_with_Additional_Backbones} for details.

\noindent\textbf{Effectiveness in Other Dense Binary Classification Tasks:} Experiments on Smoke Detection \cite{yan2022transmission} and Camouflaged Object Detection \cite{fan2020camouflaged} demonstrate that our method can be generalised to dense binary classification tasks to improve model calibration degrees. See Appendix~\ref{A_sec:Additional_Experiments_on_Dense_Binary_Classification_Tasks}.

\noindent\textbf{Generalisation to Multi-Class Segmentation task:} Experiments on Semantic Segmentation \cite{everingham2010pascal} demonstrate that our method can also be generalised to dense multi-class classification tasks. See Appendix~\ref{A_sec:Additional_Experiments_on_Semantic_Segmentation}.

\subsection{Hyperparameters}
\noindent\textbf{Static Stochastic Label Perturbation:} 
Tab.~\ref{tab:Ablation_Study} shows that, under a small label perturbation probability, the four label perturbation techniques can alleviate the model over-confidence issues of the baseline model, \enquote{B}, on the six testing datasets. They also achieve similar results to Label Smoothing \cite{muller2019does}, setting $\beta = 0.03$ and $\alpha = 1$. 
Each SLP has a wide range of effective label perturbation probabilities that improves model calibrations (See Appendix~\ref{A_sub_Sec:SLP_Model_Calibration} Tab.~\ref{A_tab:Static_SLP_Effective_Alpha_Range}), and these improvements do not sacrifice the model's classification performance (see Appendix~\ref{A_sub_sec:SLP_Classification_Performance}).
Larger values of the label perturbation probability eventually lead to increasing ECE scores as the model transitions to being under-confident (see Appendix~\ref{A_sub_Sec:SLP_Model_Calibration}).

\noindent\textbf{Updating Rate $\eta$:} 
$\text{ASLP}_{\text{MC}}$ models trained with $\eta \in [0.0002, 0.005]$ are generally stable, producing similar calibration degrees and classification performances. Values smaller than 0.001 require longer training and high values lead to sub-optimal results (See Appendix~\ref{A_sec:Hyperparameters}).

\noindent\textbf{Regularisation Strength $\lambda$:} 
$\lambda$ spanning between 500 and 10,000 are optimal. 
Very high values for $\lambda$ can lead to oscillation resulting in poor performance (See Appendix~\ref{A_sec:Hyperparameters}).


\section{Conclusion}
\label{sec:conclusion}
This work first introduces a Self-Calibrating Binary Cross Entropy loss that unifies label perturbation processes including stochastic approaches and label smoothing to improve model calibration 
while preserving classification accuracy. We further propose an Adaptive Stochastic Label Perturbation that learns a unique label perturbation level for individual training image. Following Maximum Entropy Inference, ASLP adopts classification / calibration as proxy for known data and maximises the prediction entropy with respect to missing data. The proposed $\text{ASLP}_{\text{MC}}$ improves model calibration degrees on both in-distribution samples and out-of-distribution samples, without negatively impacting classification performance. The approach can be easily applied to different models, which we demonstrate with several SOTA models. It is also demonstrated to be effective on a semantic segmentation task and other binary tasks.

\small{\vspace{.1in}\noindent\textbf{Acknowledgments.}\quad
This research was in-part supported by the ANU-Optus Bushfire Research Center of Excellence.}

{\small
\bibliographystyle{ieee_fullname}
\bibliography{egbib}
}

\clearpage
\onecolumn
\appendix
\etocdepthtag.toc{mtappendix}
\etocsettagdepth{mtchapter}{none}
\etocsettagdepth{mtappendix}{subsection}
\tableofcontents
\clearpage

\input{Appendix}

\end{document}

%% file: Appendix.tex
\section{Derivation}
\subsection{Self-Calibration Binary Cross Entropy (SC-BCE) Loss}
\label{A_sec:SC-BCE_loss_derivation}

We show that our SC-BCE loss is close to label smoothing in binary classification. Label smoothing, as defined in Eq.~\eqref{A_eq:label_smoothing}, is a typical data augmentation that softens the training supervision signals \cite{muller2019does,lukasik2020does,xu2020towards,zhang2021delving}.
\begin{equation}
    S(Y, \sigma) = \text{LS}(Y, \sigma) = (1 - \sigma) Y + \frac{\sigma}{K}, \quad \forall y \in Y.
    \label{A_eq:label_smoothing}
\end{equation}
where $\sigma$ is the label smoothing strength hyperparameter and $K$ is the number of classes, thus is set to $K = 2$ for a binary task. For image label pairs $X,Y\sim P$, the BCE loss with label smoothing takes the form:
\begin{equation}
\begin{aligned}
    \mathcal{L}_{\text{BCE}}(\theta, X, S(Y, \sigma)) =  \mathbb{E}_{x,y \in X,Y} \biggl[ - \Bigl((1 - \sigma)y + \frac{\sigma}{2} \Bigr) \log f_{\theta}(x) - \Bigl(1 - \bigl((1 - \sigma)y + \frac{\sigma}{2} \bigr)\Bigr) \log (1 - f_{\theta}(x) ) \biggr].
\end{aligned}
\end{equation}
\noindent

On the other hand, our proposed SC-BCE loss, taking expectation over the Bernoulli variable $Z_{t}(x, y)$, can be written as:
\begin{equation}
\begin{aligned}
\mathbb{E}_{Z_{t}} \biggl[ \mathcal{L}_{\text{SC-BCE}}(\theta, X,Y, \alpha, \beta) \biggr] = & \mathbb{E}_{Z_{t}} \biggl[ (1-Z_{t})\mathcal{L}_{\text{BCE}}(X,Y;\theta) + Z_{t} \mathcal{L}_{\text{BCE}}(X, P(Y, \beta), \theta) \biggr]\\
= & (1 - \alpha)\mathcal{L}_{\text{BCE}}(X,Y;\theta) + \alpha  \mathcal{L}_{\text{BCE}}(X, P(Y, \beta), \theta) \\
= & \mathbb{E}_{x,y \in X, Y} \Bigl[ - \bigl( (1 - \alpha)y + \alpha p \bigr) \log f_{\theta}(x) - \bigl( 1 - (1 - \alpha)y - \alpha p \bigr) \log (1 - f_{\theta}(x)) \Bigr]\\
\end{aligned}
\end{equation}
Substitute: $p(Y, \beta) = (1 - \beta) \cdot y + \frac{\beta}{2}$, then we have:
\begin{equation}
\begin{aligned}
& \mathbb{E}_{x,y \in X,Y} \Bigl[ - \bigl( (1 - \alpha)y + \alpha p \bigr) \log f_{\theta}(x) - \bigl( 1 - (1 - \alpha)y - \alpha p \bigr) \log (1 - f_{\theta}(x)) \Bigr]\\
= & \mathbb{E}_{x,y \in X, Y} \biggl[ - \Bigl( (1 - \alpha)y + \alpha \bigl( (1 - \beta)y + \frac{\beta}{2} \bigr) \Bigr) \log f_{\theta}(x) - \bigg( 1 - \Bigl( (1 - \alpha)y + \alpha \bigl( (1 - \beta)y + \frac{\beta}{2} \Bigr) \bigg) \log(1 - f_{\theta}(x)) \biggr]\\
= & \mathbb{E}_{x,y \in X, Y} \biggl[ - \Bigl((1 - \alpha \beta)y + \frac{\alpha \beta}{2} \Bigr) \log f_{\theta}(x) - \Bigl(1 - \bigl((1 - \alpha \beta)y + \frac{\alpha \beta}{2} \bigr)\Bigr) \log (1 - f_{\theta}(x) ) \biggr]\\
= & \mathcal{L}_{\text{bce}}(\theta, X, S(Y, \alpha \beta)),
\end{aligned}
\end{equation}
where we let $\alpha \beta = \sigma$ to show that the expectation of SC-BCE loss over with a stochastically perturbed label over a Bernoulli variable is equivalent to a BCE loss with a smoothed label.



\subsection{Connection between SC-BCE and Maximum Entropy Inference}
\label{A_sec:connection_between_SC-BCE_and_MEI}

We prove that the SC-BCE loss maximises prediction entropy as well as minimising cross entropy between the prediction distribution and groundtruth distribution. Given the SC-BCE loss written as:
\begin{equation}
\begin{aligned}
    \mathcal{L}_{\text{SC-BCE}}(\theta, X,Y,\alpha, \beta) = & (1-Z_{t}) \mathcal{L}_{\text{BCE}}(\theta, X, Y) + Z_{t} \mathcal{L}_{\text{BCE}}(\theta, X, P(Y, \beta))\\
    = & (1-Z_{t})\mathcal{L}_{\text{BCE}}(\theta, X,Y) + Z_{t} \Bigg[(1-\frac{\beta}{2})\mathcal{L}_{\text{BCE}}(\theta,X,Y) + \frac{\beta}{2}\mathcal{L}_{\text{BCE}}(\theta,X,P(Y, 2))\Bigg] \\
    = & (1 - \beta Z_{t})\mathcal{L}_{\text{BCE}}(\theta, X,Y) + \frac{\beta Z_{t}}{2}\big[\mathcal{L}_{\text{BCE}}(\theta, X, P(Y, 2)) + \mathcal{L}_{\text{BCE}}(\theta,X,Y)\big]
\end{aligned}
\label{A_eq:Expanded_SC-BCE_Loss}
\end{equation}
where the first term includes a regular BCE loss $\mathcal{L}_{\text{BCE}}(\theta,X,Y)$ with random weight $1 - \beta Z_{t}$ and $P(Y, 2)$ represents an inverted label. Aside from the coefficient $Z\beta/2$, the second term can be expanded as a simpler form without label $Y$ by collecting the Y terms:

\begin{equation}
\begin{aligned}
    \mathcal{L}_{\text{BCE}}(\theta,X,P(Y,2)) + \mathcal{L}_{\text{BCE}}(\theta,X,Y) = & - \mathbb{E}_{x,y \in X,Y}\Big[(1 - y)\log f_{\theta}(x) + y\log (1 - f_{\theta}(x))\Big]\\
    & - \mathbb{E}_{x,y \in X, Y}\Big[y\log f_{\theta}(x) + (1 - y)\log (1 - f_{\theta}(x))\Big] \\
    = & - \mathbb{E}_{x \in X} \Big[\log f_{\theta}(x) + \log (1 - f_{\theta}(x))\Big] \\
    = & 2 \cdot \mathbb{E}_{x \in X} \left[ - \frac{1}{2}\log f_{\theta}(X) - \frac{1}{2}\log (1 - f_{\theta}(X))\right] \\
    = & 2 \cdot \mathcal{L}_{\text{BCE}}(\theta, X, U)\\
\end{aligned}
\label{A_eq:Expanded_Second_Term_of_SC-BCE_Loss}
\end{equation}
where $U$ is a uniform binary categorical distribution.
Substituting Eq.~\eqref{A_eq:Expanded_Second_Term_of_SC-BCE_Loss} into Eq.~\ref{A_eq:Expanded_SC-BCE_Loss} yields:
\begin{equation}
    \mathcal{L}_{\text{SC-BCE}}(\theta,X,Y,\alpha, \beta) = (1 - \beta Z_{t}) \cdot \mathcal{L}_{\text{BCE}}(\theta,X,Y) + \beta Z_{t} \cdot \mathcal{L}_{\text{BCE}}(\theta, X, U)
\end{equation}

\subsection{Derivation of Grad-$\alpha$}
\label{A_sec:Grad_Alpha_Derivation}
\begin{figure}[h!]
    \centering
    \includegraphics[width=\textwidth]{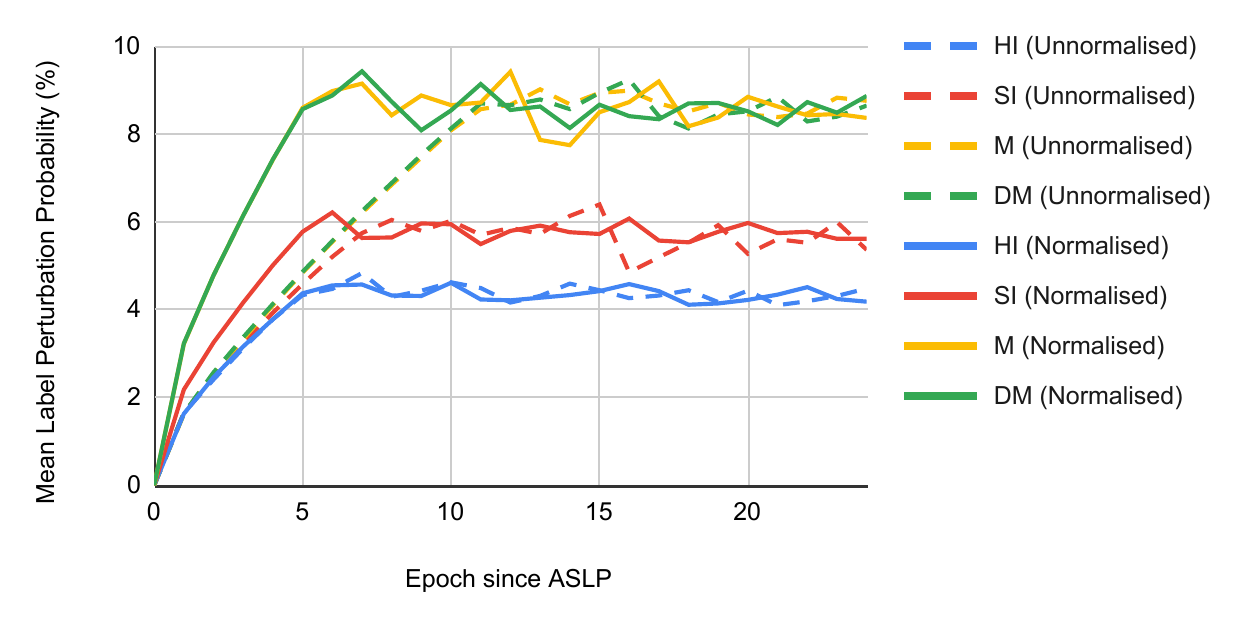}
    \caption{Convergence speed of unnormalised (dashed line) and normalised (solid line) Grad-$\alpha$ with different perturbation strengths: (1) HI: $\beta = 2$, (2) SI: $\beta = 1.5$, (3) M: $\beta = 1$, (4) DM: $\beta = 1$.}
    \label{fig:convergence_speed_of_grad_alpha}
\end{figure}

We start with the SC-BCE loss with sample-wise Bernoulli variable on a finite training dataset $\mathcal{D}_{\text{TR}} = \{x_{i}, y_{i}\}_{i=1}^{N}$ as:
\begin{equation}
    \mathcal{L}_{\text{SC-BCE}}(\theta, X, Y, \alpha, \beta) = \sum_{i=1}^{N} (1 -  Z_{t}(x_{i}, y_{i})) \cdot \mathcal{L}_{\text{BCE}}(\theta, x_{i}, y_{i}) + Z_{t}(x_{i}, y_{i}) \cdot \mathcal{L}_{\text{BCE}}(\theta, x_{i}, p(y_{i}, \beta)).
\end{equation}
where the variable is drawn from sample-specific Bernoulli distributions: $Z_{t}(x_{i}, y_{i}) \sim B(1, \alpha_{i}), \, i = 1, \dots, N$. Further, we take expectation over the Bernoulli variable for each individual training sample to recover:
\begin{equation}
\begin{aligned}
    & \sum_{i=1}^{N} \mathbb{E}_{Z_{t}(x_{i}, y_{i})} \Bigl[ (1 - Z_{t}(x_{i}, y_{i})) \cdot \mathcal{L}_{\text{BCE}}(\theta, x_{i}, y_{i}) + Z_{t}(x_{i}, y_{i}) \cdot \mathcal{L}_{\text{BCE}}(\theta, x_{i}, p(y_{i}, \beta)) \Bigr]\\
    =& \sum_{i=1}^{N} (1 - \alpha_{i}) \cdot  \mathcal{L}_{\text{BCE}}(\theta, x_{i}, y_{i}) + \alpha_{i} \cdot \mathcal{L}_{\text{BCE}}(\theta, x_{i}, p(y_{i}, \beta)).
\end{aligned}
\label{A_eq:Expected_SC-BCE}
\end{equation}
We further differentiate the above equation \wrt sample-specific label perturbation probability $\alpha_{i}, \, i = 1, \dots, N$ to obtain:
\begin{equation}
\begin{aligned}
    \frac{\partial \sum_{i=1}^{N} (1 - \alpha_{i}) \cdot  \mathcal{L}_{\text{BCE}}(\theta, x_{i}, y_{i}) + \alpha_{i} \cdot \mathcal{L}_{\text{BCE}}(\theta, x_{i}, p(y_{i}, \beta))}{\partial \alpha_{i}} =& - \mathcal{L}_{\text{BCE}}(\theta, x_{i}, y_{i}) + \mathcal{L}_{\text{BCE}}(\theta, x_{i}, p(y_{i}, \beta)),\\
    & \text{for} \quad i = 1, \dots, N,
\end{aligned}
\label{eq:unnormalised_Grad_alpha}
\end{equation} 
Performing gradient descent according to this gradient will lead to an optimal value for $\alpha$ with the regularization term.
We find Eq.~\ref{eq:unnormalised_Grad_alpha} (Unnormalised $\nabla_{\alpha_{i}}$) favours perturbation methods with higher perturbation strength $\beta$, leading them to to converge faster. 
This is because label perturbation techniques with higher strengths,  $\beta$, by definition have lower label perturbation probabilities, $\alpha$, overall to achieve optimal model calibration degrees whereas unnormalised Grad-$\alpha$ agnostic to label perturbation strength.
As illustrated in Fig.~\ref{fig:convergence_speed_of_grad_alpha}, with unnormalised Grad-$\alpha$, Hard Inversion (HI) with the largest perturbation strength $\beta = 2$ converges with only 5 epochs of ASLP training whereas it takes Moderation (M) and Dynamic Moderation (DM) with moderate perturbation strength ($\beta = 1$) around 11 epochs to converge. 

We propose a normalised version that allows ASLP under different perturbation strengths $\beta \in (0, 2]$ to converge equally fast. The unnormalised version (Eq.~\ref{eq:unnormalised_Grad_alpha}) is divided by $\beta / 2$ and the normalised $\nabla_{\alpha_{i}}$ is as:
\begin{equation}
    \nabla_{\alpha_{i}} = \frac{2 \cdot \Bigl(\mathcal{L}_{\text{BCE}}(\theta, x_{i}, y_{i}) + \mathcal{L}_{\text{BCE}}(\theta, x_{i}, p(y_{i}, \beta)) \Bigr)}{\beta}, \qquad i = 1, \dots, N
\end{equation}
Fig.~\ref{fig:convergence_speed_of_grad_alpha} illustrates that ASLP with different perturbation strengths with normalised $\nabla_{\alpha_{i}}$ can converge equally fast.

\subsection{Confidence of the Expectation of Stochastically Perturbed Label}
\label{A_sec:confidence_of_ecpectation_of_augmented_label}
We define the expectation of the stochastically perturbed label as:
\begin{equation}
    \mathbb{E}_{Z_{t}} \Bigl[ (1 - Z_{t}) \cdot Y + Z_{t} \cdot P(Y, \beta) \Bigr] = (1 - \alpha \beta) \cdot Y + \frac{\alpha \beta}{2},
\label{eq:expectation_of_stochastically_perturbed_label}
\end{equation}
where we require $\beta \in [0, 2]$ and $\alpha \in [0, \frac{1}{\beta})$. The resultant product is $\alpha \beta \in [0, 1)$. The expected confidence of perturbed label is:
\begin{equation}
\begin{aligned}
    C\Biggl( \mathbb{E}_{Z_{t}} \Bigl[ (1 - Z_{t}) \cdot Y + Z_{t} \cdot P(Y, \beta) \Bigr] \Biggr) =& \Bigl| (1 - \alpha \beta) \cdot Y + \frac{\alpha \beta}{2} - 0.5 \Bigr| + 0.5 \\
    = & 1 - \frac{\alpha \beta}{2}, \; \forall Y = \{0, 1\}
\end{aligned}
\end{equation}

\subsection{Adaptive Label Smoothing (ALS)}
\label{A_sec:Adaptive_Label_Smoothing}
Adaptive Label Smoothing (ASL) applies Label Smoothing with per-image label perturbation strength ($\alpha = 1$ and $\{\beta_{i}\}_{i=1}^{N}$). Similar to the derivation of $\nabla_{\alpha_{i}}$, we differentiate Eq.~\eqref{A_eq:Expected_SC-BCE} \wrt image-specific label perturbation strength as:
\begin{equation}
\begin{aligned}
    \nabla_{\beta_{i}} = &\frac{\partial \sum_{i=1}^{N} (1 - 1 \cdot \beta_{i}) \cdot  \mathcal{L}_{\text{BCE}}(\theta, x_{i}, y_{i}) + 1 \cdot \beta_{i} \cdot \mathcal{L}_{\text{BCE}}(\theta, x_{i}, p(y_{i}, \beta_{i}))}{\partial \beta_{i}}\\
    =& - \mathcal{L}_{\text{BCE}}(\theta, x_{i}, y_{i}) + \mathcal{L}_{\text{BCE}}(\theta, x_{i}, p(y_{i}, \beta_{i})), \qquad \text{for} \quad i = 1, \dots, N,
\end{aligned}
\label{eq:Unnormalised_Grad_Beta}
\end{equation}
The updating rule ($\text{ALS}_{\text{MC}}$) that incorporates adaptive label smoothing to maximise model calibration is formulated as:
\begin{equation}
\begin{aligned}
    \beta_{i}^{n + 1} 
    = \; & \beta_{i}^{n} + \eta \cdot \biggl( \underbrace{ \mathcal{L}_{\text{BCE}}(\theta, x_{i}, p(y_{i}, \beta_{i})) - \mathcal{L}_{\text{BCE}}(\theta, x_{i}, y_{i}) }_{\nabla_{\beta_{i}}} + \lambda \cdot \underbrace{\min \Bigl( \bigl(\mathrm{1} - \frac{1 \cdot \beta_{i}}{2} \bigr) - \mathbb{A}(\theta_{lm}, \mathcal{D}_{\text{VAL}}), 0\Bigr)}_{\text{Reg}_{\text{C}}} \biggr)\\
    &\text{for} \quad i = 1, \dots, N,
\end{aligned}
\end{equation}

\clearpage

\section{Implementations}
\subsection{Model}
\label{A_sub_sec:Model_Implementation}
Our model adopts a simple U-Net \cite{ronneberger2015u} structure consisting of an encoder and a decoder. Feature maps $\{F_{i} \in i \cdot C \times \frac{H}{i \cdot 8} \times \frac{W}{i \cdot 8}\}_{i=1}^{4}$ are extracted by the encoder, where $C = 256$ and $i$ indexes from low level to high level with an increasing value. 

The model outputs pixel-wise logits $\sigma(x_{i}) \in (- \infty, \infty)^{1 \times H \times W}, \; i = 1, \dots, N$ where $N$ is the total number of samples, which is further processed with a Sigmoid function to produce the prediction probability as:
\begin{equation}
\begin{aligned}
    f_{\theta}(x_{i}) &= \text{Sigmoid} (\sigma(x_{i})) = \frac{1}{1 + e^{- \sigma(x_{i})}},    \quad i = 1, \dots, N.
\end{aligned}
\end{equation}
The prediction probability after the Sigmoid function is in the range $f_{\theta}(x) \in (0, 1)^{1 \times H \times W}$. The predicted label is \enquote{foreground} (Labeled as \enquote{1}) if the prediction probability is larger than 0.5 and is \enquote{background} (labeled as \enquote{0}) otherwise as:
\begin{equation}
    \hat{y}_{i} = \mathbbm{1} (f_{\theta}(x_{i}) > 0.5), \quad i = 1, \dots, N.
\end{equation}
The probability of predicted label $\hat{y}$, also known as the winning class, is:
\begin{equation}
    P_{\hat{y}_{i}} = |f_{\theta}(x_{i}) - 0.5| + 0.5, \quad i = 1, \dots, N.
\end{equation}

\subsection{Evaluation Metrics - Model Calibration Degree}
\label{A_sec:Evaluation_Metrics_Model_Calibration_Degree}

\subsubsection{Equal-Width Expected Calibration Error ($\text{ECE}_{\text{EW}}$) \cite{guo2017calibration}}
\begin{equation}
    \text{ECE}_{\text{EW}} = \sum_{i=1}^{M} \frac{|B_{i}|}{\mathcal{|D|}} | C_{i} - A_{i} |,
\end{equation}
where $M$ is the total number of bins, $B_{i}$ and $\mathcal{D}$ denote the size of the $i^{\text{th}}$ bin and the dataset respectively, $C_{i} = \frac{1}{|B_{i}|} \sum_{j \in B_{i}} P_{\hat{y}_{j}}$ is the mean prediction confidence of the $i^{\text{th}}$ bin, and $A_{i} = \frac{1}{|B_{i}|} \sum_{j \in B_{i}} \mathbbm{1}(\hat{y}_{i} == y_{i})$ is the mean accuracy of the $i^{\text{th}}$ bin. $\text{ECE}_{\text{EW}}$ has fixed-width bins, with the range $\Bigl[ \frac{i}{M}, \frac{i + 1}{M} \Bigr), \; i = 0, \dots, M - 1$ for the $i^{\text{th}}$ bin.

\subsubsection{Equal-Mass Expected Calibration Error ($\text{ECE}_{\text{EM}}$) \cite{nguyen-oconnor-2015-posterior}}
\begin{equation}
    \text{ECE}_{\text{EW}} = \sum_{i=1}^{M} \frac{|B_{i}|}{\mathcal{|D|}} \cdot | C_{i} - A_{i} |, \qquad \text{where} \; |B_{j}| = |B_{k}|, \forall j, k \in [1, M].
\end{equation}
Equal-Mass Expected Calibration Error ($\text{ECE}_{\text{EM}}$) is different from Equal-Width Expected Calibration Error ($\text{ECE}_{\text{EW}}$) by constraining all bins to have equal size.

\subsubsection{SWEEP Expected Calibration Error ($\text{ECE}_{\text{SWEEP}}$) \cite{roelofs2022mitigating}}
\begin{equation}
    \text{ECE}_{\text{SWEEP}} = (\sum_{i = 1}^{b^{*}} \frac{|B_{i}|}{\mathcal{|D|}} | C_{i} - A_{i} \cdot |^{p})^{\frac{1}{p}}, \qquad \text{where} \; b^{*} = \max(b | 1 \leq b \leq n, \forall b' \leq b^{*}, A_{1} \leq \dots \leq A_{b'})
\end{equation}
$p$ is a hyperparameter that is set to $p = 1$ and $n$ is the largest bin number to be tested which we set to $n = 100$. $\text{ECE}_{\text{SWEEP}}$ follows $\text{ECE}_{\text{EM}}$ to constrain equal-size bins. $\text{ECE}_{\text{SWEEP}}$ starts with bin number $B = 1$ and keeps increasing the bin number until monotony in bin accuracy breaks.

\subsubsection{DEBIAS Expected Calibration Error ($\text{ECE}_{\text{DEBIAS}}$) \cite{kumar2019verified}}
\begin{equation}
    \text{ECE}_{\text{DEBIAS}} = \sum_{i=1}^{M} \frac{|B_{i}|}{\mathcal{|D|}} \Bigl[ (C_{i} - A_{i})^{2} - \frac{A_{i} \cdot (1 - A_{i})}{|B_{i}| - 1} \Bigr]
\end{equation}
DEBIAS Expected Calibration Error ($\text{ECE}_{\text{DEBIAS}}$) adopts equal-width bins.

\subsubsection{Over-confidence Error (OE)}
\begin{equation}
    \text{OE} = \sum_{i=1}^{M} \frac{|B_{i}|}{|\mathcal{D}|} \cdot \mathbbm{1}(C_{i} > A_{i}) \cdot | C_{i} - A_{i} |,
\end{equation}
We adapt OE to different binning schemes of $\text{ECE}_{\text{EW}}$, $\text{ECE}_{\text{EM}}$, $\text{ECE}_{\text{SWEEP}}$ to produce $\text{OE}_{\text{EW}}$, $\text{OE}_{\text{EM}}$, $\text{OE}_{\text{SWEEP}}$ respectively.

\subsection{Evaluation Metrics - Dense Classification}
\subsubsection{Prediction Accuracy}
The model prediction accuracy is computed as:
\begin{equation}
    \mathbb{A}(\theta, \mathcal{D}) = \frac{1}{N \times H \times W} \sum_{i = 1}^{N} \sum_{j = 1}^{H} \sum_{k = 1}^{W} \mathbbm{1} (\hat{y}_{i}^{j,k} = y_{i}^{j,k}),
\end{equation}
where $\mathcal{D} = \{x_{i}, y_{i}\}_{i=1}^{N}$ denotes the dataset with $N$ samples, $H$ and $W$ is the height and the width of sample respectively. 


\subsubsection{F-measure}
F-measure is computed as:
\begin{equation}
    F_{\xi} = \frac{(1 + \xi^{2}) \times \text{Precision} \times \text{Recall}}{\xi^{2} \times \text{Precision} + \text{Recall}},
\label{A_eq:F_measure}
\end{equation}
where $\xi$ is a hyperparameter. We follow previous works \cite{wu2022edn,liu2018picanet,zhang2017amulet,Liu21PamiPoolNet} to set $\xi^{2} = 0.3$. We report the maximum F-measure which selects the best results computed with various binarising threshold.

\subsubsection{E-measure}
Enhanced-alignment measure (E-measure) \cite{fan2018enhanced} is computed as:
\begin{equation}
\begin{aligned}
    Q_{FM} &= \frac{1}{H \times W} \sum_{i=1}^{H} \sum_{j=1}^{W} \phi_{FM}(i, j), \quad \text{where}\\
    \phi_{FM} &= f(\xi_{FM}) = \frac{1}{4}(1 + \xi_{FM})^{2},\\
    \xi_{FM} &= \frac{2 \cdot \varphi_{GT} \circ \varphi_{FM}}{\varphi_{GT} \circ \varphi_{GT} + \varphi_{FM} \circ \varphi_{FM}},\\
    \varphi_{I} &= I - \mu_{i} \cdot A,
\end{aligned}
\end{equation}
where $I \in (0, 1)$ is a dense binary prediction map with mean value $\mu_{I}$, $A$ is an one matrix whose dimension matches that of $I$, $\varphi_{GT}$ and $\varphi_{FM}$ denote groundtruth map and model prediction respectively, $H$ and $W$ is image height and width. Maximum E-measure replaces the mean value with a range of binarising thresholds and report the highest result.


\subsection{Datasets}
\noindent\textbf{DUTS-TR \cite{DUTS-TE}:} is commonly used training dataset for Salient Object Detection task. It consists of 10,553 pairs of image and pixel-wise annotations. We take a subset consisting 1,000 training samples as a validation set and uses the remaining 9,553 samples for training.

\noindent\textbf{DUTS-TE \cite{DUTS-TE}:} is a testing dataset consisting of 5,019 images. Both DUTS-TE and DUTS-TR belong to the DUTS dataset.

\noindent\textbf{DUT-OMRON \cite{DUT-OMRON}:} consists of 5,168 testing images, each of which includes at least one structurally complex foreground object(s).

\noindent\textbf{PASCAL-S \cite{PASCAL-S}:} contains 850 testing samples that are obtained from PASCAL-VOC dataset, which is designed for semantic segmentation task.

\noindent\textbf{SOD \cite{SOD}:} includes 300 testing images of a wide variety of natural scenes.

\noindent\textbf{ECSSD \cite{ECSSD}: } has 1,000 semantically meaningful images for testing.

\noindent\textbf{HKU-IS \cite{HKU-IS}:} is comprised of 4,447 testing images, each having multiple foreground objects.

\begin{figure}[hb!]
    \centering
    \includegraphics[width=0.9\textwidth]{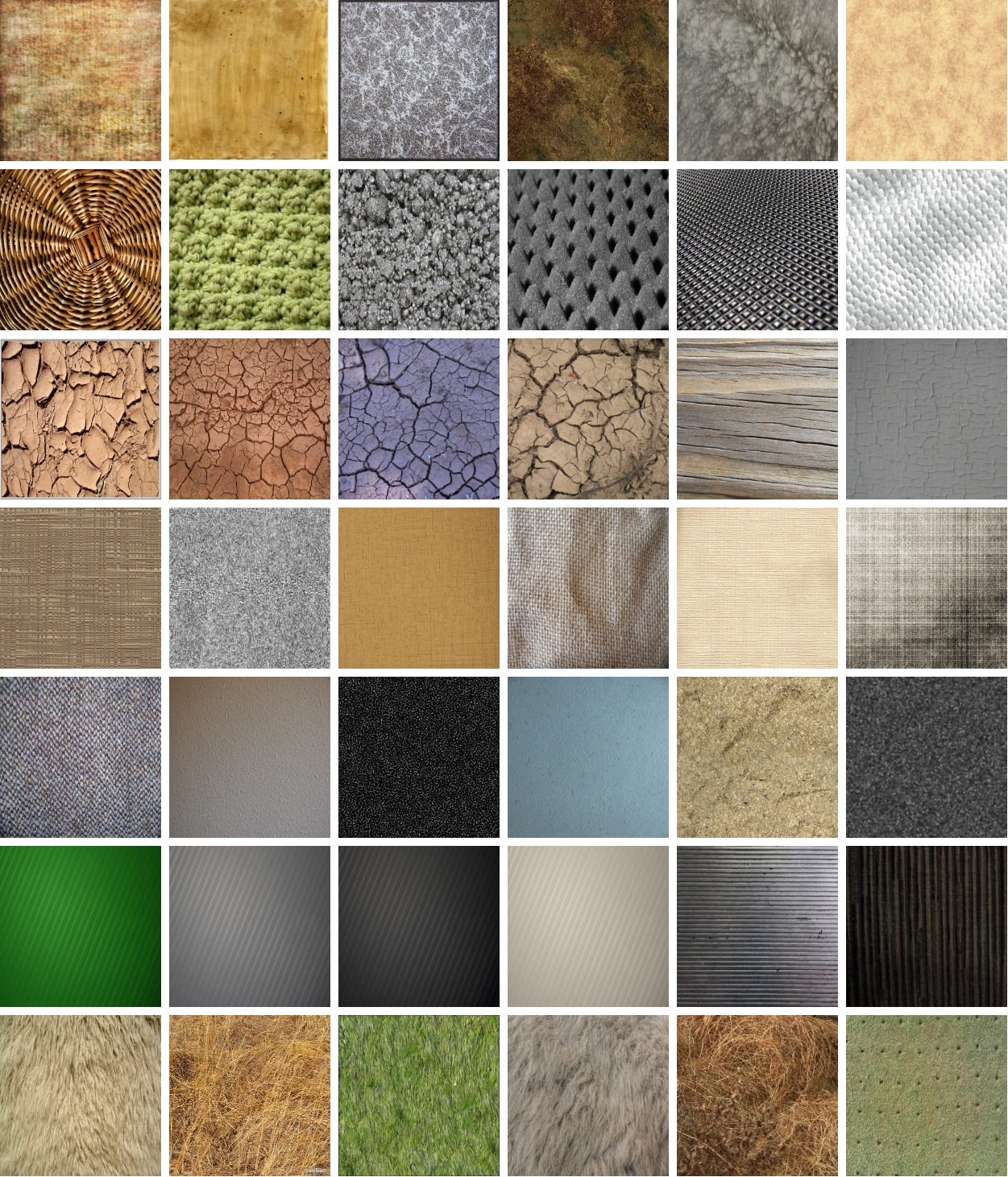}
    \caption{Texture image samples from Describable Texture Dataset \cite{cimpoi2014describing}.}
    \label{fig:DTD_Samples}
\end{figure}

\noindent\textbf{Describable Texture Dataset (DTD) \cite{cimpoi2014describing}:} contains 5,640 real-world texture images. These images are grouped into 47 categories described by adjectives such as \enquote{\textit{grooved}}, \enquote{\textit{woven}}, \enquote{\textit{matted}}. Some texture images have a distinct region that could be considered to be salient. We selectively choose only 500 texture images that have no obvious salient object 
and show some examples in Fig.~\ref{fig:DTD_Samples}. We consider the selected texture images an Out-of-Distribution samples for salient object detection. The complete collection of the 500 selected texture images are presented in Fig.~\ref{A_fig:Full_Texture_Image_Collection} at the end of the Appendix.

\clearpage



\section{Model Calibration Benchmark with $\text{ECE}_{\text{EM}}$, $\text{ECE}_{\text{SWEEP}}$ and $\text{ECE}_{\text{DEBIAS}}$}
\label{A_sec:model_calibration_benchmark_with_ECE_EM_SWEEP_DEBIAS}
We present the model calibration degrees of existing SOD methods, model calibration methods and our proposed methods evaluated in terms of: (i) Equal-Mass Expected Calibration Error $\text{ECE}_{\text{EM}}$ and Equal-Mass Over-confidence Error $\text{OE}_{\text{EM}}$ in Tab.~\ref{A_tab:ECE_EM_Benchmark}, (ii) $\text{ECE}_{\text{SWEEP}}$ and $\text{OE}_{\text{EM}}$ in Tab.~\ref{A_tab:ECE_SWEEP_Benchmark}, and (iii) $\text{ECE}_{\text{DEBIAS}}$ in Tab.~\ref{A_tab:ECE_DEBIAS_Benchmark}. Our proposed method, $\text{ASLP}_{\text{MC}}$, still outperforms existing salient object detection and model calibration methods with these model calibration evaluation metrics.

\begin{table}[htb!]
\centering
\scriptsize
\renewcommand{\arraystretch}{1.2}
\renewcommand{\tabcolsep}{1.0mm}
\caption{Salient object detection model calibration degree benchmark evaluated with $\text{ECE}_{\text{EM}}$ (\%) and $\text{OE}_{\text{EM}}$ (\%). We set the number of bins to $B = 10$. (values are shown in \% and {\color{red} red} and {\color{blue} blue} indicate the best and the second-best performance respectively.)}
\begin{tabular}{cl|c|cc|cc|cc|cc|cc|cc}
\toprule
\multicolumn{2}{c|}{\multirow{2}{*}{Methods}} & {\multirow{2}{*}{Year}} & \multicolumn{2}{c|}{DUTS-TE \cite{DUTS-TE}} & \multicolumn{2}{c|}{DUT-OMRON \cite{DUT-OMRON}} & \multicolumn{2}{c|}{PASCAL-S \cite{PASCAL-S}} & \multicolumn{2}{c|}{SOD \cite{SOD}} & \multicolumn{2}{c|}{ECSSD \cite{ECSSD}} & \multicolumn{2}{c}{HKU-IS \cite{HKU-IS}}\\
& & & $\text{ECE}_{\text{EM}} \downarrow$ & $\text{OE}_{\text{EM}} \downarrow$ & $\text{ECE}_{\text{EM}} \downarrow$ & $\text{OE}_{\text{EM}} \downarrow$ & $\text{ECE}_{\text{EM}} \downarrow$ & $\text{OE}_{\text{EM}} \downarrow$ & $\text{ECE}_{\text{EM}} \downarrow$ & $\text{OE}_{\text{EM}} \downarrow$ & $\text{ECE}_{\text{EM}} \downarrow$ & $\text{OE}_{\text{EM}} \downarrow$ & $\text{ECE}_{\text{EM}} \downarrow$ & $\text{OE}_{\text{EM}} \downarrow$ \\
\midrule
\multirow{18}{*}{\parbox{1.0cm}{SOD \\ Methods}} 
& MSRNet \cite{li2017instance} & 2017 & 3.35 & 3.03 & 3.64 & 3.40 & 4.23 & 3.93 & 5.52 & 5.13 & 1.12 & 1.08 & 1.05 & 0.96\\
& SRM \cite{Wang_2017_ICCV} & 2017 & 4.45 & 4.05 & 4.10 & 3.78 & 4.92 & 4.53 & 7.69 & 7.22 & 2.81 & 2.57 & 2.20 & 2.00\\
& Amulet \cite{zhang2017amulet} & 2017 & 5.63 & 5.10 & 5.46 & 4.98 & 5.69 & 5.23 & 8.24 & 7.63 & 2.64 & 2.45 & 2.09 & 1.94\\
& BMPM \cite{zhang2018bi} & 2018 & 3.47 & 3.21 & 4.52 & 4.18 & 4.77 & 4.57 & 8.00 & 7.88 & 1.89 & 1.83 & 1.55 & 1.50 \\
& DGRL \cite{wang2018detect} & 2018 & 4.42 & 4.04 & 3.87 & 3.57 & 4.91 & 4.57 & 5.69 & 5.35 & 2.23 & 2.07 & 1.69 & 1.53\\
& PAGR \cite{zhang2018progressive} & 2018 & 4.00 & 3.63 & 3.28 & 3.00 & 5.06 & 4.67 & 7.60 & 7.14 & 2.49 & 2.29 & 1.40 & 1.25\\
& PiCANet \cite{liu2018picanet} & 2018 & 5.37 & 4.98 & 5.17 & 4.82 & 5.78 & 5.41 & 8.75 & 8.33 & 2.67 & 2.44 & 2.31 & 2.09\\
& CPD \cite{wu2019cascaded} & 2019 & 3.10 & 2.90 & 3.62 & 3.42 & 4.11 & 3.86 & 6.76 & 6.42 & 2.07 & 1.94 & 1.81 & 1.70\\
& BASNet \cite{qin2019basnet} & 2019 & 6.07 & 5.85 & 6.15 & 5.96 & 5.72 & 5.48 & 5.07 & 4.88 & 2.12 & 2.04 & 2.36 & 2.28\\
& EGNet \cite{zhao2019egnet} & 2019 & 3.54 & 3.29 & 3.55 & 3.33 & 4.92 & 4.61 & 6.42 & 6.07 & 1.96 & 1.84 & 1.64 & 1.55\\
& AFNet \cite{feng2019attentive} & 2019 & 3.58 & 3.33 & 3.02 & 2.81 & 4.08 & 3.79 & 6.65 & 6.14 & 2.19 & 2.04 & 1.78 & 1.66\\
& PoolNet \cite{Liu21PamiPoolNet} & 2019 & 3.80 & 3.52 & 3.53 & 3.30  & 5.44 & 5.09 & 6.87 & 6.49 & 2.18 & 2.04 & 1.61 & 1.52\\
& GCPANet \cite{chen2020global} & 2020 & 4.40 & 4.12 & 4.84 & 4.61 & 4.92 & 4.64 & 4.20 & 3.94 & 1.87 & 1.76 & 1.54 & 1.47\\
& MINet \cite{pang2020multi} & 2020 & 5.02 & 4.76 & 5.40 & 5.13 & 6.17 & 5.86 & 8.29 & 8.01 & 2.84 & 2.67 & 2.31 & 2.17 \\
& $\text{F}^{3}\text{Met}$ \cite{wei2020f3net} & 2020 & 3.47 & 3.26 & 3.88 & 3.68 & 4.56 & 4.32 & 7.34 & 6.95 & 2.45 & 2.31 & 1.91 & 1.80 \\
& EBMGSOD \cite{jing_ebm_sod21} & 2021 & 3.64 & 3.41 & 3.78 & 3.55 & 4.79 & 4.52 & 5.83 & 5.56 & 2.30 & 2.15 & 1.85 & 1.72\\
& ICON \cite{zhuge2021salient} & 2021 & 2.40 & 2.26 & 2.95 & 2.81 & 3.45 & 3.29 & 4.27 & 4.09 & 1.34 & 1.25 & 1.23 & 1.16\\
& PFSNet \cite{ma2021pyramidal} & 2021 & 3.07 & 2.84 & 3.44 & 3.16 & 4.99 & 4.64 & 5.82 & 5.48 & 2.43 & 2.17 & 2.87 & 2.70\\
& EDN \cite{wu2022edn} & 2022 & 3.89 & 3.68 & 4.35 & 4.18 & 4.62 & 4.41 & 4.02 & 3.85 & 1.60 & 1.52 & 1.34 & 1.26\\
\hline
\multirow{7}{*}{\parbox{1.0cm}{Model \\ Calibration \\ Methods}}
& Brier Loss \cite{brier1950verification} & 1950 & 2.78 & 2.61 & 3.55 & 3.40 & 3.90 & 3.72 & 6.40 & 6.18 & 1.34 & 1.31 & 1.04 & 1.00 \\
& TS \cite{guo2017calibration} & 2017 & 2.77 & 2.60 & 3.44 & 3.30 & 3.85 & 3.67 & 6.64 & 6.40 & 1.21 & 1.17 & 0.95 & 0.91\\
& MMCE \cite{MMCE} & 2018 & 2.86 & 2.69 & 3.56 & 3.42 & 4.07 & 3.89 & 6.85 & 6.63 & \color{blue}{1.41} & 1.35 & 1.18 & 1.13\\
& LS \cite{muller2019does} & 2019 & 2.74 & 2.10 & 3.51 & 2.81 & 3.97 & 3.35 & 4.50 & 4.10 & 1.50 & 0.99 & 1.44 & 0.84\\
& Mixup \cite{thulasidasan2019mixup} & 2019 & 3.00 & 2.73 & 3.40 & 3.13 & \color{blue}{2.14} & 0.59 & 4.94 & 4.62 & 1.86 & 0.45 & 4.94 & 0.20\\
& Focal Loss \cite{focal_loss} & 2020 & 2.15 & 2.03 & 2.69 & 2.38 & 2.95 & 2.70 & 4.61 & 4.38 & 1.57 & 1.16 & 1.29 & 0.87\\
& AdaFocal \cite{ghosh2022adafocal} & 2022 & \color{blue}{1.74} & 1.50 & \color{blue}{1.96} & 1.45 & 2.45 & 2.02 & \color{blue}{3.88} & 3.09 & 1.79 & 0.74 & 1.45 & 0.44\\
\hline
\multirow{2}{*}{\parbox{1.0cm}{Our \\ Methods}}
& $\text{ASLP}_{\text{ECE}}$  & 2023 & \color{red}{\textbf{1.53}} & \color{blue}{1.41} & \color{red}{\textbf{1.72}} & \color{blue}{1.43} & \color{red}{\textbf{1.58}} & \color{blue}{1.55} & \color{red}{\textbf{2.30}} & \color{blue}{1.66} & \color{red}{\textbf{0.71}} & \color{blue}{0.35} & \color{red}{\textbf{0.84}} & \color{blue}{0.19}\\
& $\text{ASLP}_{\text{MEI}}$ & 2023 & 21.00 & \color{red}{\textbf{0.08}} & 20.24 & \color{red}{\textbf{0.00}} & 19.89 & \color{red}{\textbf{0.00}} & 18.14 & \color{red}{\textbf{0.00}} & 22.15 & \color{red}{\textbf{0.00}} & 22.58 & \color{red}{\textbf{0.00}}\\
\bottomrule
\end{tabular}
\label{A_tab:ECE_EM_Benchmark}
\end{table}

\begin{table}[htb!]
\centering
\scriptsize
\renewcommand{\arraystretch}{1.2}
\renewcommand{\tabcolsep}{1.0mm}
\caption{Salient object detection model calibration degree benchmark evaluated with $\text{ECE}_{\text{SWEEP}}$ (\%) and $\text{OE}_{\text{SWEEP}}$ (\%). The number of bins for each evaluation is selected to ensure a monotonically increasing accuracy in the bins \cite{roelofs2022mitigating} (values are shown in \% and {\color{red} red} and {\color{blue} blue} indicate the best and the second-best performance respectively.)}
\begin{tabular}{cl|c|cc|cc|cc|cc|cc|cc}
\toprule
\multicolumn{2}{c|}{\multirow{2}{*}{Methods}} & {\multirow{2}{*}{Year}} & \multicolumn{2}{c|}{DUTS-TE \cite{DUTS-TE}} & \multicolumn{2}{c|}{DUT-OMRON \cite{DUT-OMRON}} & \multicolumn{2}{c|}{PASCAL-S \cite{PASCAL-S}} & \multicolumn{2}{c|}{SOD \cite{SOD}} & \multicolumn{2}{c|}{ECSSD \cite{ECSSD}} & \multicolumn{2}{c}{HKU-IS \cite{HKU-IS}}\\
& & & $\text{ECE}_{\text{SW}} \downarrow$ & $\text{OE}_{\text{SW}} \downarrow$ & $\text{ECE}_{\text{SW}} \downarrow$ & $\text{OE}_{\text{SW}} \downarrow$ & $\text{ECE}_{\text{SW}} \downarrow$ & $\text{OE}_{\text{SW}} \downarrow$ & $\text{ECE}_{\text{SW}} \downarrow$ & $\text{OE}_{\text{SW}} \downarrow$ & $\text{ECE}_{\text{SW}} \downarrow$ & $\text{OE}_{\text{SW}} \downarrow$ & $\text{ECE}_{\text{SW}} \downarrow$ & $\text{OE}_{\text{SW}} \downarrow$ \\
\midrule
\multirow{18}{*}{\parbox{1.0cm}{SOD \\ Methods}} 
& MSRNet \cite{li2017instance} & 2017 & 3.16 & 2.85 & 4.10 & 3.86 & 4.09 & 3.85 & 5.30 & 5.05 & 1.04 & 1.00 & 1.01 & 0.94\\
& SRM \cite{Wang_2017_ICCV} & 2017 & 4.66 & 4.32 & 4.92 & 4.61 & 5.77 & 5.43 & 8.04 & 7.56 & 2.98 & 2.74 & 2.12 & 1.95\\
& Amulet \cite{zhang2017amulet} & 2017 & 6.52 & 6.04 & 7.31 & 6.85 & 6.50 & 6.08 & 8.47 & 7.88 & 2.17 & 2.06 & 2.47 & 2.32\\
& BMPM \cite{zhang2018bi} & 2018 & 4.77 & 4.38 & 4.27 & 3.98 & 6.13 & 5.74 & 8.74 & 8.31 & 2.09 & 1.72 & 2.03 & 1.85 \\
& DGRL \cite{wang2018detect} & 2018 & 4.51 & 4.30 & 3.98 & 3.81 & 4.61 & 4.46 & 5.23 & 4.89 & 1.98 & 1.84 & 1.88 & 1.73\\
& PAGR \cite{zhang2018progressive} & 2018 & 4.40 & 4.07 & 5.20 & 5.26 & 5.71 & 5.44 & 12.07 & 11.45 & 2.80 & 2.62 & 1.58 & 1.50\\
& PiCANet \cite{liu2018picanet} & 2018 & 4.81 & 4.52 & 4.17 & 3.86 & 5.34 & 4.91 & 7.71 & -7.27 & 2.75 & 2.46 & 2.08 & 1.89\\
& CPD \cite{wu2019cascaded} & 2019 & 4.00 & 3.80 & 4.45 & 4.33 & 4.76 & 4.58 & 6.98 & 6.65 & 2.29 & 2.16 & 2.26 & 2.15\\
& BASNet \cite{qin2019basnet} & 2019 & 7.17 & 6.94 & 7.10 & 6.91 & 7.70 & 7.48 & 7.84 & 7.74 & 2.14 & 2.11 & 2.59 & 2.51\\
& EGNet \cite{zhao2019egnet} & 2019 & 3.91 & 3.68 & 4.29 & 4.08 & 4.75 & 4.55 & 5.89 & 5.56 & 1.84 & 1.71 & 1.29 & 1.23\\
& AFNet \cite{feng2019attentive} & 2019 & 4.31 & 4.06 & 4.48 & 4.27 & 4.56 & 4.49 & 6.79 & 6.24 & 2.21 & 2.06 & 2.06 & 1.95\\
& PoolNet \cite{Liu21PamiPoolNet} & 2019 & 3.58 & 3.36 & 4.30 & 4.10 & 6.09 & 5.75 & 6.72 & 5.75 & 1.98 & 1.85 & 1.53 & 1.45\\
& GCPANet \cite{chen2020global} & 2020 & 4.45 & 4.18 & 5.26 & 5.04 & 5.01 & 4.75 & 5.74 & 5.60 & 1.63 & 1.52 & 1.58 & 1.51\\
& MINet \cite{pang2020multi} & 2020 & 4.97 & 4.69 & 6.03 & 5.77 & 6.97 & 6.67 & 8.17 & 7.97 & 1.99 & 1.93 & 1.48 & 1.45 \\
& $\text{F}^{3}\text{Met}$ \cite{wei2020f3net} & 2020 & 3.29 & 3.15 & 4.56 & 4.36 & 4.26 & 4.10 & 7.74 & 7.29 & 2.20 & 2.08 & 2.29 & 2.17 \\
& EBMGSOD \cite{jing_ebm_sod21} & 2021 & 4.32 & 4.10 & 5.03 & 4.81 & 4.40 & 4.29 & 5.46 & 5.18 & 2.53 & 2.39 & 2.30 & 2.17\\
& ICON \cite{zhuge2021salient} & 2021 & 2.64 & 2.54 & 4.16 & 4.02 & 3.93 & 3.90 & 5.13 & 5.01 & 1.32 & 1.24 & 1.20 & 1.14\\
& PFSNet \cite{ma2021pyramidal} & 2021 & 4.89 & 4.79 & 5.89 & 5.61 & 7.73 & 7.54 & 10.74 & 10.45 & 2.31 & 2.28 & 2.21 & 2.19\\
& EDN \cite{wu2022edn} & 2022 & 4.28 & 4.07 & 4.78 & 4.60 & 5.10 & 4.92 & 5.63 & 5.55 & 1.48 & 1.45 & 1.54 & 1.45\\
\hline
\multirow{7}{*}{\parbox{1.0cm}{Model \\ Calibration \\ Methods}} 
& Brier Loss \cite{brier1950verification} & 1950 & 3.43 & 3.17 & 4.39 & 4.15 & 4.44 & 4.22 & 5.03 & 4.22 & 1.48 & 1.38 & 1.21 & 1.15 \\
& TS \cite{guo2017calibration} & 2017 & 3.30 & 3.03 & 4.12 & 3.91 & 3.48 & 3.30 & 5.33 & 4.97 & 1.29 & 1.22 & 1.13 & 1.08\\
& MMCE \cite{MMCE} & 2018 & 3.44 & 3.20 & 4.38 & 4.17 & 3.66 & 3.48 & 5.55 & 5.19 & 1.40 & 1.31 & 1.36 & 1.29\\
& LS \cite{muller2019does} & 2019 & 2.97 & 2.92 & 3.88 & 3.81 & 4.08 & 4.99 & 5.67 & 5.42 & 1.46 & 1.27 & 1.32 & 0.99\\
& Mixup \cite{thulasidasan2019mixup} & 2019 & 3.01 & 2.76 & 4.47 & 4.21 & \color{blue}{1.84} & \color{blue}{1.26} & 5.26 & 4.99 & 1.28 & 1.11 & 1.73 & 1.48\\
& Focal Loss \cite{focal_loss} & 2020 & 2.23 & 2.14 & 3.73 & 3.43 & 3.03 & 2.93 & 4.77 & 4.59 & 1.30 & 1.16 & 1.40 & 1.08\\
& AdaFocal \cite{ghosh2022adafocal} & 2022 & \color{blue}{1.79} & 1.60 & \color{blue}{2.44} & 2.08 & 1.88 & 1.78 & \color{blue}{4.16} & 3.46 & \color{blue}{1.16} & 0.97 & \color{blue}{1.03} & 0.86\\
\hline
\multirow{2}{*}{\parbox{1.0cm}{Our \\ Methods}}
& $\text{ASLP}_{\text{ECE}}$  & 2023 & \color{red}{\textbf{1.37}} & \color{blue}{1.21} & \color{red}{\textbf{1.67}} & \color{blue}{1.33} & \color{red}{\textbf{1.77}} & 1.51 & \color{red}{\textbf{2.73}} & \color{blue}{2.41} & \color{red}{\textbf{0.97}} & \color{blue}{0.61} & \color{red}{\textbf{0.89}} & \color{blue}{0.41}\\
& $\text{ASLP}_{\text{MEI}}$ & 2023 & 20.78 & \color{red}{\textbf{0.00}} & 19.64 & \color{red}{\textbf{0.00}} & 19.74 & \color{red}{\textbf{0.00}} & 17.35 & \color{red}{\textbf{0.00}} & 22.47 & \color{red}{\textbf{0.00}} & 22.90 & \color{red}{\textbf{0.00}}\\
\bottomrule
\end{tabular}
\label{A_tab:ECE_SWEEP_Benchmark}
\end{table}

\begin{table}[htb!]
\centering
\scriptsize
\renewcommand{\arraystretch}{1.2}
\renewcommand{\tabcolsep}{1.8mm}
\caption{Salient object detection model calibration degree benchmark evaluated with $\text{ECE}_{\text{DEBIAS}}$ \cite{kumar2019verified}. We set he number of bins to $B = 10$. (values are shown in \% and {\color{red} red} and {\color{blue} blue} indicate the best and the second-best performance respectively.)}
\begin{tabular}{cl|c|c|c|c|c|c|c}
\toprule
\multicolumn{2}{c|}{\multirow{2}{*}{Methods}} & {\multirow{2}{*}{Year}} & 
\multicolumn{6}{c}{$\text{ECE}_{\text{DEBIAS}} (\%) \downarrow$}\\
& & & DUTS-TE \cite{DUTS-TE} & DUT-OMRON \cite{DUT-OMRON} & PASCAL-S \cite{PASCAL-S} & SOD \cite{SOD} & ECSSD \cite{ECSSD} & HKU-IS \cite{HKU-IS}\\
\midrule
\multirow{18}{*}{\parbox{1.0cm}{SOD \\ Methods}} 
& MSRNet \cite{li2017instance} & 2017 & 0.167 & 0.188 & 0.235 & 0.524 & 0.020 & 0.015\\
& SRM \cite{Wang_2017_ICCV} & 2017 & 0.419 & 0.358 & 0.436 & 1.221 & 0.186 & 0.110\\
& Amulet \cite{zhang2017amulet} & 2017 & 0.553 & 0.536 & 0.508 & 1.165 & 0.235 & 0.079\\
& BMPM \cite{zhang2018bi} & 2018 & 0.471 & 0.378 & 0.440 & 1.175 & 0.191 & 0.134 \\
& DGRL \cite{wang2018detect} & 2018 & 0.420 & 0.370 & 0.430 & 0.807 & 0.096 & 0.072\\
& PAGR \cite{zhang2018progressive} & 2018 & 0.340 & 0.418 & 0.470 & 1.568 & 0.137 & 0.053\\
& PiCANet \cite{liu2018picanet} & 2018 & 0.456 & 0.359 & 0.461 & 0.985 & 0.175 & 0.124\\
& CPD \cite{wu2019cascaded} & 2019 & 0.390 & 0.353 & 0.567 & 1.233 & 0.145 & 0.109\\
& BASNet \cite{qin2019basnet} & 2019 & 0.544 & 0.536 & 0.683 & 1.190 & 0.138 & 0.127\\
& EGNet \cite{zhao2019egnet} & 2019 & 0.318 & 0.304 & 0.576 & 0.860 & 0.109 & 0.066\\
& AFNet \cite{feng2019attentive} & 2019 & 0.381 & 0.348 & 0.471 & 0.934 & 0.132 & 0.091\\
& PoolNet \cite{Liu21PamiPoolNet} & 2019 & 0.335 & 0.326 & 0.612 & 0.907 & 0.107 & 0.055\\
& GCPANet \cite{chen2020global} & 2020 & 0.388 & 0.318 & 0.372 & 0.569 & 0.068 & 0.043\\
& MINet \cite{pang2020multi} & 2020 & 0.448 & 0.505 & 0.606 & 1.041 & 0.172 & 0.142\\
& $\text{F}^{3}\text{Met}$ \cite{wei2020f3net} & 2020 & 0.457 & 0.468 & 0.556 & 0.816 & 0.193 & 0.167\\
& EBMGSOD \cite{jing_ebm_sod21} & 2021 & 0.374 & 0.406 & 0.508 & 0.733 & 0.154 & 0.130\\
& ICON \cite{zhuge2021salient} & 2021 & 0.306 & 0.390 & 0.382 & 0.607 & 0.098 & 0.101\\
& PFSNet \cite{ma2021pyramidal} & 2021 & 0.323 & 0.339 & 0.539 & 0.594 & 0.588 & 0.435\\
& EDN \cite{wu2022edn} & 2022 & 0.285 & 0.281 & 0.407 & 0.745 & 0.068 & 0.061\\
\hline
\multirow{7}{*}{\parbox{1.0cm}{Model \\ Calibration \\ Methods}} 
& Brier Loss \cite{brier1950verification} & 1950 & 0.241 & 0.265 & 0.330 & 0.572 & 0.051 & 0.035\\
& TS \cite{guo2017calibration} & 2017 & 0.230 & 0.246 & 0.338 & 0.631 & 0.040 & 0.024\\
& MMCE \cite{MMCE} & 2018 & 0.250 & 0.269 & 0.378 & 0.752 & 0.054 & 0.039\\
& LS \cite{muller2019does} & 2019 & 0.218 & 0.241 & 0.303 & 0.570 & 0.047 & 0.034\\
& Mixup \cite{thulasidasan2019mixup} & 2019 & 0.143 & 0.211 & 0.110 & 0.423 & 0.078 & 0.482\\
& Focal Loss \cite{focal_loss} & 2020 & 0.135 & 0.193 & 0.262 & 0.518 & 0.070 & 0.061\\
& AdaFocal \cite{ghosh2022adafocal} & 2022 & \color{blue}{0.069} & 0.133 & 0.103 & 0.383 & 0.108 & 0.102\\
\hline
\multirow{2}{*}{\parbox{1.0cm}{Our \\ Methods}}
& $\text{ASLP}_{\text{ECE}}$  & 2023 & \color{red}{\textbf{0.056}} & \color{red}{\textbf{0.103}} & \color{red}{\textbf{0.061}} & \color{red}{\textbf{0.083}} & \color{red}{\textbf{0.024}} & \color{red}{\textbf{0.027}}\\
& $\text{ASLP}_{\text{MEI}}$ & 2023 & 4.565 & 4.027 & 4.079 & 3.112 & 5.095 & 5.301\\
\bottomrule
\end{tabular}
\label{A_tab:ECE_DEBIAS_Benchmark}
\end{table}

\clearpage

\section{Joint Distribution of Prediction Confidence and Prediction Accuracy on 6 Testing Datasets}
\label{A_sec:More_Joint_Plot}
Fig.~\ref{fig:more_joint_plot} presents the joint distribution of prediction confidence and prediction accuracy of our methods, existing model calibration methods and some of the salient object detection models on the six SOD testing datasets.

\begin{figure}[h!]
    \centering
    \includegraphics[width=0.95\textwidth]{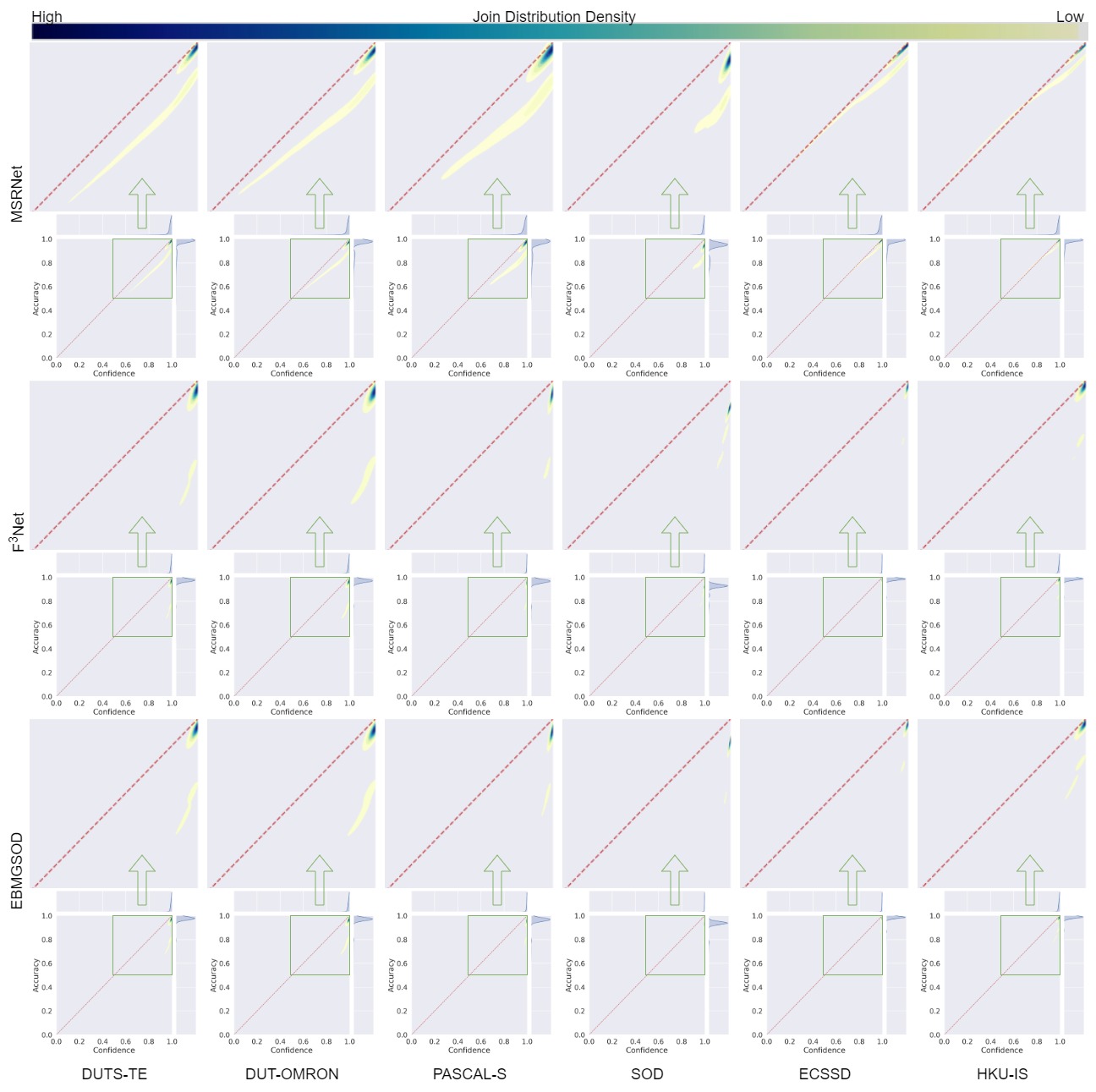}
    \caption{Joint distribution of prediction confidence (horizontal axis) and prediction accuracy (vertical axis) on the six SOD testing datasets.}
\end{figure}

\begin{figure}[h!]
    \ContinuedFloat
    \centering
    \includegraphics[width=0.95\textwidth]{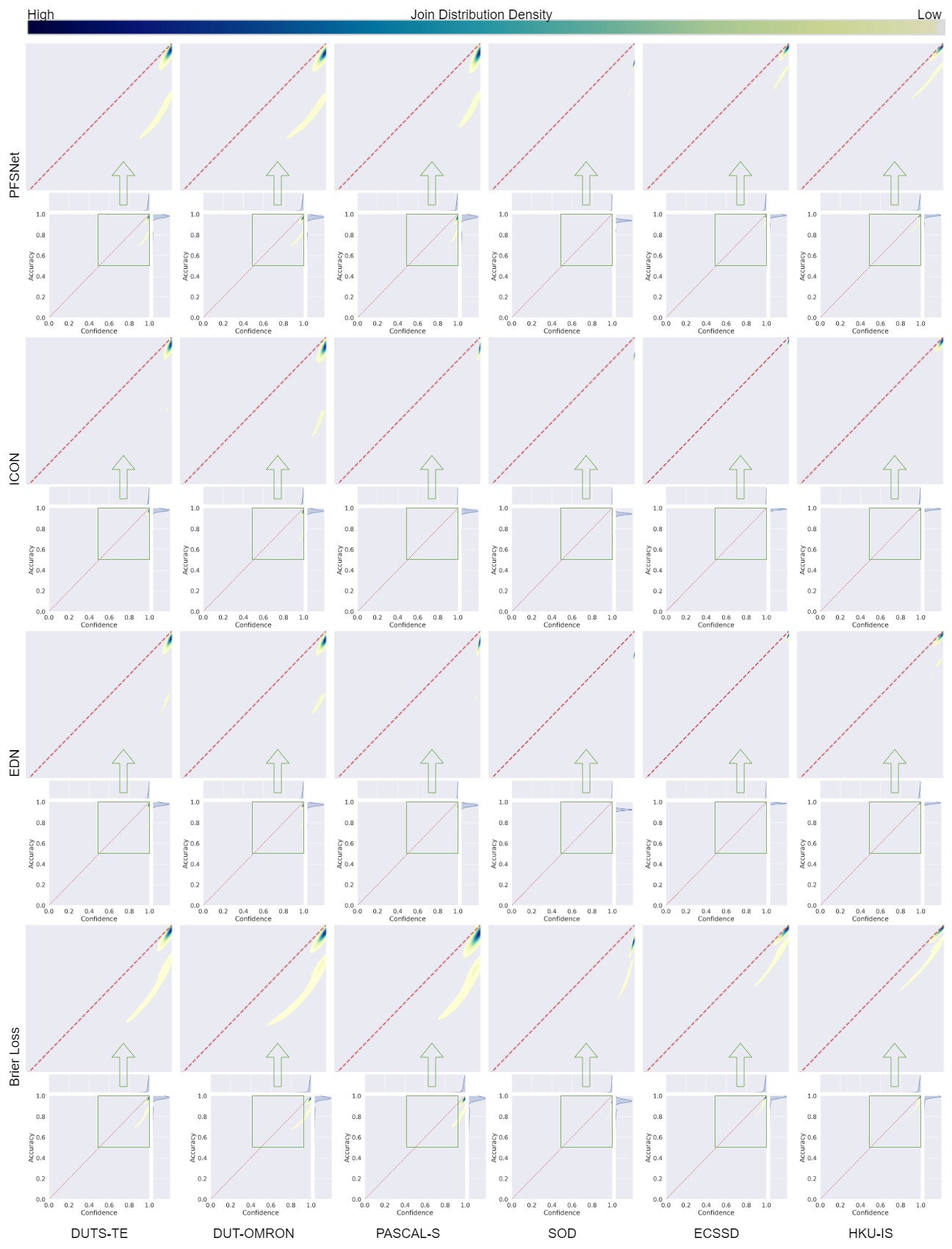}
    \caption{Joint distribution of prediction confidence (horizontal axis) and prediction accuracy (vertical axis) on the six SOD testing datasets.}
\end{figure}

\begin{figure}[h!]
    \ContinuedFloat
    \centering
    \includegraphics[width=0.95\textwidth]{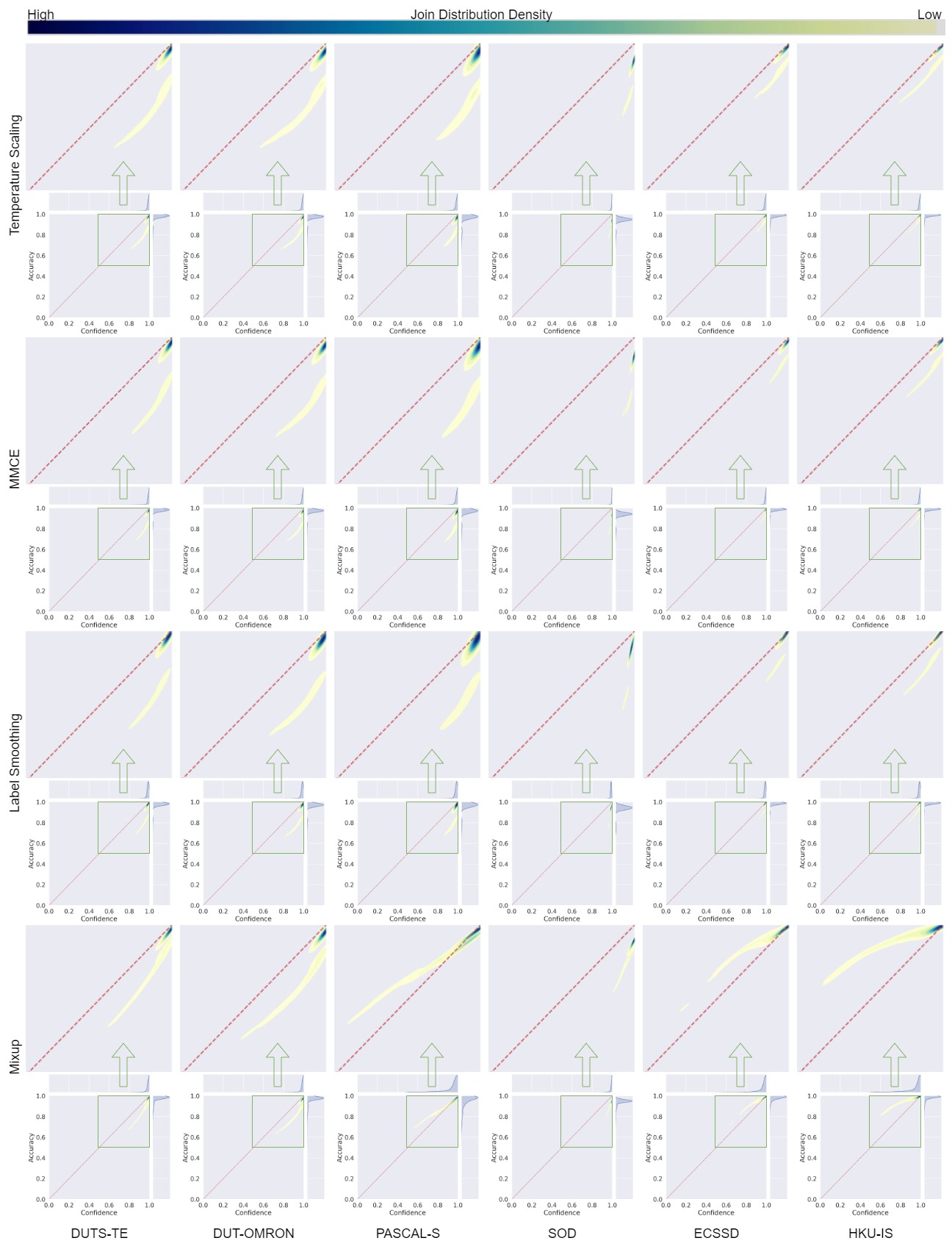}
    \caption{Joint distribution of prediction confidence (horizontal axis) and prediction accuracy (vertical axis) on the six SOD testing datasets.}
\end{figure}

\begin{figure}[h!]
    \ContinuedFloat
    \centering
    \includegraphics[width=0.95\textwidth]{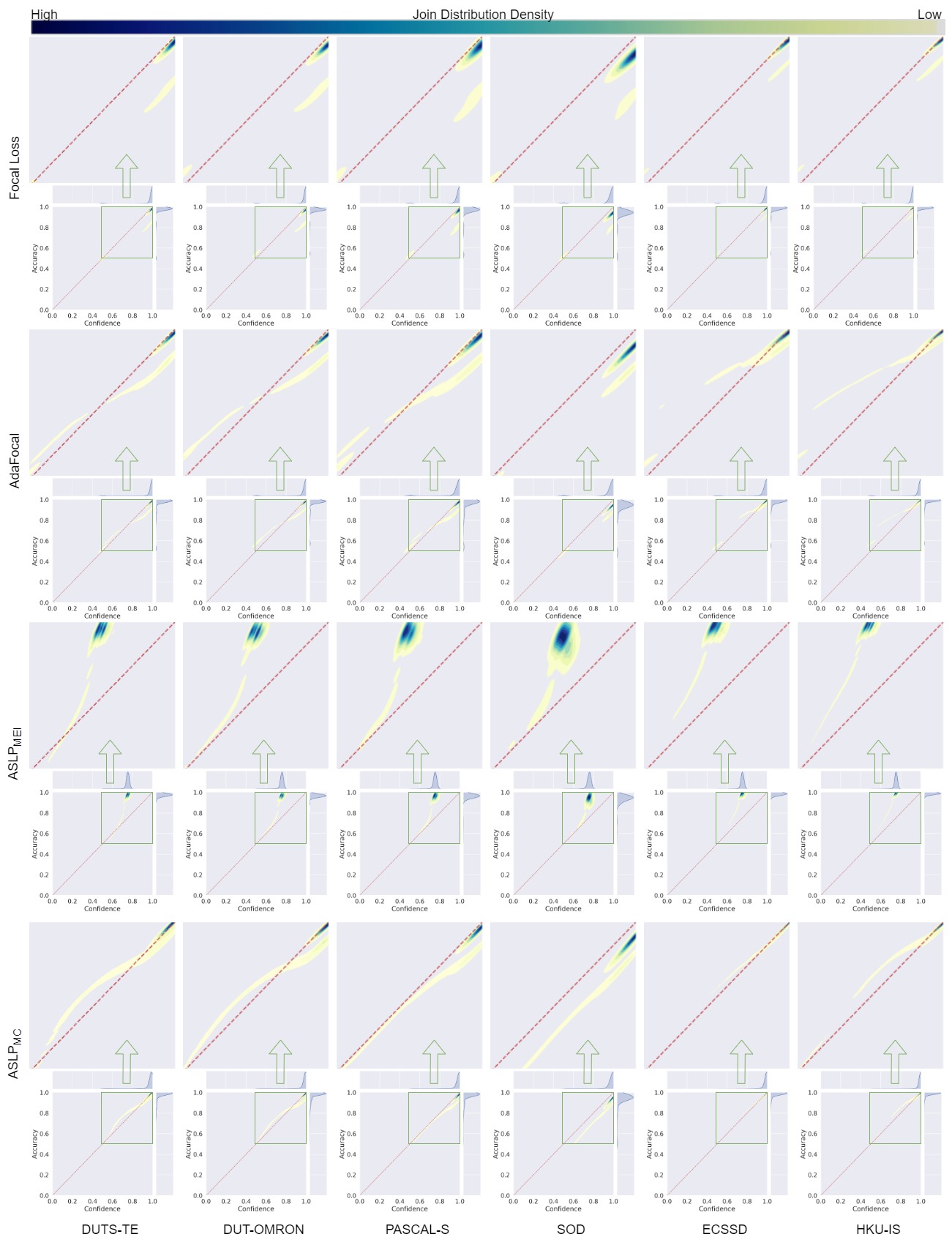}
    \caption{Joint distribution of prediction confidence (horizontal axis) and prediction accuracy (vertical axis) on the six SOD testing datasets.}
    \label{fig:more_joint_plot}
\end{figure}

\clearpage

\section{Generalisation to Existing SOD Methods}
\label{A_sec:Generalisation_to_SOD_Models}
We study the compatibility of the proposed updating rule $\text{ASLP}_{\text{MC}}$ with some of the existing state-of-the-art SOD models, including EBMGSOD \cite{jing_ebm_sod21}, ICON \cite{zhuge2021salient}, and EDN \cite{wu2022edn}, and present the model calibration results in Tab.~\ref{A_tab:Generalisation_to_SOD_Methods_Calibration_Results}. We implement the $\text{ASLP}_{\text{MC}}$ with the Hard Inversion (HI) label perturbation technique. The results demonstrate that our proposed method is readily compatible with existing SOD methods to improve their respective model calibration degrees. Further, we find that incorporation of our proposed $\text{ASLP}_{\text{MC}}$ into the training of existing SOD models do not negatively impact their classification performances as demonstrated in Tab.~\ref{A_tab:Generalisation_to_SOD_Methods_Classification_Results}. 

\begin{table}[htb!]
\centering
\scriptsize
\renewcommand{\arraystretch}{1.2}
\renewcommand{\tabcolsep}{1.0mm}
\caption{The model calibration degrees of existing Salient Object Detection models with or without the proposed Adaptive Label Augmentation are evaluated in terms of Equal-Width Expected Calibration Error, $\text{ECE}_{\text{EW}}$, and Equal-Width Over-confidence Error, $\text{OE}_{\text{EW}}$, with 10 bins ($B = 10$).}
\begin{tabular}{l|c|c|cc|cc|cc|cc|cc|cc}
\toprule
\multirow{2}{*}{Methods} & {\multirow{2}{*}{Year}} & {\multirow{2}{*}{$\text{ASLP}_{\text{MC}}$}} & \multicolumn{2}{c|}{DUTS-TE \cite{DUTS-TE}} & \multicolumn{2}{c|}{DUT-OMRON \cite{DUT-OMRON}} & \multicolumn{2}{c|}{PASCAL-S \cite{PASCAL-S}} & \multicolumn{2}{c|}{SOD \cite{SOD}} & \multicolumn{2}{c|}{ECSSD \cite{ECSSD}} & \multicolumn{2}{c}{HKU-IS \cite{HKU-IS}}\\
& & & $\text{ECE} \downarrow$ & $\text{OE} \downarrow$ & $\text{ECE} \downarrow$ & $\text{OE} \downarrow$ & $\text{ECE} \downarrow$ & $\text{OE} \downarrow$ & $\text{ECE} \downarrow$ & $\text{OE} \downarrow$ & $\text{ECE} \downarrow$ & $\text{OE} \downarrow$ & $\text{ECE} \downarrow$ & $\text{OE} \downarrow$ \\
\midrule
EBMGSOD \cite{jing_ebm_sod21} & 2021 & \xmark & 3.45 & 3.29 & 4.11 & 3.95 & 4.79 & 4.61 & 7.48 & 7.30 & 2.14 & 2.05 & 1.79 & 1.70\\
ICON \cite{zhuge2021salient} & 2021 & \xmark & 2.89 & 2.76 & 3.84 & 3.71 & 4.08 & 3.95 & 6.70 & 6.55 & 1.56 & 1.49 & 1.38 & 1.32\\
EDN \cite{wu2022edn} & 2022 & \xmark & 3.62 & 3.47 & 4.02 & 3.90 & 4.89 & 4.74 & 8.81 & 8.66 & 2.20 & 2.13 & 1.65 & 1.58\\
\hline
EBMGSOD & 2021 & \cmark & 1.60 & 1.34 & 1.91 & 1.74 & 2.45 & 2.23 & 5.48 & 5.21 & 0.77 & 0.47 & 0.75 & 0.22\\
ICON & 2021 & \cmark & 1.28 & 1.05 & 1.88 & 1.67 & 2.45 & 2.17 & 5.17 & 4.91 & 1.25 & 0.07 & 1.10 & 0.05\\
EDN & 2022 & \cmark & 2.02 & 1.77 & 2.23 & 2.03 & 2.74 & 2.55 & 6.77 & 6.46 & 0.82 & 0.52 & 0.71 & 0.35\\
\bottomrule
\end{tabular}
\label{A_tab:Generalisation_to_SOD_Methods_Calibration_Results}
\end{table}

\begin{table}[htb!]
\centering
\scriptsize
\renewcommand{\arraystretch}{1.2}
\renewcommand{\tabcolsep}{1.0mm}
\caption{The dense classification accuracy of Salient Object Detection models with or without the proposed Adaptive Label Augmentation is evaluated with maximum F-measure and maximum E-measure \cite{fan2018enhanced}.}
\begin{tabular}{l|c|c|cc|cc|cc|cc|cc|cc}
\toprule
\multirow{2}{*}{Methods} & {\multirow{2}{*}{Year}} & {\multirow{2}{*}{$\text{ASLP}_{\text{MC}}$}} & \multicolumn{2}{c|}{DUTS-TE \cite{DUTS-TE}} & \multicolumn{2}{c|}{DUT-OMRON \cite{DUT-OMRON}} & \multicolumn{2}{c|}{PASCAL-S \cite{PASCAL-S}} & \multicolumn{2}{c|}{SOD \cite{SOD}} & \multicolumn{2}{c|}{ECSSD \cite{ECSSD}} & \multicolumn{2}{c}{HKU-IS \cite{HKU-IS}}\\
& & & $F_{\text{max}} \uparrow$ & $E_{\text{max}} \uparrow$ & $F_{\text{max}} \uparrow$ & $E_{\text{max}} \uparrow$ & $F_{\text{max}} \uparrow$ & $E_{\text{max}} \uparrow$ & $F_{\text{max}} \uparrow$ & $E_{\text{max}} \uparrow$ & $F_{\text{max}} \uparrow$ & $E_{\text{max}} \uparrow$ & $F_{\text{max}} \uparrow$ & $E_{\text{max}} \uparrow$ \\
\midrule
EBMGSOD \cite{jing_ebm_sod21} & 2021 & \xmark & 0.850 & 0.927 & 0.762 & 0.867 & 0.830 & 0.896 & 0.834 & 0.800 & 0.914 & 0.944 & 0.906 & 0.952\\
ICON \cite{zhuge2021salient} & 2021 & \xmark & 0.860 & 0.924 & 0.773 & 0.876 & 0.850 & 0.899 & 0.815 & 0.854 & 0.933 & 0.954 & 0.919 & 0.953\\
EDN \cite{wu2022edn} & 2022 & \xmark & 0.893 & 0.949 & 0.821 & 0.900 & 0.879 & 0.920 & 0.840 & 0.860 & 0.950 & 0.969 & 0.940 & 0.970\\
\hline
EBMGSOD & 2021 & \cmark & 0.853 & 0.930 & 0.767 & 0.871 & 0.841 & 0.901 & 0.839 & 0.807 & 0.923 & 0.946 & 0.912 & 0.956\\
ICON & 2021 & \cmark & 0.864 & 0.929 & 0.776 & 0.877 & 0.857 & 0.904 & 0.819 & 0.855 & 0.940 & 0.959 & 0.926 & 0.959\\
EDN & 2022 & \cmark & 0.898 & 0.954 & 0.824 & 0.901 & 0.880 & 0.923 & 0.848 & 0.866 & 0.952 & 0.971 & 0.942 & 0.972\\
\bottomrule
\end{tabular}
\label{A_tab:Generalisation_to_SOD_Methods_Classification_Results}
\end{table}

\clearpage

\section{Experiments on Additional Dense Classification Tasks}
\label{A_sec:Additional_Experiments_on_Dense_Binary_Classification_Tasks}
\subsection{Camouflaged Object Detection}
We train our model on the COD10K training set \cite{fan2020camouflaged} which consists of 6,000 training samples. We partition it into a training set of 5,400 samples and a validation set of 600 samples. Four testing datasets, including the COD10K testing set \cite{fan2020camouflaged}, NC4K \cite{lv2021simultaneously}, CAMO \cite{le2019anabranch} and CHAMELEON \cite{skurowski2018animal}, are used to evaluate the model calibration degree and dense binary classification accuracy. We train the models for 50 epochs and the rest of settings follow those in Salient Object Detection.

We apply the proposed $\text{ASLP}_{\text{MC}}$ with Hard Inversion (HI) and Soft Inversion (SI) label perturbation techniques and $\text{ALS}_{\text{MC}}$ to improve the model calibration degrees with four label perturbation techniques and report the results in Tab.~\ref{tab:Model_Calibration_on_Camouflaged_Object_Detection}. It can be observed that both ASLP with various label perturbation techniques and ALS can also significantly improve model calibration degrees in Camouflaged Object Detection models. Further, we show that the improvements in model calibration degree are achieved without negatively impacting the classification accuracy as shown in Tab.~\ref{tab:Classification_Accuracy_on_Camouflaged_Object_Detection}.

\begin{table}[htb!]
\centering
\scriptsize
\renewcommand{\arraystretch}{1.2}
\renewcommand{\tabcolsep}{1.5mm}
\caption{Application Adaptive Stochastic Label Perturbation (ASLP) with different label perturbation techniques in Camouflaged Object Detection task. The model calibration degrees are evaluated with Equal-Width Expected Calibration Error ($\text{ECE}_{\text{EW}}$) and Equal-Width Over-confidence Error ($\text{OE}_{\text{EW}}$) with 10 bins. Results are presented in (\%).}
\begin{tabular}{l|ccc|cc|cc|cc|cc}
\toprule
\multirow{2}{*}{Methods} & \multicolumn{3}{c|}{Perturbation Params} & \multicolumn{2}{c|}{COD10K \cite{fan2020camouflaged}} & \multicolumn{2}{c|}{NC4K \cite{lv2021simultaneously}} & \multicolumn{2}{c|}{CHAMELEON \cite{skurowski2018animal}} & \multicolumn{2}{c}{CAMO \cite{le2019anabranch}} \\
& $\alpha$ & $\beta$ & e & 
$\text{ECE}_{\text{EW}} \downarrow$ & $\text{OE}_{\text{EW}} \downarrow$ & $\text{ECE}_{\text{EW}} \downarrow$ & $\text{OE}_{\text{EW}} \downarrow$ & $\text{ECE}_{\text{EW}} \downarrow$ & $\text{OE}_{\text{EW}} \downarrow$ & $\text{ECE}_{\text{EW}} \downarrow$ & $\text{OE}_{\text{EW}} \downarrow$ \\
\midrule
Baseline (\enquote{COD-B}) & 0 & 0 & \xmark & 1.65 & 1.55 & 2.75 & 2.60 & 0.63 & 0.57 & 3.62 & 3.46\\
\hline
$\text{COD-ASLP}_{\text{MC}}^{\text{HI}}$ & 
$\alpha_{\text{ada}}$ & 1.0 & \xmark & 1.06 & 0.81 & 1.67 & 1.51 & 0.43 & 0.12 & 2.00 & 1.80\\
$\text{COD-ASLP}_{\text{MC}}^{\text{SI}}$ & $\alpha_{\text{ada}}$ & 0.75 & \xmark & 1.05 & 0.80 & 1.72 & 1.55 & 0.44 & 0.21 & 2.03 & 1.85\\
$\text{COD-ALS}_{\text{MC}}$ & 1.0 & $\beta_{\text{ada}}$ & \xmark & 1.03 & 0.76 & 1.69 & 1.53 & 0.45 & 0.28 & 1.98 & 1.81\\
\bottomrule
\end{tabular}
\label{tab:Model_Calibration_on_Camouflaged_Object_Detection}
\end{table}

\begin{table}[htb!]
\centering
\scriptsize
\renewcommand{\arraystretch}{1.2}
\renewcommand{\tabcolsep}{1.5mm}
\caption{Application Adaptive Stochastic Label Perturbation (ASLP) with different label perturbation techniques in the Camouflaged Object Detection task. The dense classification accuracy is evaluated with maximum F-measure and maximum E-measure \cite{fan2018enhanced}.}
\begin{tabular}{l|ccc|cc|cc|cc|cc}
\toprule
\multirow{2}{*}{Methods} & \multicolumn{3}{c|}{Perturbation Params} & \multicolumn{2}{c|}{COD10K \cite{fan2020camouflaged}} & \multicolumn{2}{c|}{NC4K \cite{lv2021simultaneously}} & \multicolumn{2}{c|}{CHAMELEON \cite{skurowski2018animal}} & \multicolumn{2}{c}{CAMO \cite{le2019anabranch}} \\
& $\alpha$ & $\beta$ & e & 
$F_{\text{max}} \uparrow$ & $E_{\text{max}} \uparrow$ & $F_{\text{max}} \uparrow$ & $E_{\text{max}} \uparrow$ & $F_{\text{max}} \uparrow$ & $E_{\text{max}} \uparrow$ & $F_{\text{max}} \uparrow$ & $E_{\text{max}} \uparrow$ \\
\midrule
Baseline (\enquote{COD-B}) & 0 & 0 & \xmark & 0.715 & 0.886 & 0.803 & 0.902 & 0.843 & 0.940 & 0.749 & 0.855\\
\hline
$\text{COD-ASLP}_{\text{MC}}^{\text{HI}}$ & 
$\alpha_{\text{ada}}$ & 1.0 & \xmark & 0.716 & 0.886 & 0.803 & 0.902 & 0.845 & 0.942 & 0.756 & 0.861\\
$\text{COD-ASLP}_{\text{MC}}^{\text{SI}}$ & $\alpha_{\text{ada}}$ & 0.75 & \xmark & 0.716 & 0.887 & 0.802 & 0.904 & 0.844 & 0.943 & 0.759 & 0.867\\
$\text{COD-ALS}_{\text{MC}}$ & 1.0 & $\beta_{\text{ada}}$ & \xmark & 0.717 & 0.887 & 0.804 & 0.905 & 0.845 & 0.941 & 0.767 & 0.868\\
\bottomrule
\end{tabular}
\label{tab:Classification_Accuracy_on_Camouflaged_Object_Detection}
\end{table}

\begin{table}[h!]
\centering
\scriptsize
\renewcommand{\arraystretch}{1.2}
\renewcommand{\tabcolsep}{1.5mm}
\caption{Application Adaptive Stochastic Label Perturbation (ASLP) with different label perturbation techniques in the Smoke Detection (SD) task. Model calibration degree is evaluated with Equal-Width Expected Calibration Error ($\text{ECE}_{\text{EW}}$) and Equal-Width Over-confidence Error ($\text{OE}_{\text{EW}}$) with 10 bins. Dense classification accuracy is evaluated with maximum F-measure and maximum E-measure \cite{fan2018enhanced}.}
\begin{tabular}{l|ccc|cc|cc}
\toprule
\multirow{2}{*}{Methods} & \multicolumn{3}{c|}{Perturbation Params} & \multicolumn{4}{c}{SMOKE5K \cite{fan2020camouflaged}} \\
& $\alpha$ & $\beta$ & e & 
$\text{ECE}_{\text{EW}} (\%) \downarrow$ & $\text{OE}_{\text{EW}} (\%) \downarrow$ & $F_{\text{max}} \uparrow$ & $E_{\text{max}} \uparrow$ \\
\midrule
Baseline (\enquote{SD-B}) & 0 & 0 & \xmark & 0.164 & 0.154 & 0.763 & 0.930 \\
\hline
$\text{SD-ASLP}_{\text{MC}}^{\text{HI}}$ & 
$\alpha_{\text{ada}}$ & 1.0 & \xmark & 0.071 & 0.063 & 0.763 & 0.930\\
$\text{SD-ASLP}_{\text{MC}}^{\text{SI}}$ & $\alpha_{\text{ada}}$ & 0.75 & \xmark & 0.076 & 0.072 & 0.765 & 0.932\\
$\text{SD-ALS}_{\text{MC}}$ & 1.0 & $\beta_{\text{ada}}$ & \xmark & 0.079 & 0.072 & 0.764 & 0.930\\
\bottomrule
\end{tabular}
\label{tab:Smoke_Detection_Results}
\end{table}

\subsection{Smoke Detection}
We train our model on the SMOKE5K training set \cite{yan2022transmission} which consists of 4,600 training samples of real smoke. We partition it into a training set of 4,200 samples and a validation set of 400 samples. SMOKE5K testing set, comprising of 400 real-smoke images, is used to evaluate model calibration degree and dense binary classification accuracy. 

We apply the proposed $\text{ASLP}_{\text{MC}}$ with Hard Inversion (HI) and Soft Inversion (SI) label perturbation techniques and $\text{ALS}_{\text{MC}}$ to improve the model calibration degrees and report the results in Tab.~\ref{tab:Smoke_Detection_Results}. It can be observed that both $\text{ASLP}_{\text{MC}}$ with different label perturbation techniques and $\text{ALS}_{\text{MC}}$ can significantly improve model calibration degrees in Smoke Detection models, despite the baseline model already achieving higher calibration degrees compared with baseline models in Salient Object Detection and Camouflaged Object Detection. We can observe that our proposed methods still achieve improvements in model calibration degree without negatively impacting the classification accuracy.


\section{Experiments on Additional Dense Multi-Class Classification Task - Semantic Segmentation}
\label{A_sec:Additional_Experiments_on_Semantic_Segmentation}
We evaluate our proposed methods on the PASCAL VOC 2012 segmentation dataset \cite{everingham2010pascal} which has 20 foreground categories and 1 background category. The official split has 1,464, 1,449, and 1,456 samples in training, validation and testing sets respectively. Following previous work \cite{chen2017deeplab}, we use an augmented training set comprising of 10,582 samples, provided by \cite{hariharan2011semantic}, for model training. As we do not have access to the groundtruth of \enquote{official testing set} whose evaluation is server-based, we adopt the \enquote{official validation set} as \enquote{our testing set} to evaluate the model calibration degrees and segmentation accuracies. Similar to our implementation in dense binary classification tasks, we partition the augmented training set into \enquote{our training set} of 9,582 images and \enquote{our validation set} of 1,000 images. 


We adopt DeepLabv3+ \cite{chen2017deeplab} with a ResNet50 backbone as our baseline model (\enquote{SS-B}) and apply the proposed $\text{ASLP}_{\text{MC}}$ with with the Hard Inversion (HI) label perturbation technique and $\text{ALS}_{\text{MC}}$ to improve the model calibration degrees. We report model calibration results evaluated in terms of Equal-Width Expected Calibration Error ($\text{ECE}_{\text{EW}}$) and Equal-Width Over-confidence Error ($\text{OE}_{\text{EW}}$) with 10 bins in Tab.~\ref{tab:Semantic_Segmentation_Results}.

\begin{table}[h!]
\centering
\scriptsize
\renewcommand{\arraystretch}{1.2}
\renewcommand{\tabcolsep}{1.5mm}
\caption{Application Adaptive Stochastic Label Perturbation (ASLP) with different label perturbation techniques in a Semantic Segmentation (SS) task. Model calibration degree is evaluated with Equal-Width Expected Calibration Error ($\text{ECE}_{\text{EW}}$) and Equal-Width Over-confidence Error ($\text{OE}_{\text{EW}}$) with 10 bins. Segmentation accuracy is evaluated with Intersection-over-Union (IoU) \cite{chen2017deeplab}.}
\begin{tabular}{l|ccc|cc|c}
\toprule
\multirow{2}{*}{Methods} & \multicolumn{3}{c|}{Perturbation Params} & \multicolumn{3}{c}{PASCAL VOC 2012 \cite{everingham2010pascal}} \\
& $\alpha$ & $\beta$ & e & 
$\text{ECE}_{\text{EW}} (\%) \downarrow$ & $\text{OE}_{\text{EW}} (\%) \downarrow$ & IoU (\%) $\uparrow$ \\
\midrule
Baseline (\enquote{SS-B}) & 0 & 0 & \xmark & 6.29 & 5.37 & 71.2 \\
\hline
$\text{SS-ASLP}_{\text{MC}}^{\text{HI}}$ & $\alpha_{\text{ada}}$ & 1.0 & \xmark & 4.05 & 3.13 & 71.3\\
$\text{SS-ALS}_{\text{MC}}$ & 1.0 & $\beta_{\text{ada}}$ & \xmark & 4.10 & 3.24 & 71.5\\
\bottomrule
\end{tabular}
\label{tab:Semantic_Segmentation_Results}
\end{table}

\clearpage

\section{Static Stochastic Label Perturbation}
\subsection{Implementation}
\label{A_sub_sec:SLP_Implementation}
We implement four static stochastic label perturbation techniques each of which have a single label perturbation probability $\alpha$ for the entire training dataset. Their details are as below:
\begin{itemize}
    \item \textbf{Hard Inversion (HI)} produces the perturbed label by inverting the groundtruth label with $p = \text{LP}(y, 2) = 1 - y$. Intuitively, it switches the label category from \enquote{salient} to \enquote{non-salient} and vice versa. The label perturbation probability is limited to $\alpha \in [0, 0.5)$ to avoid learning a complete opposite task (non-salient background detection).
    
    \item \textbf{Soft Inversion (SI)} inverts the label category and softens the target with  $p = \text{LP}(y, 0.75) = -0.5y + 0.75$. Similarly, the label perturbation probability is limited to $p \in [0, \frac{1}{1.5})$ to prevent from learning a complete opposite task.
    
    \item \textbf{Moderation (M)} transforms groundtruth label into a prior distribution on the two classes (salient foreground object v.s. non-salient background), as $p = \text{LP}(y, 0.5) = 0.5$. The label perturbation probability is in the range $\alpha \in [0, 1)$.
    
    \item \textbf{Dynamic Moderation (DM)} introduces additional stochasticity on top of the \textbf{Moderation} method by adding an additional noise sampled from a truncated normal distribution\footnote{Truncated normal distribution $\mathcal{N}_{a,b}(\mu, \sigma)$, where $a$ and $b$ indicate the bound, $\mu$ is the mean and $\sigma$ is the variance.}: $p = \text{LP}(y, 0.5) + e = 0.5 + e, \,\, e \sim \mathcal{N}_{-0.5, 0.5}(0, 1)$. The label perturbation probability is in the range $\alpha \in [0, 1)$.
\end{itemize}

\subsection{Effect of Static Stochastic Label Perturbation Techniques on Model Calibration Degrees}
\label{A_sub_Sec:SLP_Model_Calibration}

Fig.~\ref{fig:Static_SLP_Model_Calibration} presents model calibration degrees, evaluated in terms of Equal-Width Expected Calibration Error ($\text{ECE}_{\text{EW}}$) and Equal-Width Over-confidence Error ($\text{OE}_{\text{EW}}$) with 100 bins ($B = 100$), of various static stochastic label perturbation techniques, in which a unique label perturbation probability $\alpha$ is set for all samples throughout the training. We can observe that, with an increasing label perturbation probability, ECE scores tend to reduce to a critical points before climbing. This is caused by the model transitioning from being over-confident to under-confident. This is evidenced in the OE scores which keep decreasing until 0 when the label perturbation probability increases. Further, \enquote{HI} has the steepest change in terms of both ECE and OE scores. This rate can be related to the product of label perturbation probability and strength $\alpha \beta$. We also find a dampening effect of additional stochasticity at high label perturbation probability range ($\alpha \in [0.4, 0.6]$) where \enquote{DM} is consistently less under-confident than \enquote{M}.

\begin{table}[htb!]
\centering
\scriptsize
\renewcommand{\arraystretch}{1.2}
\renewcommand{\tabcolsep}{1.0mm}
\caption{Effect label perturbation probability range (\%) for different static stochastic label perturbation techniques to reduce the Equal-Width Expected Calibration Error ($\text{ECE}_{\text{EW}}$) scores on the six testing datasets.}
\begin{tabular}{l|c|c|c|c|c|c}
\toprule
Static SLP Technique & DUTS-TE \cite{DUTS-TE} & DUT-OMRON \cite{DUT-OMRON} & PASCAL-S \cite{PASCAL-S} & SOD \cite{SOD} & ECSSD \cite{ECSSD} & HKU-IS \cite{HKU-IS}\\
\midrule
Hard Inversion (HI)     & 0 - 5\% & 0 - 3\% & 0 - 5\% & 0 - 10\% & 0 - 1\% & 0 - 1\%\\
Soft Inversion (SI)     & 0 - 5\% & 0 - 5\% & 0 - 5\% & 0 - 10\% & 0 - 2\% & 0 - 2\%\\
Moderation (M)          & 0 - 5\% & 0 - 5\% & 0 - 5\% & 0 - 20\% & 0 - 3\% & 0 - 3\%\\
Dynamic Moderation (DM) & 0 - 5\% & 0 - 5\% & 0 - 5\% & 0 - 20\% & 0 - 3\% & 0 - 3\%\\
\bottomrule
\end{tabular}
\label{A_tab:Static_SLP_Effective_Alpha_Range}
\end{table}

The effective label perturbation probability range for each static SLP technique on the six testing datasets is summarised in Tab.~\ref{A_tab:Static_SLP_Effective_Alpha_Range}. In general, the static SLPs have a wide range of effective label perturbation probability leading to reduced ECE scores compared to the baseline. The widest effective label perturbation probability range is found on the SOD dataset, with 0 - 10\% for \enquote{HI} and \enquote{SI} and 0 - 20\% for \enquote{M} and \enquote{DM}. This can be attributed to the baseline model being the most mis-calibrated on the SOD dataset, thus stronger label augmentation measures are required to transform the model from being over-confident to being under-confident. On the other hand, the baseline model is the most calibrated on the ECSSD and the HKU-IS datasets, indicating a small gap between the prediction confidence and prediction accuracy distributions. That leaves little space for label augmentation techniques to reduce the prediction confidence in order to match the prediction accuracy.

\subsection{Effect of Static Stochastic Label Perturbation Techniques on Dense Binary Classification Performance}
\label{A_sub_sec:SLP_Classification_Performance}
\begin{figure}
\centering
\begin{subfigure}{0.45\textwidth}
    \includegraphics[width=\textwidth]{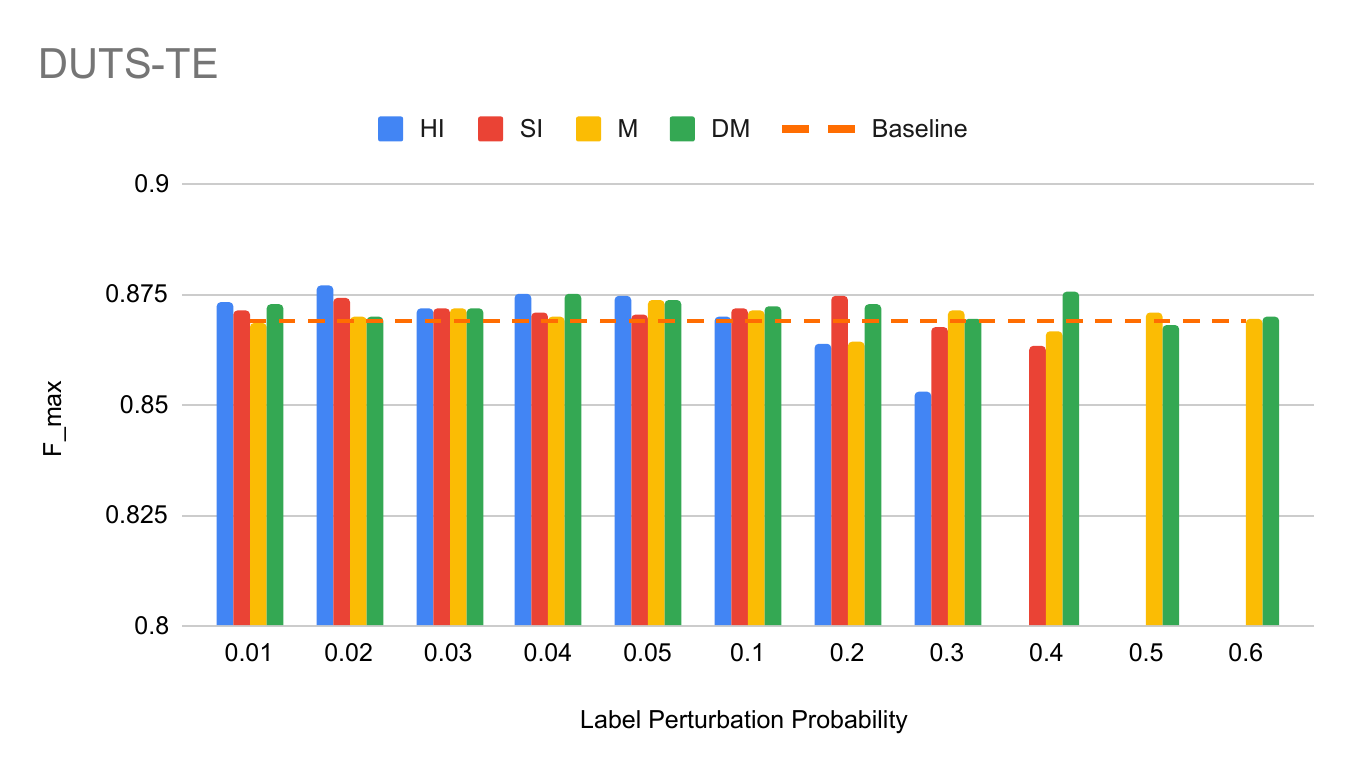}
    \caption{DUTS-TE}
\end{subfigure}
\begin{subfigure}{0.45\textwidth}
    \includegraphics[width=\textwidth]{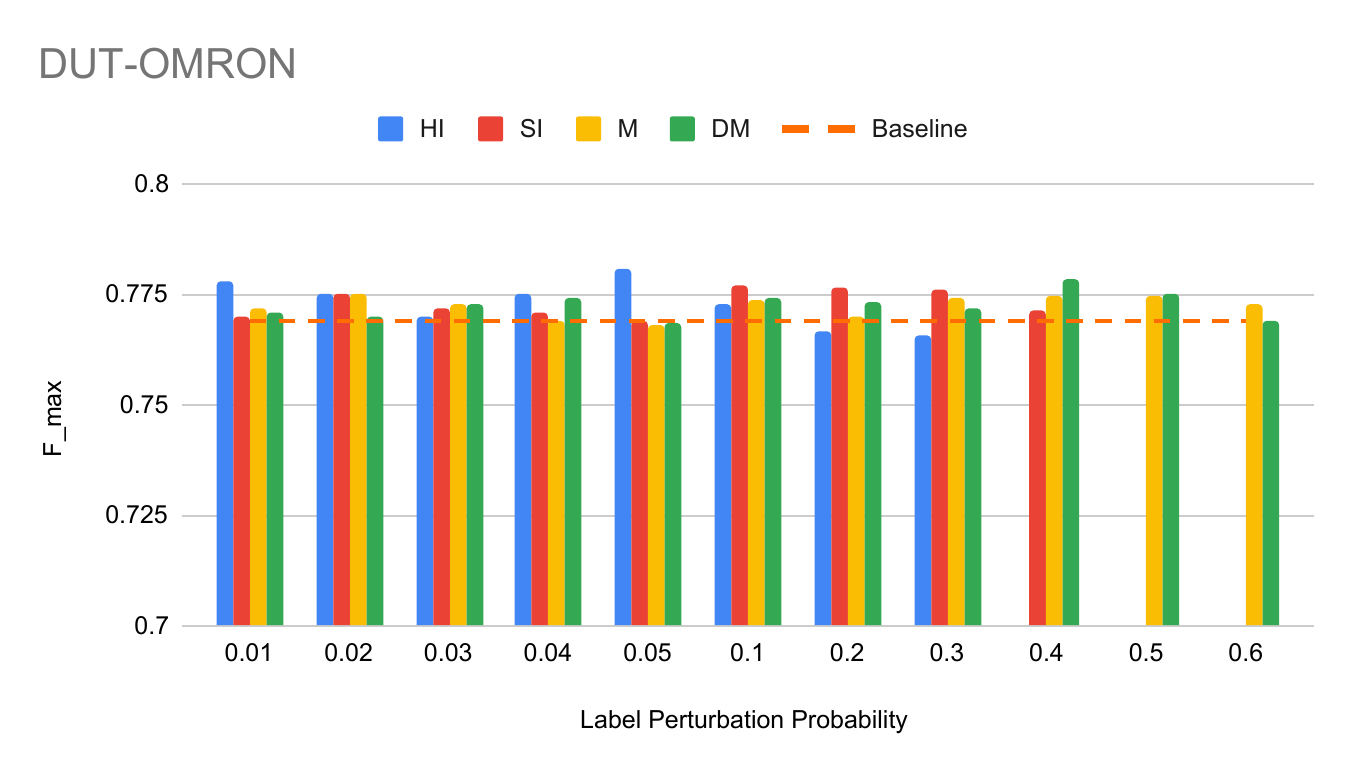}
    \caption{DUT-OMRON}
\end{subfigure}
\begin{subfigure}{0.45\textwidth}
    \includegraphics[width=\textwidth]{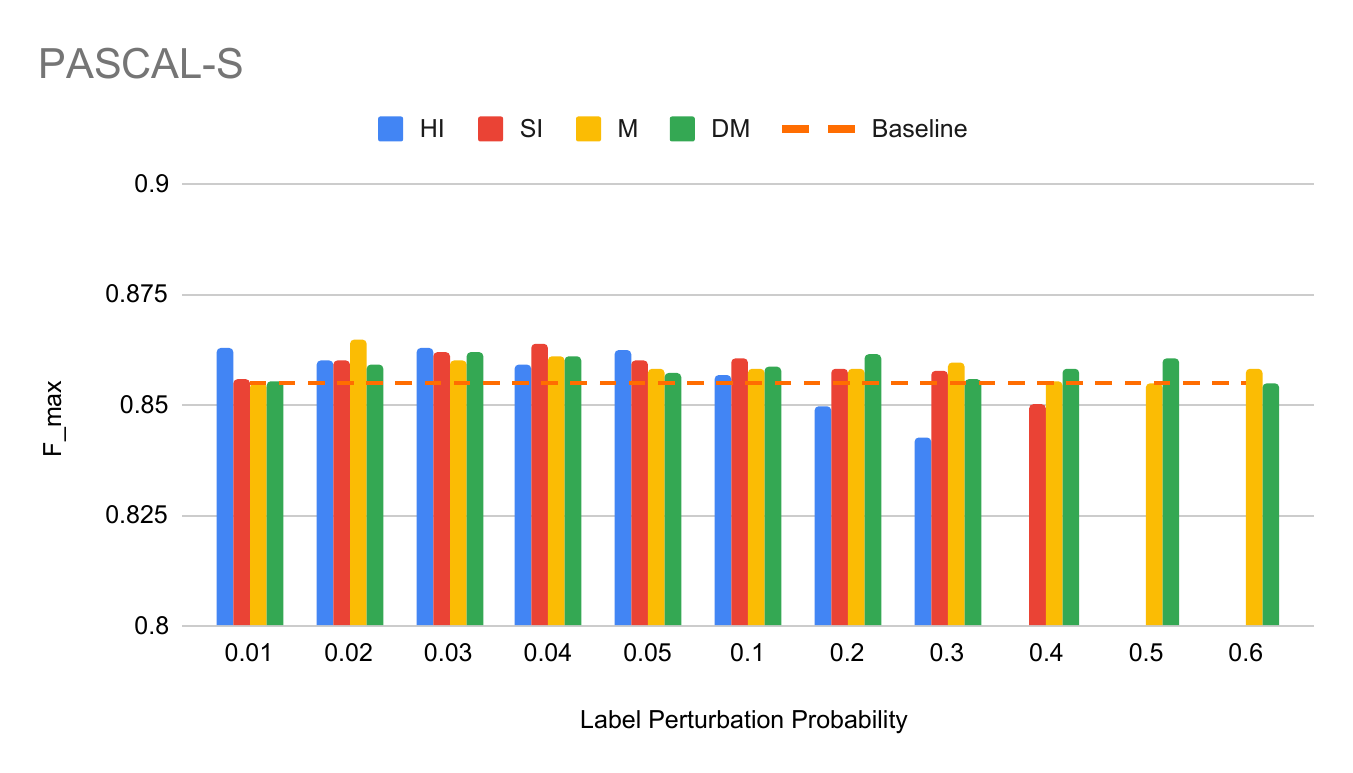}
    \caption{PASCAL-S}
\end{subfigure}
\begin{subfigure}{0.45\textwidth}
    \includegraphics[width=\textwidth]{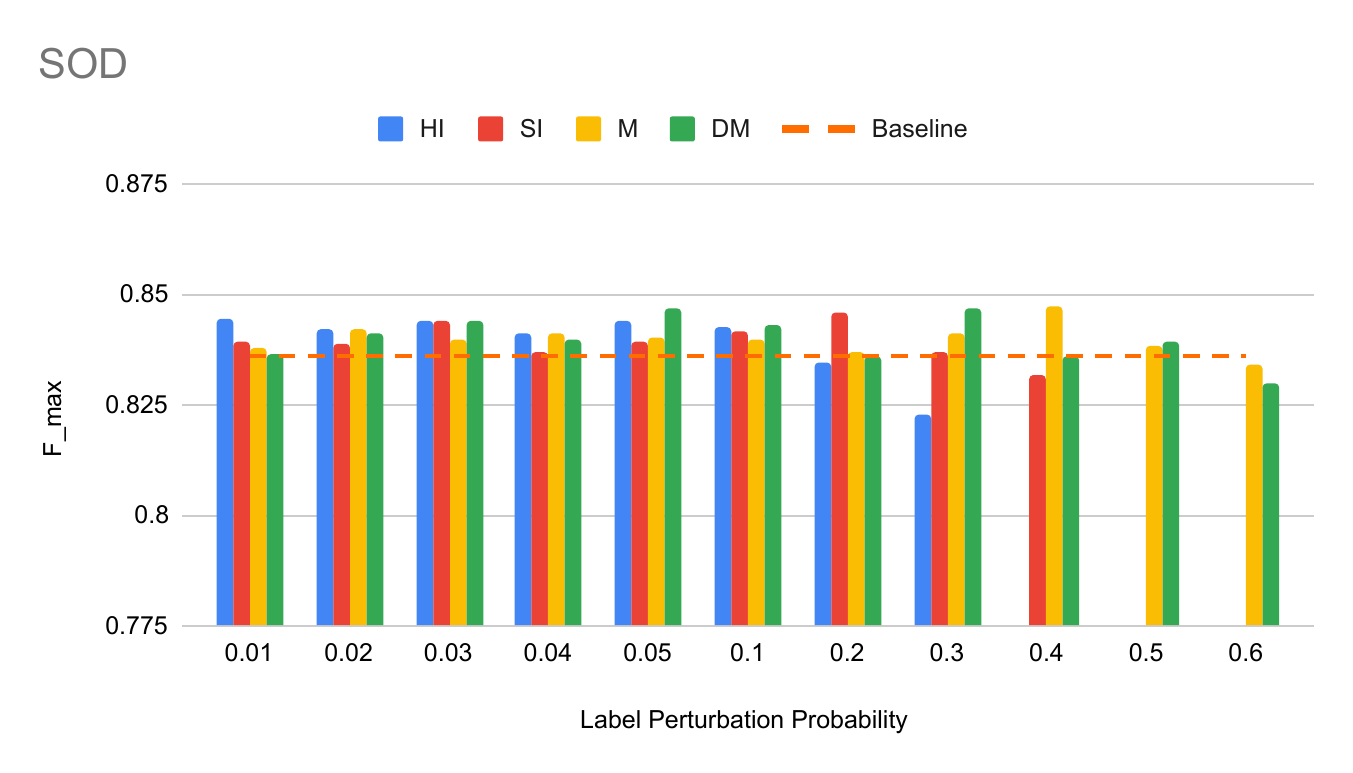}
    \caption{SOD}
\end{subfigure}
\begin{subfigure}{0.45\textwidth}
    \includegraphics[width=\textwidth]{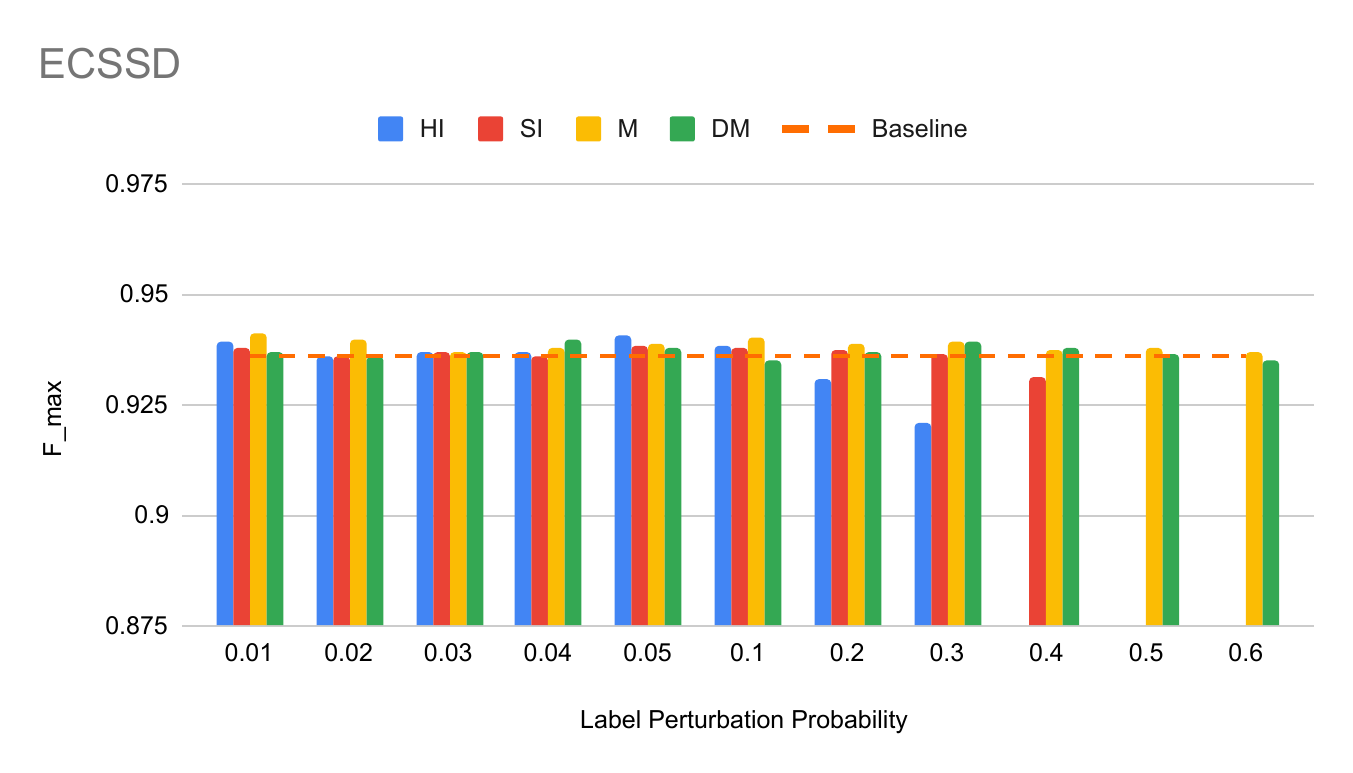}
    \caption{ECSSD}
\end{subfigure}
\begin{subfigure}{0.45\textwidth}
    \includegraphics[width=\textwidth]{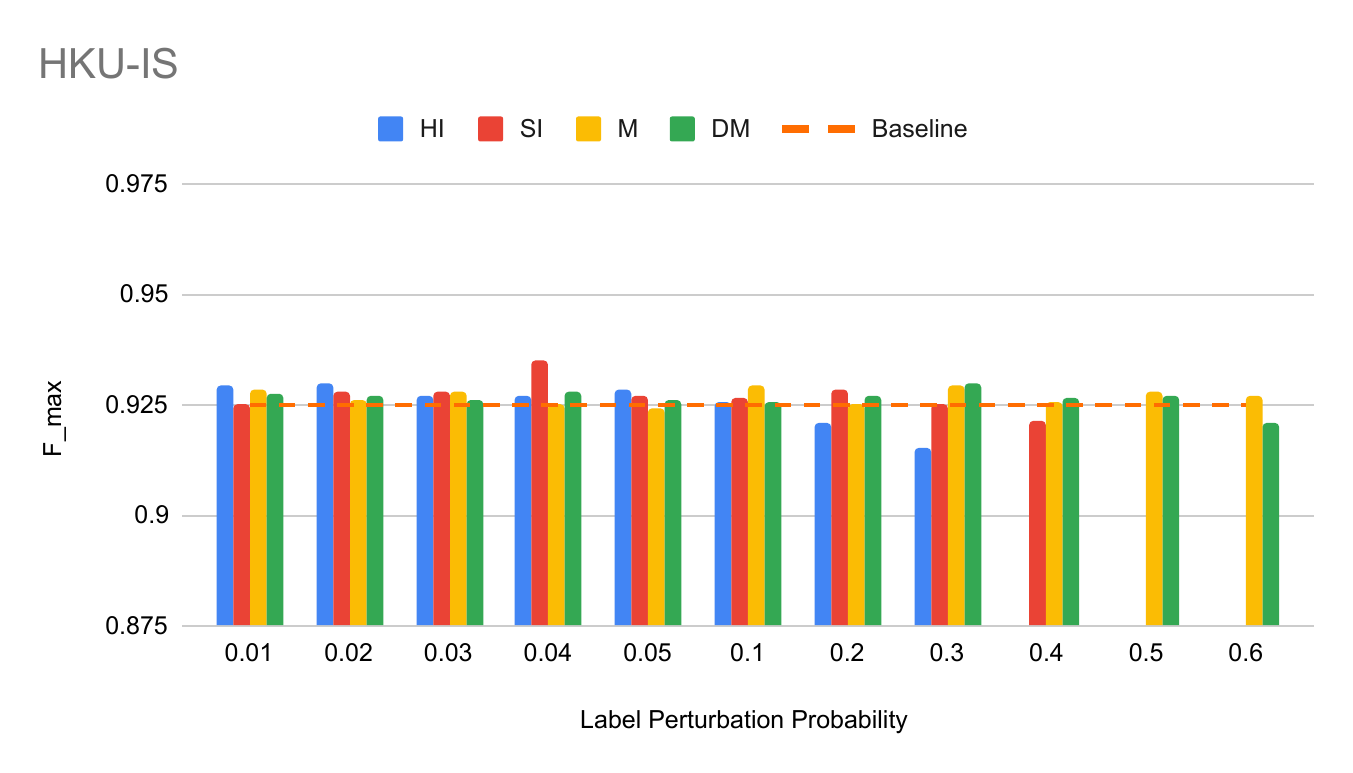}
    \caption{HKU-IS}
\end{subfigure}
\caption{Dense binary classification performance, evaluated in terms of maximum F measure, of various static stochastic label perturbation techniques under different label perturbation probabilities on the six testing datasets: (a): DUTS-TE, (b) DUT-OMRON, (c) PASCAL-S, (d) SOD, (e) ECSSD, (f) HKU-IS.}
\label{fig:Static_SLP_F_Measure}
\end{figure}

We present the dense binary classification performance, evaluated in terms of maximum F measure, of various static stochastic label perturbation techniques in Fig.~\ref{fig:Static_SLP_F_Measure}. It can be observed that in the effective label perturbation probability range for respective static SLP techniques, the dense binary classification performances are not negatively impacted. The performance drop is observed when the product $\alpha \beta$ is too high, \eg $\alpha \in [0.2, 0.3]$ for \enquote{HI}, $\alpha = 0.4$ for \enquote{SI}, and $\alpha = 0.6$ for \enquote{DM}. Overall, incorporation of static SLP techniques, with an effective label perturbation probability, can achieve improved model calibration degrees without sacrificing the dense bianry classification performance.

\begin{figure}
\centering
\begin{subfigure}{\textwidth}
    \includegraphics[width=\textwidth]{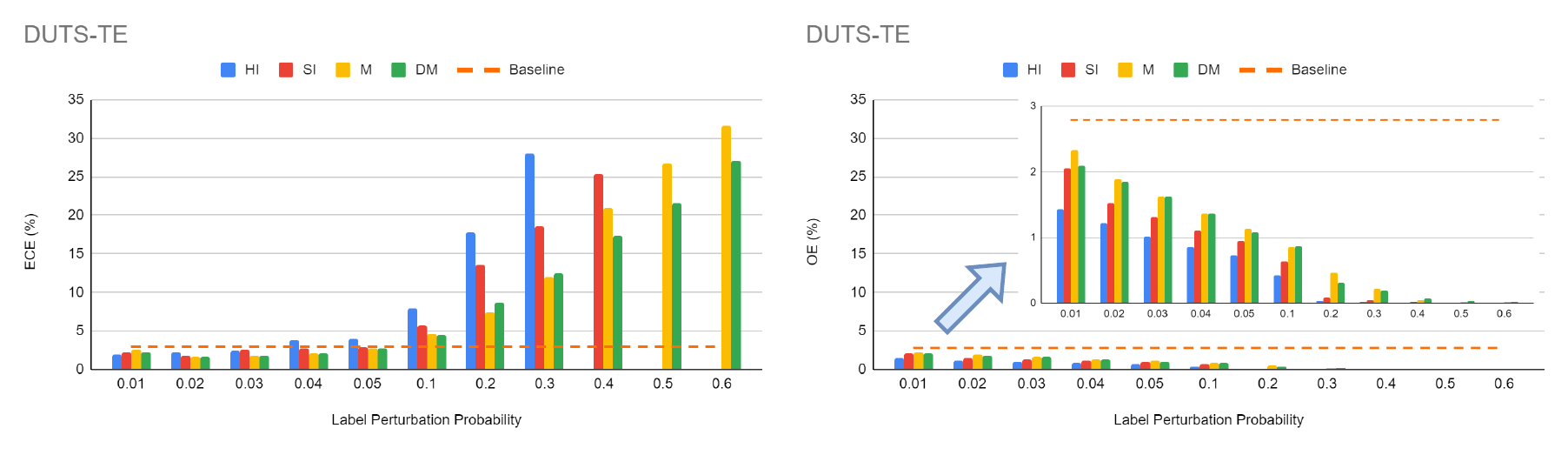}
    \caption{DUTS-TE}
\end{subfigure}
\begin{subfigure}{\textwidth}
    \includegraphics[width=\textwidth]{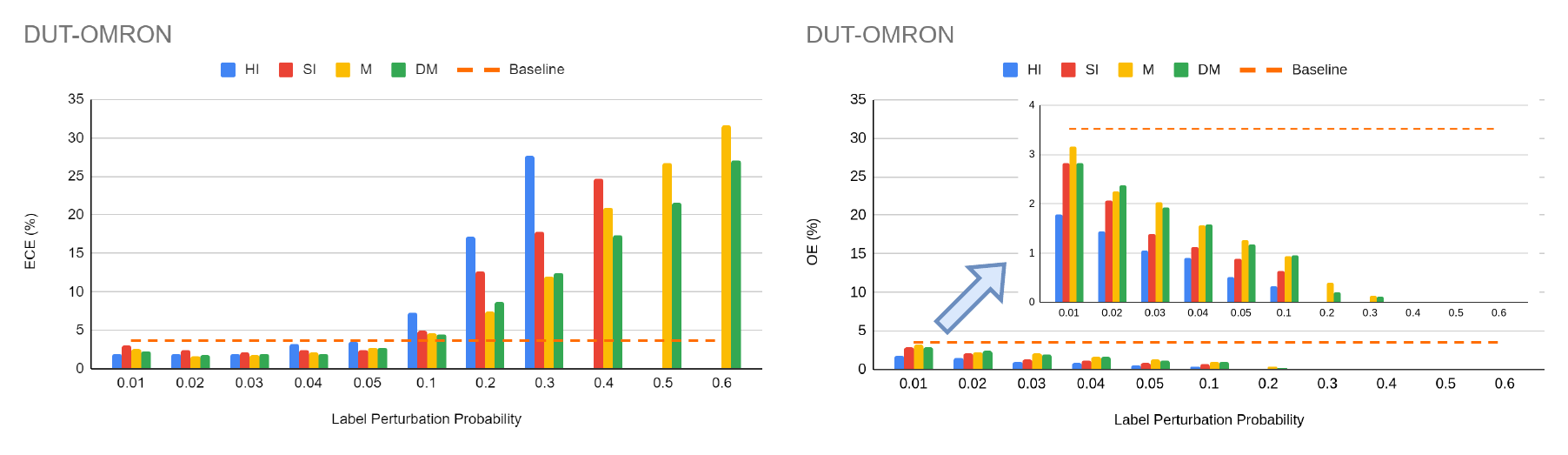}
    \caption{DUT-OMRON}
\end{subfigure}
\begin{subfigure}{\textwidth}
    \includegraphics[width=\textwidth]{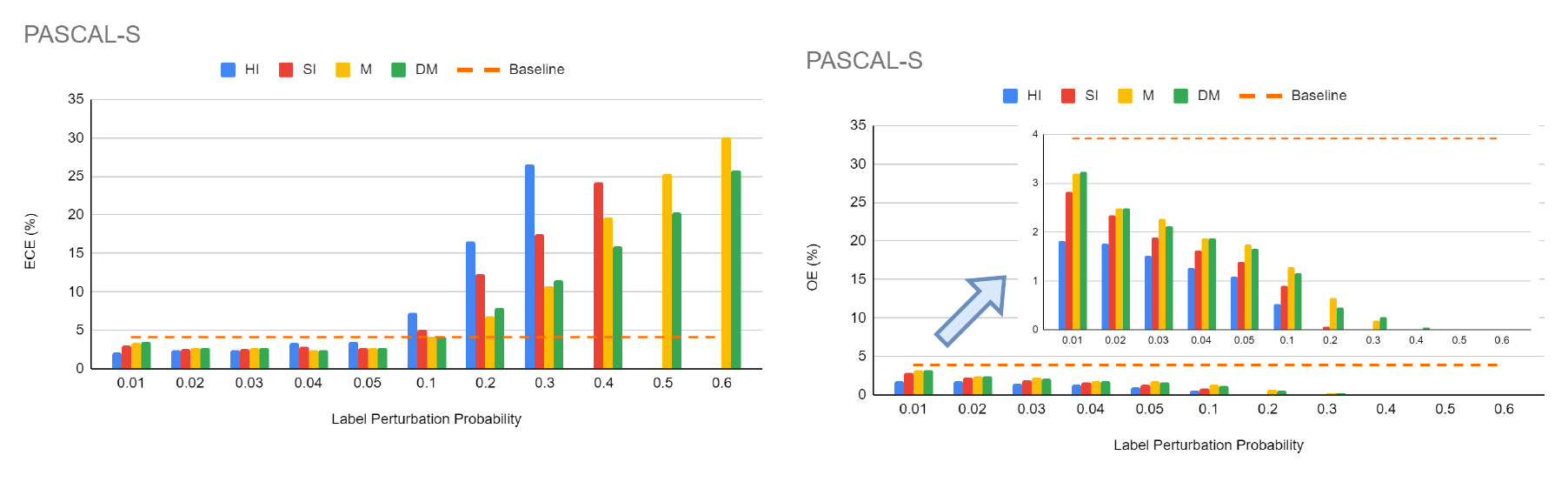}
    \caption{PASCAL-S}
\end{subfigure}
\caption{Model calibration degrees, evaluated in terms of Equal-Width Expected Calibration Error ($\text{ECE}_{\text{EW}}$) and Equal-Width Over-confidence Error ($\text{OE}_{\text{EW}}$) with 100 bins ($B = 100$), of various static stochastic label perturbation techniques under different label perturbation probabilities on the six testing datasets: (a): DUTS-TE, (b) DUT-OMRON, (c) PASCAL-S, (d) SOD, (e) ECSSD, (f) HKU-IS.}
\end{figure}

\begin{figure}
\ContinuedFloat
\centering
\begin{subfigure}{\textwidth}
    \includegraphics[width=\textwidth]{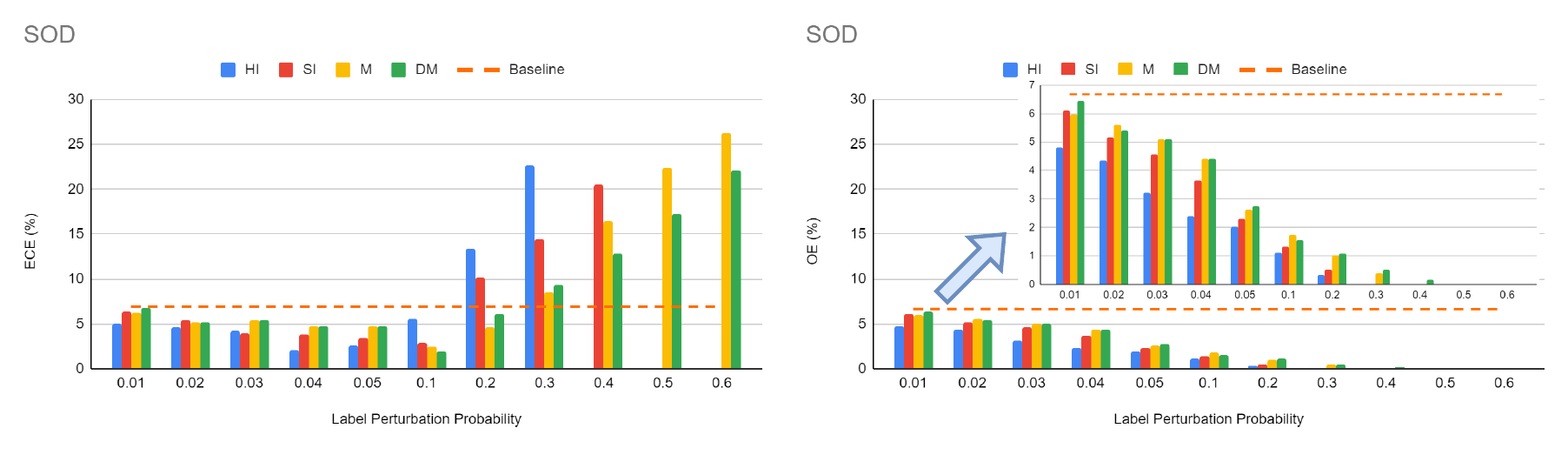}
    \caption{SOD}
\end{subfigure}
\begin{subfigure}{\textwidth}
    \includegraphics[width=\textwidth]{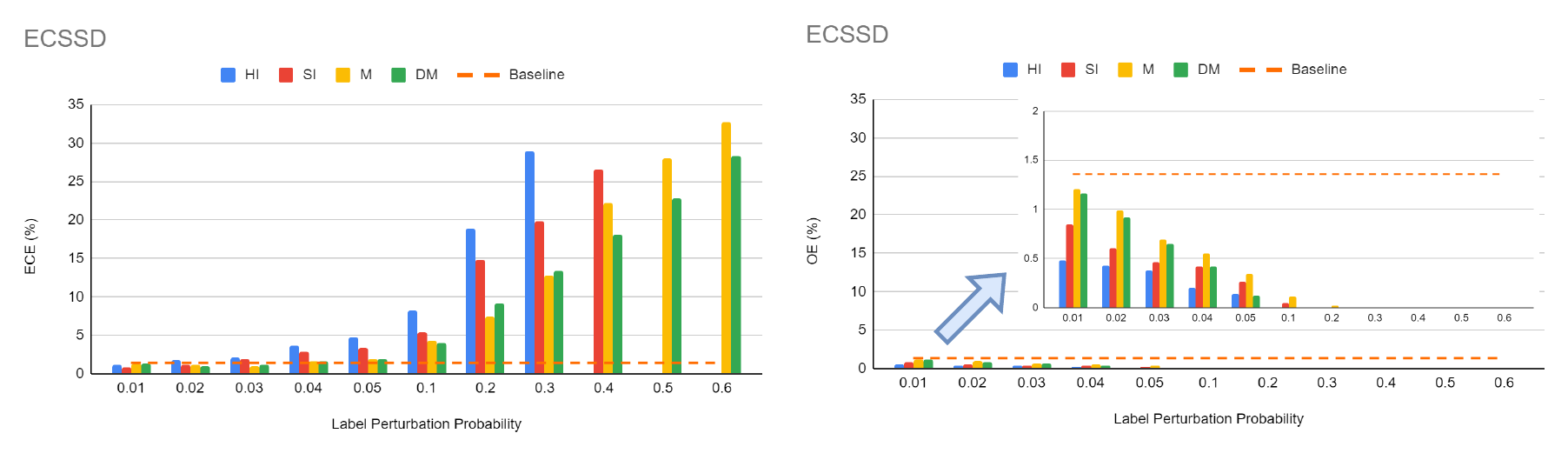}
    \caption{ECSSD}
\end{subfigure}
\begin{subfigure}{\textwidth}
    \includegraphics[width=\textwidth]{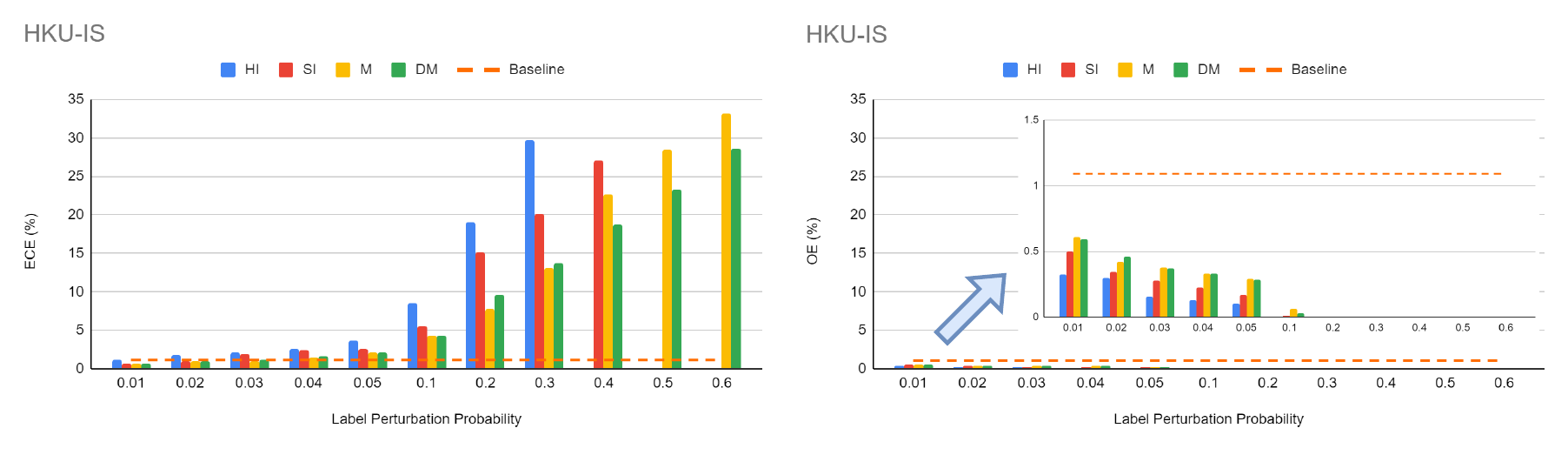}
    \caption{HKU-IS}
\end{subfigure}
\caption{Model calibration degrees, evaluated in terms of Equal-Width Expected Calibration Error ($\text{ECE}_{\text{EW}}$) and Equal-Width Over-confidence Error ($\text{OE}_{\text{EW}}$) with 100 bins ($B = 100$), of various static stochastic label perturbation techniques under different label perturbation probabilities on the six testing datasets: (a): DUTS-TE, (b) DUT-OMRON, (c) PASCAL-S, (d) SOD, (e) ECSSD, (f) HKU-IS.}
\label{fig:Static_SLP_Model_Calibration}
\end{figure}

\clearpage

\section{Experiments on Salient Object Detection with Additional Backbones}
\label{A_sec:Experiments_with_Additional_Backbones}
Experiments with additional backbones, VGG16 and Swin Transformer, are carried out on Salient Object Detection. We replace the ResNet50 backbone of the baseline model with VGG16 and Swin Transformer in respective experiments. We apply the proposed $\text{ASLP}_{\text{MC}}$ with with Hard Inversion (HI) and Soft Inversion (SI) label perturbation techniques and $\text{ALS}_{\text{MC}}$ to improve the model calibration degrees with respective backbones.

\begin{table*}[htb!]
\centering
\scriptsize
\renewcommand{\arraystretch}{1.2}
\renewcommand{\tabcolsep}{1.5mm}
\caption{Model calibration degrees with Swin transformer \cite{liu2021swin} backbone. Results are evaluated with Equal-Width Expected Calibration Error ($\text{ECE}_{\text{EW}}$) and Equal-Width Over-confidence Error ($\text{OE}_{\text{EW}}$) with 10 bins (units in (\%)).
}
\begin{tabular}{l|ccc|cc|cc|cc|cc|cc|cc}
\toprule
\multirow{2}{*}{Methods} & \multicolumn{3}{c|}{Perturbation Params} & \multicolumn{2}{c|}{DUTS-TE \cite{DUTS-TE}} & \multicolumn{2}{c|}{DUT-OMRON \cite{DUT-OMRON}} & \multicolumn{2}{c|}{PASCAL-S \cite{PASCAL-S}} & \multicolumn{2}{c|}{SOD \cite{SOD}} & \multicolumn{2}{c|}{ECSSD \cite{ECSSD}} & \multicolumn{2}{c}{HKU-IS \cite{HKU-IS}}\\
& $\alpha$ & $\beta$ & e & $\text{ECE} \downarrow$ & $\text{OE} \downarrow$ & $\text{ECE} \downarrow$ & $\text{OE} \downarrow$ & $\text{ECE} \downarrow$ & $\text{OE} \downarrow$ & $\text{ECE} \downarrow$ & $\text{OE} \downarrow$ & $\text{ECE} \downarrow$ & $\text{OE} \downarrow$ & $\text{ECE} \downarrow$ & $\text{OE} \downarrow$ \\
\midrule
Baseline (\enquote{Swin-B}) & 0 & 0 & 0 & 2.41 & 2.23 & 3.29 & 3.15 & 3.35 & 3.19 & 6.23 & 6.05 & 1.02 & 0.97 & 0.87 & 0.82\\
\hline
$\text{Swin-ASLP}_{\text{MC}}^{\text{HI}}$ & $\alpha_{\text{ada}}$ & 1.0 & \xmark & 1.44 & 1.21 & 1.73 & 1.59 & 1.74 & 1.57 & 5.08 & 4.85 & 0.57 &  0.30 & 0.81 & 0.23\\
$\text{Swin-ASLP}_{\text{MC}}^{\text{SI}}$ & $\alpha_{\text{ada}}$ & 0.75 & \xmark & 1.48 & 1.14 & 1.63 & 1.49 & 1.80 & 1.52 & 5.14 & 4.93 & 0.64 & 0.38 & 0.80 & 0.24\\
$\text{Swin-ALS}$ & 1.0 & $\beta_{\text{ada}}$ & \xmark & 1.44 & 1.14 & 1.76 & 1.57 & 1.69 & 1.55 & 5.17 & 4.82 & 0.54 & 0.36 & 0.77 & 0.24\\
\bottomrule
\end{tabular}
\label{tab:Swin_Transformer_Calibration_Degrees}
\end{table*}

\begin{table*}[htb!]
\centering
\scriptsize
\renewcommand{\arraystretch}{1.2}
\renewcommand{\tabcolsep}{1.5mm}
\caption{Dense classification accuracy with Swin transformer \cite{liu2021swin} backbone. Results are evaluated with maximum F-measure and maximum E-measure \cite{fan2018enhanced}.
}
\begin{tabular}{l|ccc|cc|cc|cc|cc|cc|cc}
\toprule
\multirow{2}{*}{Methods} & \multicolumn{3}{c|}{Perturbation Params} & \multicolumn{2}{c|}{DUTS-TE \cite{DUTS-TE}} & \multicolumn{2}{c|}{DUT-OMRON \cite{DUT-OMRON}} & \multicolumn{2}{c|}{PASCAL-S \cite{PASCAL-S}} & \multicolumn{2}{c|}{SOD \cite{SOD}} & \multicolumn{2}{c|}{ECSSD \cite{ECSSD}} & \multicolumn{2}{c}{HKU-IS \cite{HKU-IS}}\\
& $\alpha$ & $\beta$ & e & 
$F_{\text{max}} \uparrow$ & $E_{\text{max}} \uparrow$ & $F_{\text{max}} \uparrow$ & $E_{\text{max}} \uparrow$ & $F_{\text{max}} \uparrow$ & $E_{\text{max}} \uparrow$ & $F_{\text{max}} \uparrow$ & $E_{\text{max}} \uparrow$ & $F_{\text{max}} \uparrow$ & $E_{\text{max}} \uparrow$ & $F_{\text{max}} \uparrow$ & $E_{\text{max}} \uparrow$ \\
\midrule
Baseline (\enquote{Swin-B}) & 0 & 0 & 0 & 0.894 & 0.949 & 0.804 & 0.890 & 0.877 & 0.920 & 0.858 & 0.878 & 0.948 & 0.969 & 0.939 & 0.969\\
\hline
$\text{Swin-ASLP}_{\text{MC}}^{\text{HI}}$ & $\alpha_{\text{ada}}$ & 1.0 & \xmark & 0.895 & 0.953 & 0.808 & 0.892 & 0.881 & 0.924 & 0.959 & 0.879 & 0.950 & 0.969 & 0.938 & 0.969\\
$\text{Swin-ASLP}_{\text{MC}}^{\text{SI}}$ & $\alpha_{\text{ada}}$ & 0.75 & \xmark & 0.895 & 0.952 & 0.805 & 0.893 & 9,880 & 0.922 & 0.857 & 0.882 & 0.950 & 0.969 & 0.939 & 0.970\\
$\text{Swin-ALS}$ & 1.0 & $\beta_{\text{ada}}$ & \xmark & 0.895 & 0.952 & 0.804 & 0.892 & 0.879 & 0.920 & 0.859 & 0.879 & 0.948 & 0.969 & 0.939 & 0.970\\
\bottomrule
\end{tabular}
\label{tab:Swin_Transformer_Claasification_Accuracy}
\end{table*}

\begin{table*}[htb!]
\centering
\scriptsize
\renewcommand{\arraystretch}{1.2}
\renewcommand{\tabcolsep}{1.5mm}
\caption{Model calibration degrees with VGG16 \cite{simonyan2014very} backbone. Results are evaluated with Equal-Width Expected Calibration Error ($\text{ECE}_{\text{EW}}$) and Equal-Width Over-confidence Error ($\text{OE}_{\text{EW}}$) with 10 bins (units in (\%)).
}
\begin{tabular}{l|ccc|cc|cc|cc|cc|cc|cc}
\toprule
\multirow{2}{*}{Methods} & \multicolumn{3}{c|}{Perturbation Params} & \multicolumn{2}{c|}{DUTS-TE \cite{DUTS-TE}} & \multicolumn{2}{c|}{DUT-OMRON \cite{DUT-OMRON}} & \multicolumn{2}{c|}{PASCAL-S \cite{PASCAL-S}} & \multicolumn{2}{c|}{SOD \cite{SOD}} & \multicolumn{2}{c|}{ECSSD \cite{ECSSD}} & \multicolumn{2}{c}{HKU-IS \cite{HKU-IS}}\\
& $\alpha$ & $\beta$ & e & $\text{ECE} \downarrow$ & $\text{OE} \downarrow$ & $\text{ECE} \downarrow$ & $\text{OE} \downarrow$ & $\text{ECE} \downarrow$ & $\text{OE} \downarrow$ & $\text{ECE} \downarrow$ & $\text{OE} \downarrow$ & $\text{ECE} \downarrow$ & $\text{OE} \downarrow$ & $\text{ECE} \downarrow$ & $\text{OE} \downarrow$ \\
\midrule
Baseline (\enquote{VGG-B}) & 0 & 0 & 0 & 3.46 & 3.23 & 4.12 & 3.92 & 4.40 & 4.17 & 7.87 & 7.60 & 2.02 & 1.91 & 1.51 & 1.44\\
\hline
$\text{VGG-ASLP}_{\text{MC}}^{\text{HI}}$ & $\alpha_{\text{ada}}$ & 1.0 & \xmark & 1.44 & 1.28 & 1.91 & 1.82 & 2.40 & 2.16 & 5.44 & 5.08 & 0.57 & 0.21 & 0.84 & 0.16\\
$\text{VGG-ASLP}_{\text{MC}}^{\text{SI}}$ & $\alpha_{\text{ada}}$ & 0.75 & \xmark & 1.47 & 1.23 & 2.05 & 1.81 & 2.34 & 2.15 & 5.54 & 5.22 & 0.51 & 0.21 & 0.88 & 0.19\\
$\text{VGG-ALS}$ & 1.0 & $\beta_{\text{ada}}$ & \xmark & 1.48 & 1.31 & 1.99 & 1.76 & 2.33 & 2.04 & 5.53 & 5.14 & 0.45 & 0.29 & 0.82 & 0.13\\
\bottomrule
\end{tabular}
\label{tab:VGG16_Calibration_Degrees}
\end{table*}

\begin{table*}[htb!]
\centering
\scriptsize
\renewcommand{\arraystretch}{1.2}
\renewcommand{\tabcolsep}{1.5mm}
\caption{Dense classification accuracy with VGG16 \cite{simonyan2014very} backbone. Results are evaluated with maximum F-measure and maximum E-measure \cite{fan2018enhanced}.
}
\begin{tabular}{l|ccc|cc|cc|cc|cc|cc|cc}
\toprule
\multirow{2}{*}{Methods} & \multicolumn{3}{c|}{Perturbation Params} & \multicolumn{2}{c|}{DUTS-TE \cite{DUTS-TE}} & \multicolumn{2}{c|}{DUT-OMRON \cite{DUT-OMRON}} & \multicolumn{2}{c|}{PASCAL-S \cite{PASCAL-S}} & \multicolumn{2}{c|}{SOD \cite{SOD}} & \multicolumn{2}{c|}{ECSSD \cite{ECSSD}} & \multicolumn{2}{c}{HKU-IS \cite{HKU-IS}}\\
& $\alpha$ & $\beta$ & e & 
$F_{\text{max}} \uparrow$ & $E_{\text{max}} \uparrow$ & $F_{\text{max}} \uparrow$ & $E_{\text{max}} \uparrow$ & $F_{\text{max}} \uparrow$ & $E_{\text{max}} \uparrow$ & $F_{\text{max}} \uparrow$ & $E_{\text{max}} \uparrow$ & $F_{\text{max}} \uparrow$ & $E_{\text{max}} \uparrow$ & $F_{\text{max}} \uparrow$ & $E_{\text{max}} \uparrow$ \\
\midrule
Baseline (\enquote{VGG-B}) & 0 & 0 & 0 & 0.838 & 0.912 & 0.741 & 0.851 & 0.844 & 0.895 & 0.810 & 0.851 & 0.921 & 0.944 & 0.913 & 0.950\\
\hline
$\text{VGG-ASLP}_{\text{MC}}^{\text{HI}}$ & $\alpha_{\text{ada}}$ & 1.0 & \xmark & 0.844 & 0.916 & 0.746 & 0.857 & 0.844 & 0.896 & 0.812 & 0.851 & 0.921 & 0.944 & 0.913 & 0.951\\
$\text{VGG-ASLP}_{\text{MC}}^{\text{SI}}$ & $\alpha_{\text{ada}}$ & 0.75 & \xmark & 0.845 & 0.916 & 0.747 & 0.855 & 0.846 & 0.895 & 0.810 & 0.851 & 0.921 & 0.944 & 0.916 & 0.953\\
$\text{VGG-ALS}$ & 1.0 & $\beta_{\text{ada}}$ & \xmark & 0.843 & 0.914 & 0.745 & 0.857 & 0.848 & 0.898 & 0.811 & 0.852 & 0.921 & 0.945 & 0.913 & 0.952\\
\bottomrule
\end{tabular}
\label{tab:VGG16_Claasification_Accuracy}
\end{table*}

\clearpage

\section{Hyperparameters}
\label{A_sec:Hyperparameters}

\begin{figure}[h!]
\centering
\begin{subfigure}[b]{.48\textwidth}
    \centering
    \includegraphics[width=\textwidth]{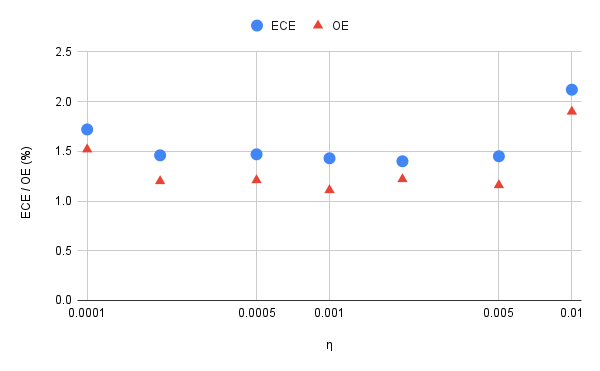}
    \caption{Learning Rate ($\eta$)}
    \label{sub_fig:ablation_study_on_learning_rate_eta}
\end{subfigure}
\hfill
\begin{subfigure}[b]{.48\textwidth}
    \centering
    \includegraphics[width=\textwidth]{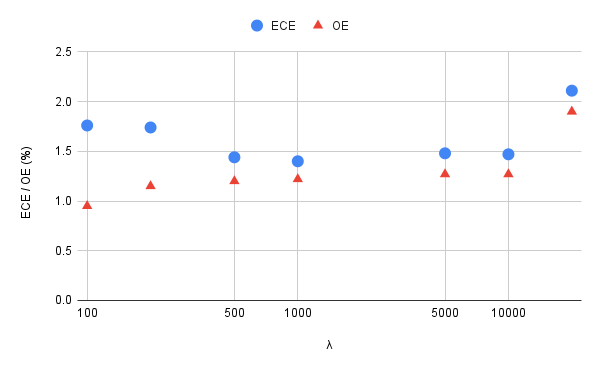}
    \caption{Regularisation Strength ($\lambda$)}
\label{sub_fig:ablation_study_on_regularisation_strength_lambda}
\end{subfigure}
\caption{Ablation study on hyperparameters: (1) learning rate ($\eta$) and (2) regularisation strength ($\lambda$) evaluated in terms of $\text{ECE}_{\text{EW}}$ and $\text{OE}_{\text{EW}}$ with 100 bins on the DUTS-TE dataset.}
\label{fig:ablation_hyperparameters}
\end{figure}

\section{Training and Inference Time}
In SOD, the training of ASLP on DUTS-TR requires 2.5 hours, which is 0.2 hours longer (or $\sim8.7$\% more) than training the base model (2.3 hours). The inference speed of ASLP on the six SOD testing datasets averages: 53.40 samples per second, which is the same as that of the base model because of the same network architecture. Both training and inference time are evaluated on a single Geforce RTX 3090 GPU.


\clearpage

\section{500 Texture Images from Describable Texture Dataset}

\begin{figure}[h!]
    \centering
    \includegraphics[width=\textwidth]{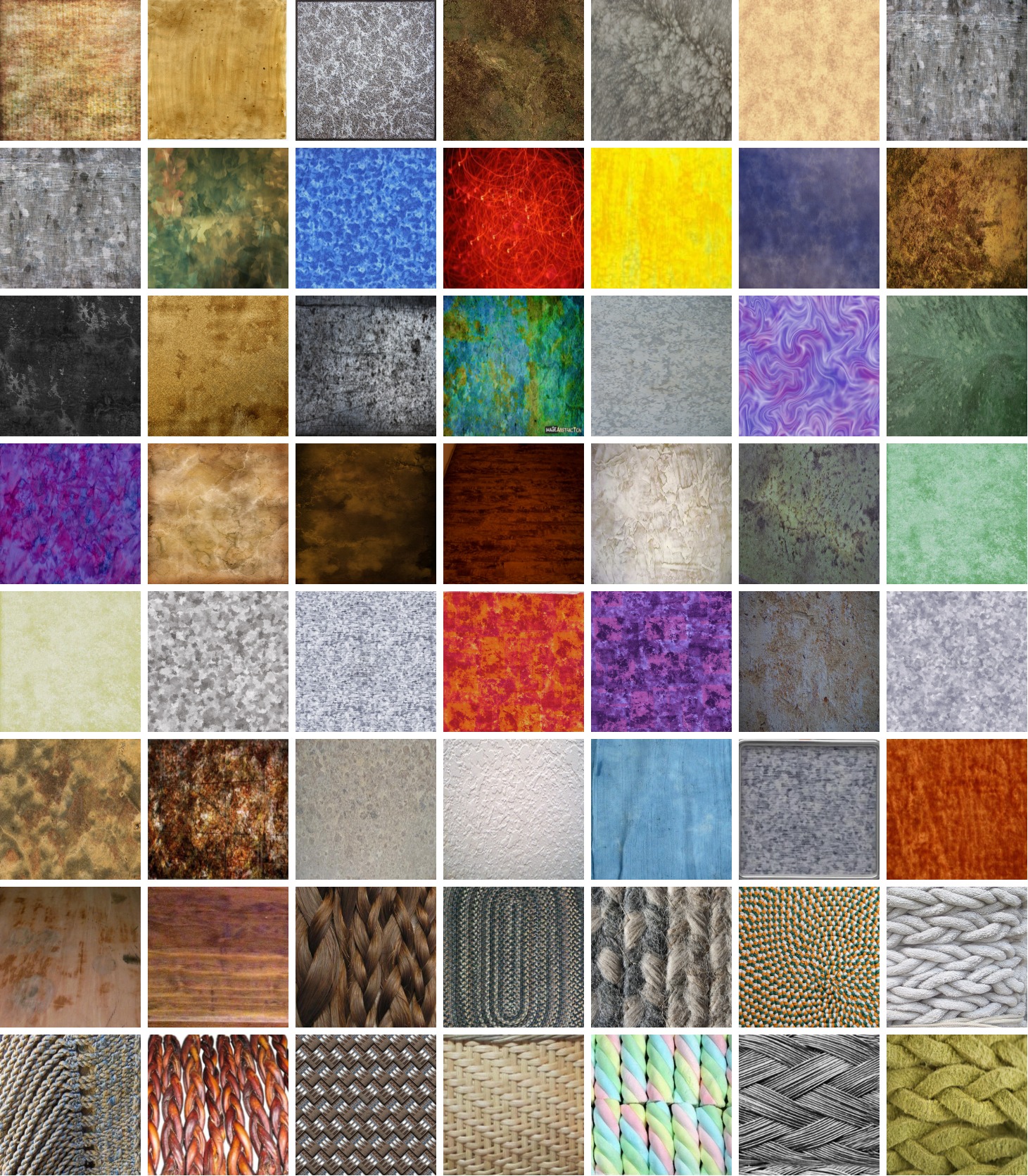}
    \caption{Texture images without visually salient objects selected from Describable Texture Dataset \cite{cimpoi2014describing}.}
\end{figure}

\begin{figure}[h!]
    \ContinuedFloat
    \centering
    \includegraphics[width=\textwidth]{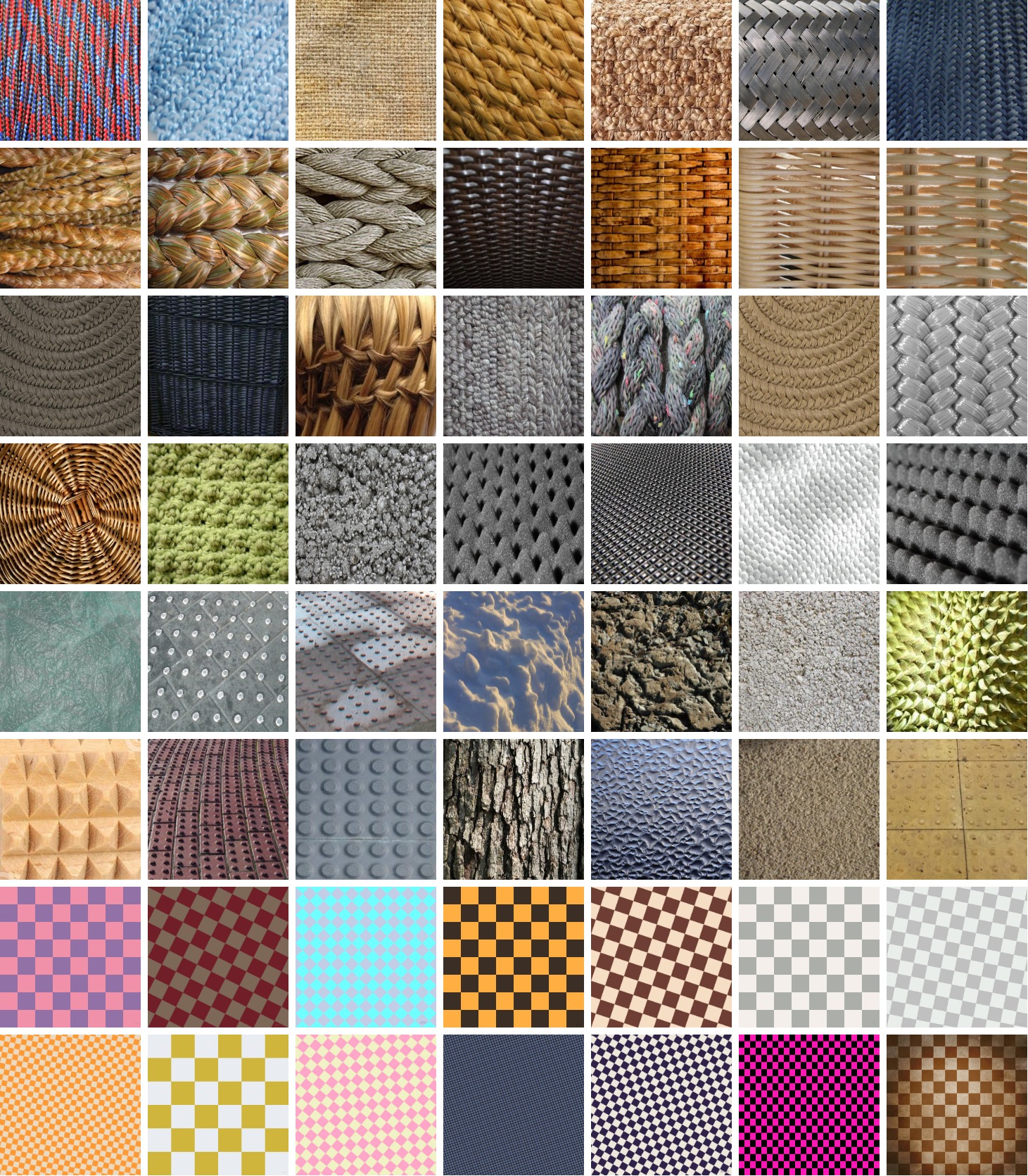}
    \caption{Texture images without visually salient objects selected from Describable Texture Dataset \cite{cimpoi2014describing}.}
\end{figure}

\begin{figure}[h!]
    \ContinuedFloat
    \centering
    \includegraphics[width=\textwidth]{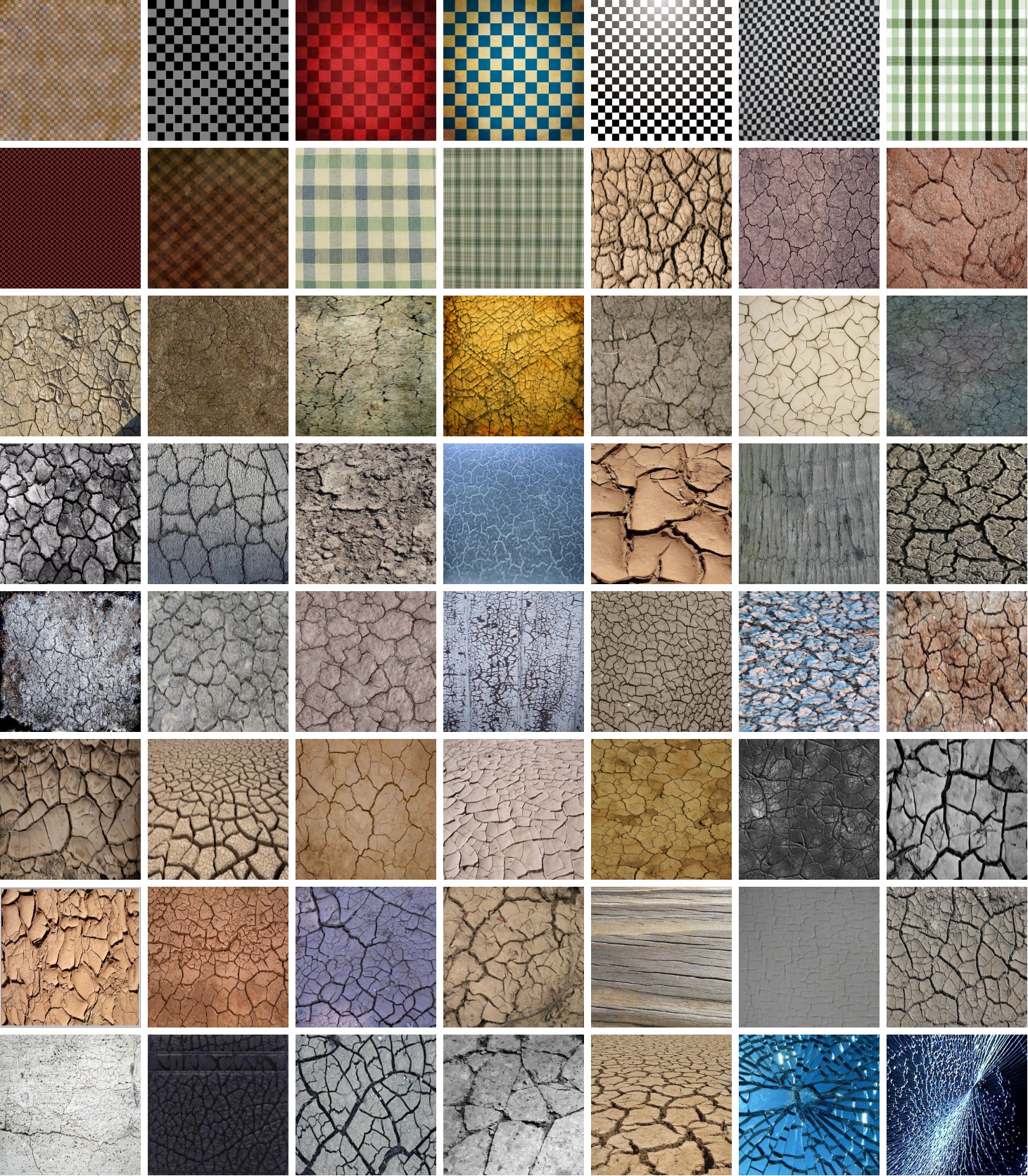}
    \caption{Texture images without visually salient objects selected from Describable Texture Dataset \cite{cimpoi2014describing}.}
\end{figure}

\begin{figure}[h!]
    \ContinuedFloat
    \centering
    \includegraphics[width=\textwidth]{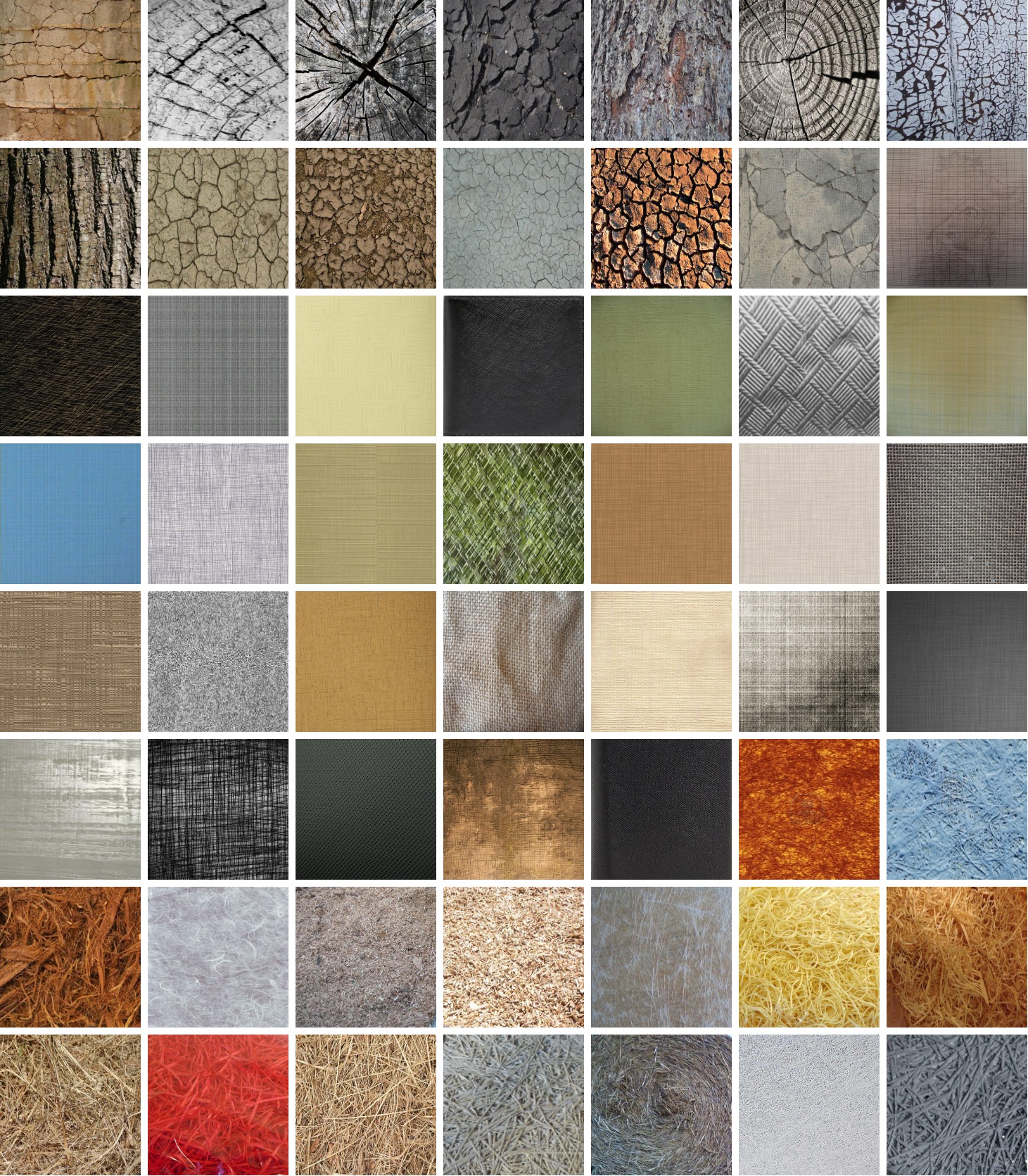}
    \caption{Texture images without visually salient objects selected from Describable Texture Dataset \cite{cimpoi2014describing}.}
\end{figure}

\begin{figure}[h!]
    \ContinuedFloat
    \centering
    \includegraphics[width=\textwidth]{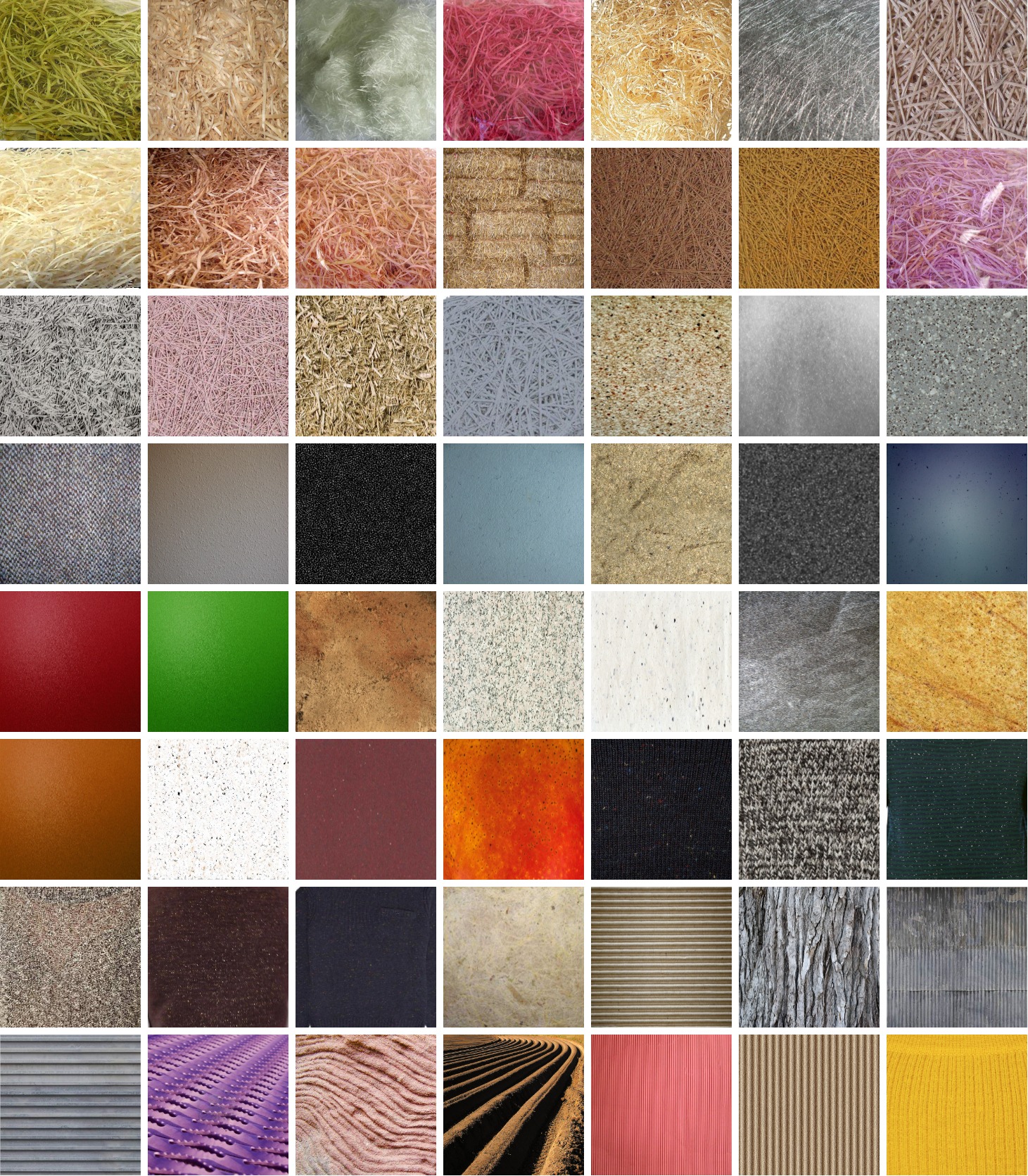}
    \caption{Texture images without visually salient objects selected from Describable Texture Dataset \cite{cimpoi2014describing}.}
\end{figure}

\begin{figure}[h!]
    \ContinuedFloat
    \centering
    \includegraphics[width=\textwidth]{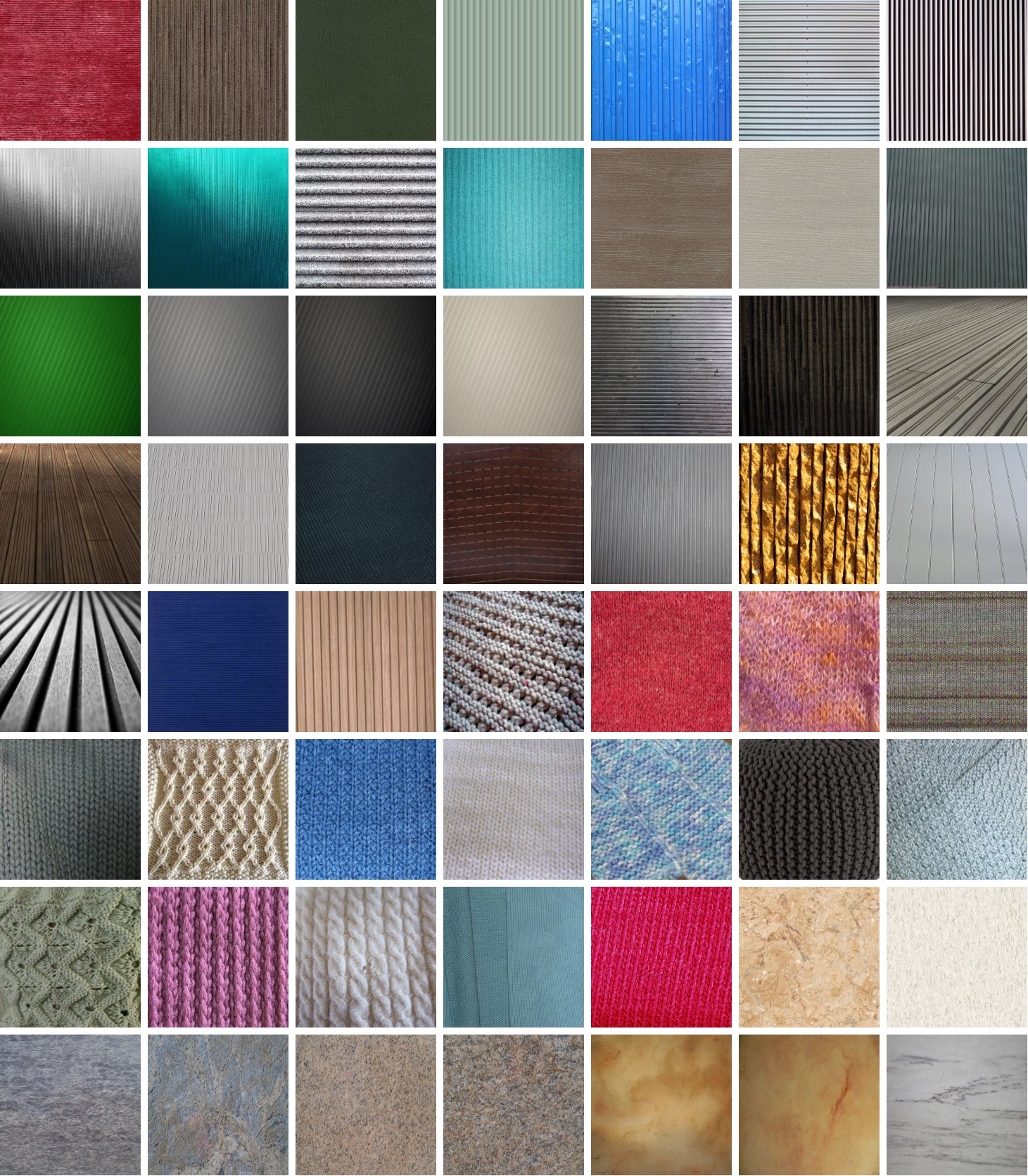}
    \caption{Texture images without visually salient objects selected from Describable Texture Dataset \cite{cimpoi2014describing}.}
\end{figure}

\begin{figure}[h!]
    \ContinuedFloat
    \centering
    \includegraphics[width=\textwidth]{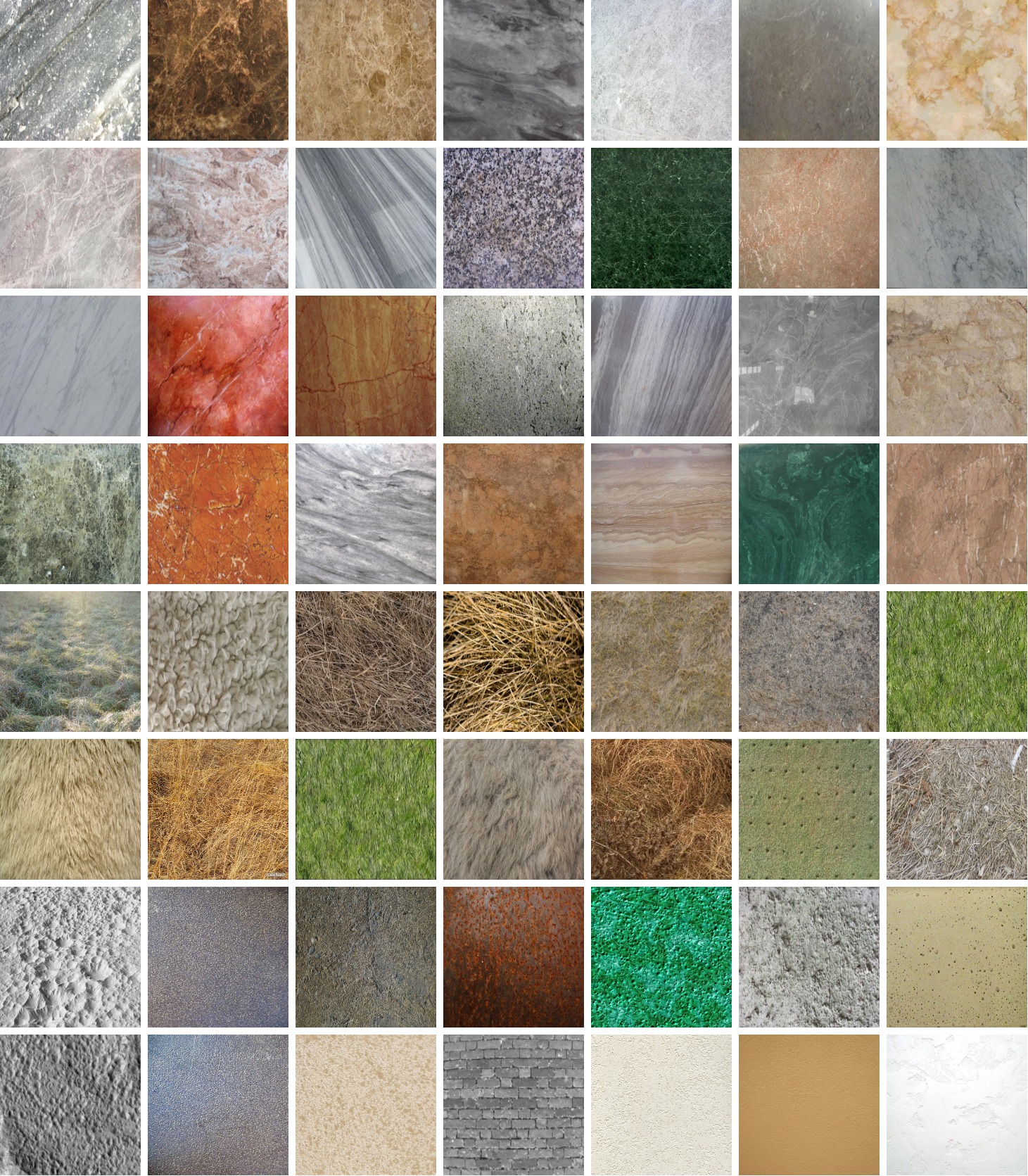}
    \caption{Texture images without visually salient objects selected from Describable Texture Dataset \cite{cimpoi2014describing}.}
\end{figure}

\begin{figure}[h!]
    \ContinuedFloat
    \centering
    \includegraphics[width=\textwidth]{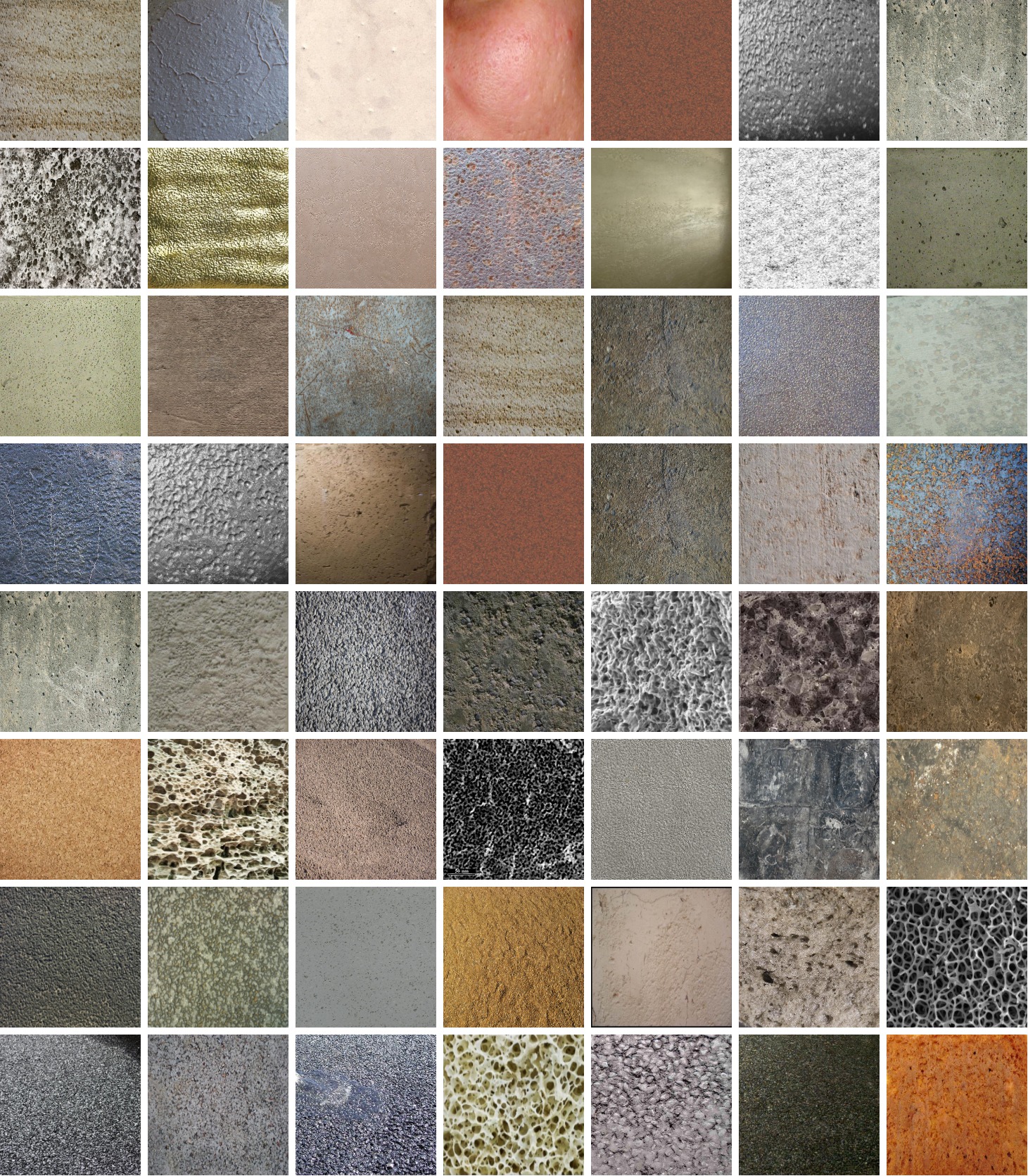}
    \caption{Texture images without visually salient objects selected from Describable Texture Dataset \cite{cimpoi2014describing}.}
\end{figure}

\begin{figure}[h!]
    \ContinuedFloat
    \centering
    \includegraphics[width=\textwidth]{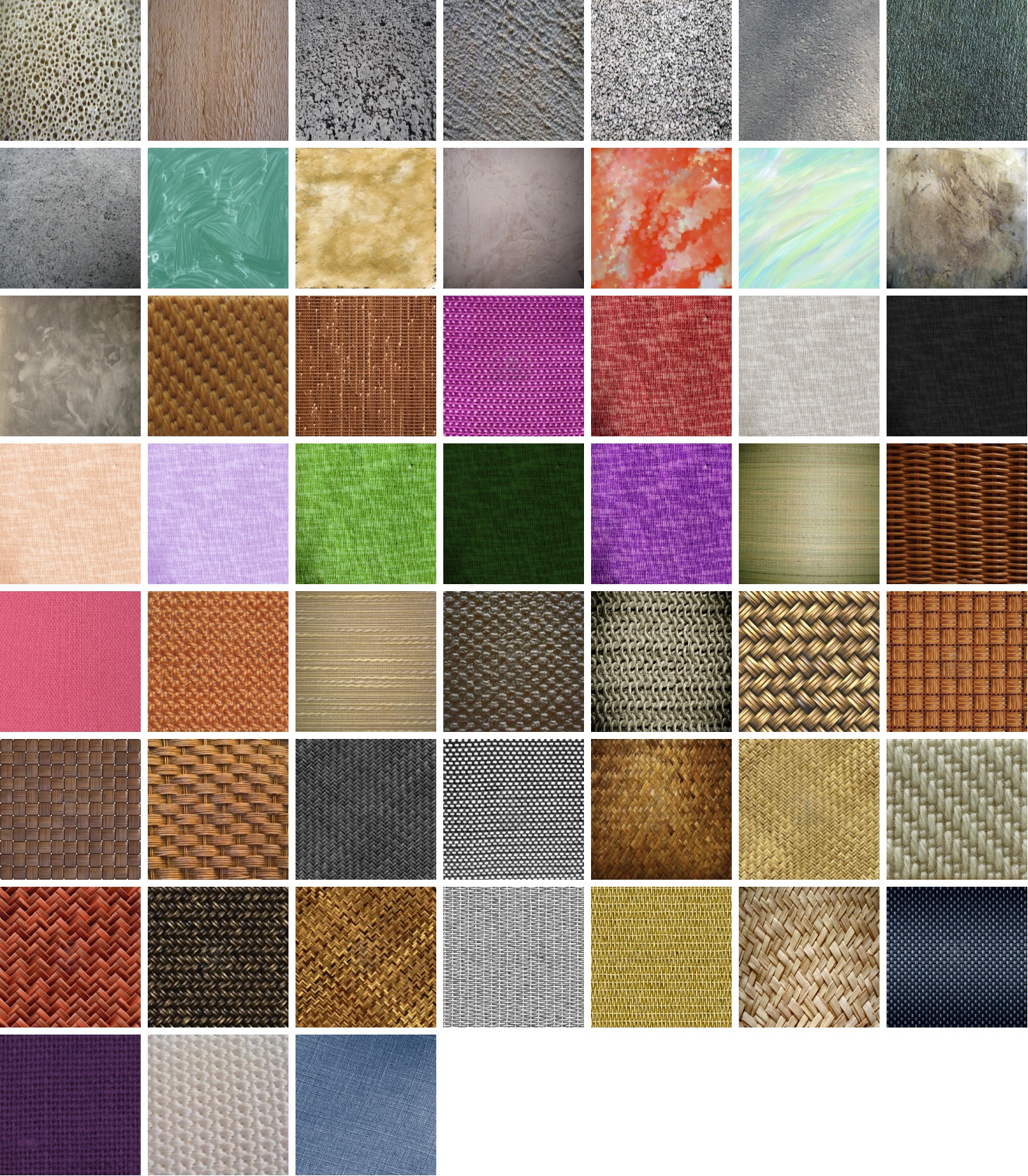}
    \caption{Texture images without visually salient objects selected from Describable Texture Dataset \cite{cimpoi2014describing}.}
    \label{A_fig:Full_Texture_Image_Collection}
\end{figure}